\documentclass[11pt, a4paper, logo, copyright]{hail_lab}

\usepackage[utf8]{inputenc}
\usepackage[T1]{fontenc}
\usepackage{microtype}

\usepackage{amsmath,amssymb,amsfonts,bm}
\usepackage{amsthm}
\usepackage{mathtools}
\usepackage{bbm}

\usepackage{booktabs}
\usepackage{array}
\usepackage{tabularx}
\usepackage{ragged2e}
\usepackage{makecell}
\usepackage{multirow}
\usepackage{colortbl}
\usepackage{longtable}

\usepackage{graphicx}
\usepackage{subcaption}
\usepackage[export]{adjustbox}
\usepackage{wrapfig}
\usepackage{float}

\colorlet{darkgreen}{green!65!black}
\colorlet{darkblue}{blue!75!black}
\colorlet{darkred}{red!80!black}
\definecolor{statistical}{HTML}{8c564b}
\definecolor{structural}{HTML}{0070C0}
\definecolor{semantic}{HTML}{008080}
\definecolor{yellow}{HTML}{f7c600}
\definecolor{lightblue}{HTML}{0071bc}
\definecolor{lightgreen}{HTML}{39b54a}
\definecolor{mypurple}{HTML}{412F8A}
\definecolor{myorange}{HTML}{fc8e62}
\definecolor{textgreen}{RGB}{57, 172, 57}
\definecolor{textred}{RGB}{200, 10, 10}
\definecolor{deemph}{gray}{0.55}
\definecolor{baselinecolor}{gray}{.95}
\definecolor{graycolor}{gray}{.95}

\usepackage[pagebackref=false, breaklinks=true, colorlinks, linkcolor=darkred, citecolor=lightblue, urlcolor=lightblue, bookmarks=false]{hyperref}
\usepackage{url}
\usepackage[capitalize,noabbrev]{cleveref}
\usepackage{cite}

\usepackage{enumitem}

\usepackage{xspace}
\usepackage{pbox}
\usepackage{nicefrac}
\usepackage{afterpage}
\usepackage{textcomp}
\usepackage{siunitx}
\usepackage{soul}
\usepackage{etoolbox}
\usepackage[most]{tcolorbox}
\tcbset{
    frame code={}
    center title,
    left=0pt,
    right=0pt,
    top=0pt,
    bottom=0pt,
    colframe=white,
    width=\dimexpr\textwidth\relax,
    enlarge left by=0mm,
    boxsep=5pt,
    arc=0pt,outer arc=0pt,
}
\usepackage{pifont}
\usepackage{bbding}
\usepackage[dvipsnames]{xcolor}

\newcolumntype{L}[1]{>{\RaggedRight\arraybackslash}p{#1}}
\newcolumntype{Y}{>{\raggedright\arraybackslash}X}
\newcolumntype{x}[1]{>{\centering\arraybackslash}p{#1pt}}
\newcolumntype{y}[1]{>{\raggedright\arraybackslash}p{#1pt}}
\newcolumntype{z}[1]{>{\raggedleft\arraybackslash}p{#1pt}}

\newlength\savewidth
\newcommand{\grayrow}{\rowcolor[gray]{.95}}

\renewcommand{\tilde}{\widetilde}
\renewcommand{\hat}{\widehat}

\newcommand{\dsr}{Deep Spurious Regression\xspace}

\newcommand{\labelmds}{\texttt{L-MDS}\xspace}
\newcommand{\featuremds}{\texttt{F-MDS}\xspace}

\newcommand{\ColoredRotatedMNIST}{\texttt{ColoredRotatedMNIST}\xspace}
\newcommand{\UTKFace}{\texttt{UTKFace}\xspace}
\newcommand{\SkyFinder}{\texttt{SkyFinder}\xspace}
\newcommand{\PovertyMap}{\texttt{PovertyMap}\xspace}
\newcommand{\CodeNet}{\texttt{CodeNet}\xspace}

\newenvironment{Itemize}{
    \begin{itemize}[leftmargin=*]
    \setlength{\itemsep}{0pt}
    \setlength{\topsep}{0pt}
    \setlength{\partopsep}{0pt}
    \setlength{\parskip}{0pt}}
{\end{itemize}}

\theoremstyle{plain}

\theoremstyle{definition}

\theoremstyle{remark}

\usepackage{titletoc}
\usepackage{letltxmacro}

\title{Shortcut to Nowhere: Demystifying Deep Spurious Regression}

\author[1]{Guanrong Xu}
\author[1]{Jessica Li}
\author[2]{Hao Wang}
\author[1$\dagger$]{Yuzhe Yang}

\affil[1]{University of California, Los Angeles}
\affil[2]{Rutgers University}

\correspondingauthor{$^\dagger$Correspondence to: yuzhey@ucla.edu.}

\hflink{https://hf.co/yang-ai-lab/Deep-Spurious-Regression}
\codelink{https://github.com/yang-ai-lab/Deep-Spurious-Regression}
\projecturl{https://yang-ai-lab.github.io/Deep-Spurious-Regression}

\begin{abstract}
\vspace{-5pt}
Real-world regression often exhibits shortcuts: attributes that are spuriously correlated with continuous targets in training, yet unreliable under deployment shifts; 
regressing targets using such shortcuts may fail catastrophically at test time. 
Existing studies on spurious correlations focus primarily on classification, where labels are categorical and groups are naturally defined. However, many real-world tasks require continuous prediction, where hard label boundaries or discrete group-label pairs do not exist.
We define \dsr (DSR) as learning from regression data with attribute-label confounding, addressing continuous spurious correlations, and generalizing to all attribute-label combinations at test time. Motivated by the intrinsic difference between classification and regression shortcuts, we propose to exploit the similarity among spurious attributes in both label and feature spaces, thereby accounting for nearby targets and related groups while calibrating both label and learned feature distributions across attributes.
Extensive experiments on common real-world DSR datasets that span computer vision, environmental sensing, and large language model (LLM) regression verify the superior performance of our strategies. Our work fills the gap in benchmarks and techniques for studying spurious correlations in continuous prediction.

\end{abstract}

\begin{document}

\maketitle

\section{Introduction}
\label{sec:intro}

Spurious correlations are ubiquitous and inherent in real-world observational data \cite{yang2024limits,yang2023change}. Rather than preserving a stable relationship between target-relevant features and labels, the data often exhibit shortcut correlations, where certain attributes are highly predictive of the target during training but unreliable at deployment \cite{geirhos2020shortcut, yang2023change}. This phenomenon poses great challenges for deep learning models and has motivated many prior techniques for addressing spurious correlations and subgroup failures \cite{sagawa2020dro, liu2021jtt, nam2020lff, kirichenko2023dfr, creager2021environment, zhang2022cnc, holste2024towards}.

\begin{figure}[!t]
\centering
\includegraphics[width=1\textwidth]{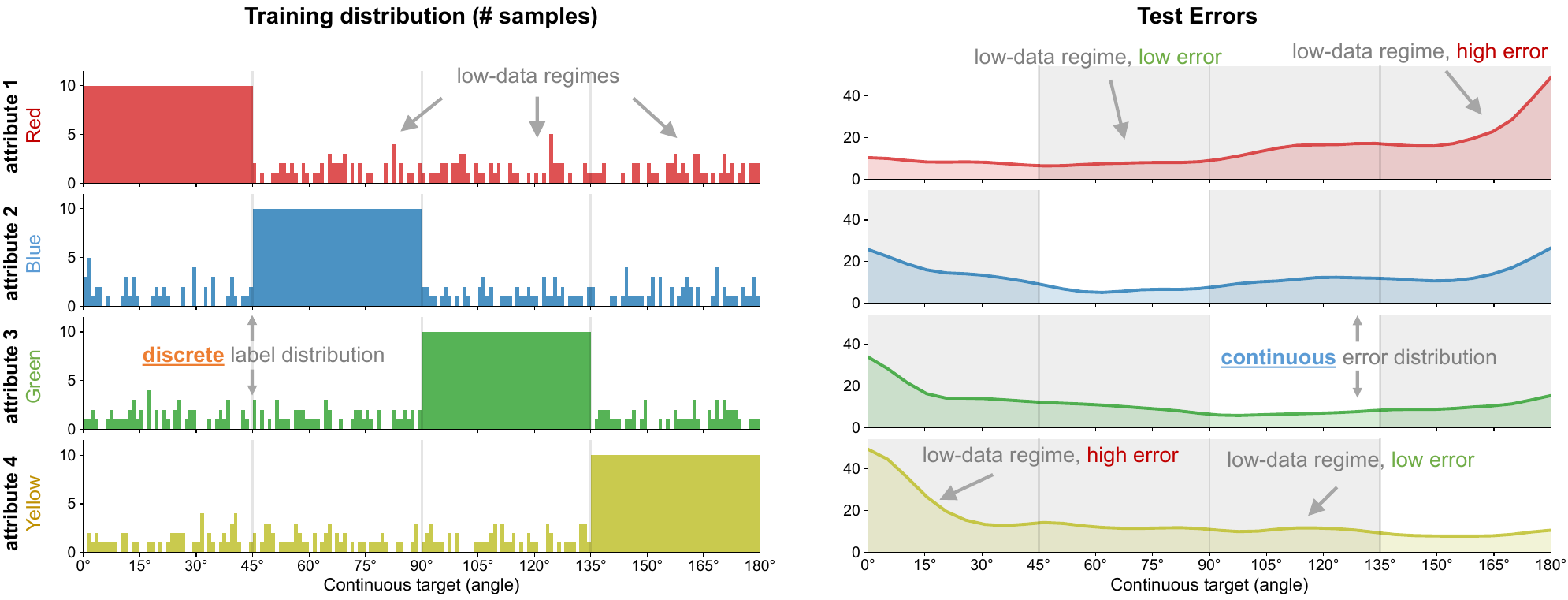}
\caption{\small{
\textbf{Example illustration of \dsr (DSR).} \textbf{Left:} In \ColoredRotatedMNIST, each spurious attribute, represented by a background color, is strongly associated with a dominant angle range in training, while other angle ranges have few samples. \textbf{Right:} ERM produces continuous test-error curves across the target angle. Importantly, low-data regimes do not always lead to high error: certain sparse regions remain easy, while others fail sharply when the color-angle shortcut breaks. This shows that DSR cannot be captured by discrete group counts alone. More details are in Appendix \ref{app:colored_rotated_mnist_results}.
}}
\label{fig:toy_dsr}
\vspace{-8pt}
\end{figure}

Existing solutions for spurious correlations, however, focus on targets with \textit{\textbf{categorical}} indices, i.e., the targets are different classes, and subgroups can be naturally defined by finite label-attribute combinations.
However, many real-world tasks involve \textit{\textbf{continuous}} and even infinite target values, where discrete group definitions do not exist.
Unfortunately, standard models trained on such data learn fragmented and shortcut-driven mappings that are incapable of capturing the continuous relationships that underlie regression tasks.
Fig. \ref{fig:toy_dsr} illustrates this failure mode on \ColoredRotatedMNIST, where the target is the rotation angle and the spurious attribute is the background color (details in Appendix \ref{app:colored_rotated_mnist_details}).
Rather than learning the continuous rotation structure, standard empirical risk minimization (ERM) \cite{vapnik1998statistical} produces error patterns tied to the \textit{color-angle shortcut}: some low-data regions have low error, while others fail sharply when the shortcut breaks. Such a \textit{continuous} and \textit{attribute-dependent} error distribution is suboptimal for regression and cannot be fully captured by discrete group counts.

In this work, we systematically investigate \dsr (DSR) arising in real-world settings. We define DSR as learning continuous targets from data with attribute-label confounding, dealing with potentially sparse or missing data for certain attribute-label combinations, and generalizing to a test set that is balanced over the entire range of continuous target values and spurious attributes. This definition is analogous to the spurious correlation problem in classification \cite{sagawa2020dro}, but focuses on the continuous setting.

DSR brings new challenges distinct from its classification counterpart. First, given continuous and potentially infinite target values, discrete groups are no longer naturally defined, causing ambiguity when directly applying traditional debiasing methods such as re-sampling, re-weighting, and group-robust optimization \cite{sagawa2020dro}. 
Second, both nearby labels and related attributes carry meaningful information for interpreting spurious correlations. For example, two weakly observed targets under one attribute may differ substantially if one is supported by nearby labels or similar attributes, while the other lies in a sparse neighborhood.
Finally, unlike classification, certain attribute-label combinations may have no data at all, which motivates the need for interpolation and extrapolation across targets and attributes.

To fill these gaps, we propose two simple yet effective methods for addressing DSR:
label multi-dimensional scaling (\labelmds) and feature multi-dimensional scaling (\featuremds). A key idea underlying both approaches is to leverage the \textit{similarity} among spurious attributes by pooling information across similar groups and attributes using Multi-Dimensional Scaling (MDS) \cite{borg2005modern} and kernel smoothing to perform explicit calibration in the label and feature spaces.
Both techniques can be easily embedded into existing deep networks or large language models (LLMs) and allow optimization in an end-to-end fashion. We verify that our techniques not only calibrate for the intrinsic underlying spurious structure, but also provide large and consistent gains when combined with standard regression objectives.

To support practical evaluation of spurious regression, we curate and benchmark DSR datasets for common real-world tasks in computer vision, environmental sensing, and natural language processing. They range from visual regression tasks such as age prediction, to LLM regression tasks such as code metric prediction. We further set up benchmarks for proper DSR performance evaluation. Our contributions are as follows:
\vspace{-6pt}
\begin{Itemize}
\item We formally define DSR as regression learning with spurious correlations, generalizing to the entire target range and all attribute-label combinations. DSR provides thorough and unbiased evaluation of learning algorithms in practical continuous prediction settings.
\vspace{3pt}
\item We develop two simple, effective, and interpretable algorithms for DSR, \labelmds and \featuremds, which exploit the similarity among spurious attributes in both label and feature spaces.
\vspace{3pt}
\item We curate benchmark DSR datasets in different domains: computer vision, environmental sensing, and natural language processing, covering diverse tasks from synthetic data to LLM regression. We set up strong baselines as well as benchmarks for proper DSR performance evaluation.
\vspace{3pt}
\item Extensive experiments on real-world DSR datasets verify the consistent and superior performance of our strategies. We further reveal intriguing properties of DSR on robustness and generalization.
\end{Itemize}

\section{Related Work}
\label{sec:related_work}
\vspace{-4pt}

\textbf{Spurious Correlations in Classification.}
Spurious correlations, which arise when a model relies on a feature correlated with the target during training but not causally related to it, have been widely documented in classification tasks \cite{geirhos2020shortcut}. This phenomenon, also termed \textit{shortcut learning}, is closely tied to the tendency of neural networks to exploit the simplest available signal \cite{shah2020pitfalls}.
For example, classifiers trained on biased datasets may rely on background cues \cite{beery2018recognition}, textures \cite{geirhos2019texture}, or demographic attributes \cite{buolamwini2018gender, sagawa2020dro}, rather than semantically meaningful features. Benchmarks such as SubpopBench further systematize this challenge with shifts across diverse domains \cite{yang2023change}.
However, existing works have focused exclusively on classification, where labels are discrete and groups can be naturally defined by label-attribute pairs. In contrast, we study the underexplored setting of \textit{regression}, where the target is real-valued and such discrete group structure no longer directly applies.

\textbf{Mitigating Spurious Correlation in Classification.}
Existing mitigation methods for spurious correlations broadly aim to reduce a model’s reliance on non-causal but predictive shortcuts. One line of work uses distributionally robust optimization to emphasize high-loss or underperforming groups, often assuming that group annotations are available during training \cite{duchi2021dro, sagawa2020dro}.
Another line identifies biased or misclassified samples with a preliminary model, and then upweights them to improve worst-group performance \cite{liu2021jtt, nam2020lff}. Other approaches encourage invariant representations across environments, suppress domain-discriminative information, or retrain the classifier on a more balanced subset to reduce shortcut reliance \cite{ganin2016dann, arjovsky2019irm, kirichenko2023dfr, zhang2022cnc}.
Despite this progress, these methods largely assume discrete labels or finite group-label pairs, and therefore do not directly generalize to regression settings where targets are continuous and label ordering carries semantic meaning.

\textbf{Regression Learning and Imbalanced Regression.}
Real-world prediction problems often require estimating \textit{continuous} targets from observational data, including visual, environmental, language, and physiological signals \cite{zhifei2017utkface, hu2019uav, zha2023rnc, yang2023simper}.
Deep regression models are typically trained to minimize average prediction error, which can be insufficient when target values are long-tailed or shifted across domains \cite{branco2016survey}.
Recent work on Deep Imbalanced Regression improves learning across the full target range by smoothing label and feature distributions over nearby targets \cite{yang2021delving}, while other methods reformulate regression losses, balance cross-domain transfer, or learn representations that preserve the ordinal structure of continuous labels \cite{ren2022balancedmse, yang2022mdlt, zha2023rnc}. However, these methods mainly address marginal label imbalance, domain imbalance, or target-aware representation learning, and do not explicitly model \textit{spurious} attribute correlations in regression.
Our work fills this gap by modeling geometric relationships among spurious attribute groups, directly addressing the joint $(y, a)$ imbalance and continuous spurious correlations that existing regression methods overlook.
\section{Methods}
\label{sec:methods}
\vspace{-4pt}

\noindent\textbf{Problem Setup.}
We consider supervised regression with training data $\mathcal{D} = \{(\mathbf{x}_i, y_i, a_i)\}_{i=1}^{N}$, where $\mathbf{x} \in \mathcal{X}$ is the input, $y \in \mathcal{Y} \subset \mathbb{R}$ is a continuous target, and $a \in \mathcal{A}$ is the spurious attribute.
The attribute $a$ is correlated with $y$ in the training distribution, but is unreliable at test time, inducing a shift $p_{\mathrm{train}}(y\mid a)\neq p_{\mathrm{test}}(y\mid a)$. A model trained by ERM \cite{vapnik1998statistical} may exploit this attribute-label correlation as a shortcut, encoding $a$ rather than the true causal features of $y$, and thus fail to generalize uniformly across all $(y,a)$ combinations at test time.

For density estimation and subgroup analysis, we discretize $\mathcal{Y}$ into $K$ non-overlapping bins $\{B_k\}_{k=1}^{K}$ following \cite{yang2021delving}, assigning each sample a bin index $b_i = k$ if $y_i \in B_k$. A \textit{group} is then defined as a unique combination of bin index and attribute value:
\begin{equation}
g = (k, a) \in \{1, \ldots, K\} \times \mathcal{A},
\quad n_g = \left|\{i : b_i = k,\, a_i = a\}\right|.
\label{eq:group_counts}
\end{equation}

Unlike classification, these groups are not natural task labels, but auxiliary partitions of a continuous target space. Next, we identify two structural properties of DSR that motivate our approach: continuity in the target space $\mathcal{Y}$ and similarity structure in the attribute space $\mathcal{A}$.

\begin{figure}[!t]
\centering
\includegraphics[width=.95\linewidth]{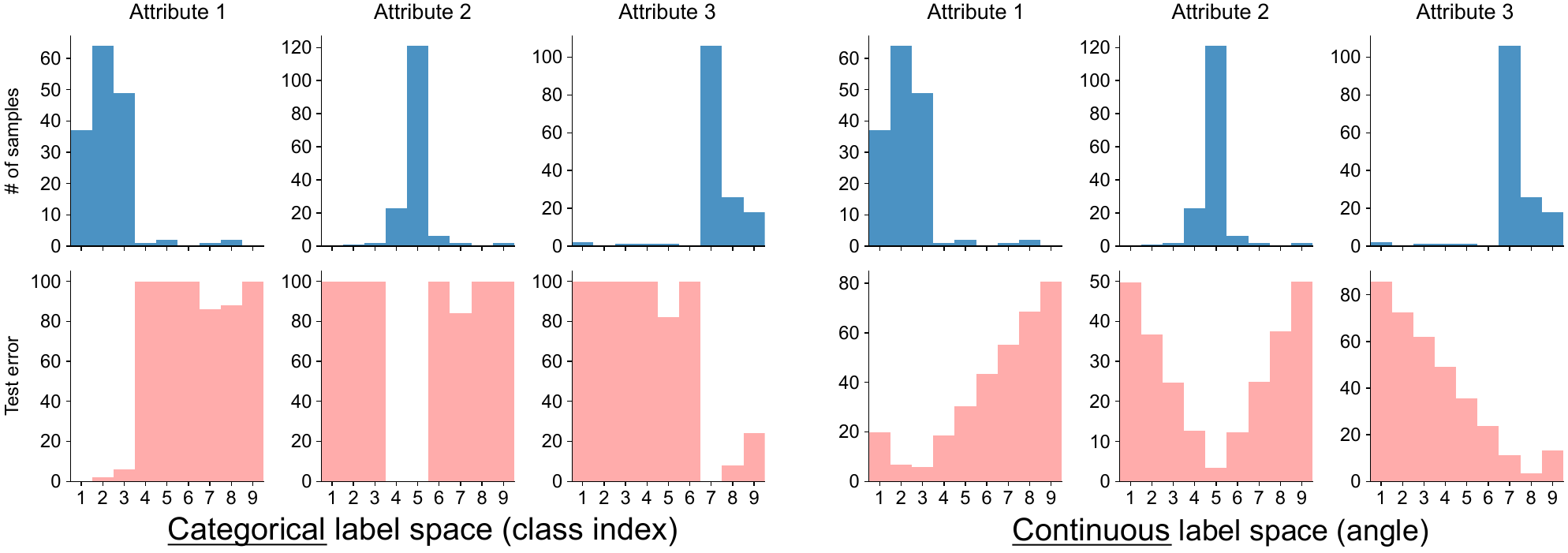}
\vspace{-3pt}
\caption{\small{
\textbf{Classification \textit{vs.} regression under the same spurious structure.} 
\textbf{Top:} We construct classification and regression tasks with identical per-attribute training distributions, with three spurious attributes and target bins. \textbf{Bottom left:} In classification, test error is nearly binary across off-diagonal class-attribute combinations. \textbf{Bottom right:} In regression, test error changes smoothly with the distance from the dominant target region. 
}}
\label{fig:reg_cls}
\vspace{-8pt}
\end{figure}

\textbf{Observation 1: Target continuity enables within-attribute smoothing.}
We first compare classification and regression under the same spurious structure using two MNIST-based tasks \cite{lecun1998mnist} with colored backgrounds as spurious attributes. In both tasks, the spurious attribute is the background color $a \in \{\text{Red}, \text{Blue}, \text{Green}\}$, and each color is strongly associated with a dominant target region in the training set.
The \textit{classification} task uses digit identity (i.e., digit ``1'' to ``9'') as the categorical label, while the \textit{regression} task uses a rotated digit ``2'' and takes the rotation angle as the continuous label.
Both tasks have identical per-attribute training distributions and balanced test sets over all target-attribute combinations.
As shown in Fig. \ref{fig:reg_cls}, the classification setting produces nearly binary off-diagonal failures: once a class falls outside the dominant region of an attribute, the model has no notion of how close or far that class is from the training support.
In contrast, the regression setting produces graded errors: prediction error changes smoothly as the angle moves away from the dominant target region of each attribute.
This reveals a key structure unique to regression: Target bins should not be treated as unrelated classes, because nearby target values carry useful information for each other within the same attribute.
We therefore exploit label continuity by smoothing along the target axis \cite{yang2021delving} inside each attribute group.

\begin{figure}[!t]
\centering
\includegraphics[width=\linewidth]{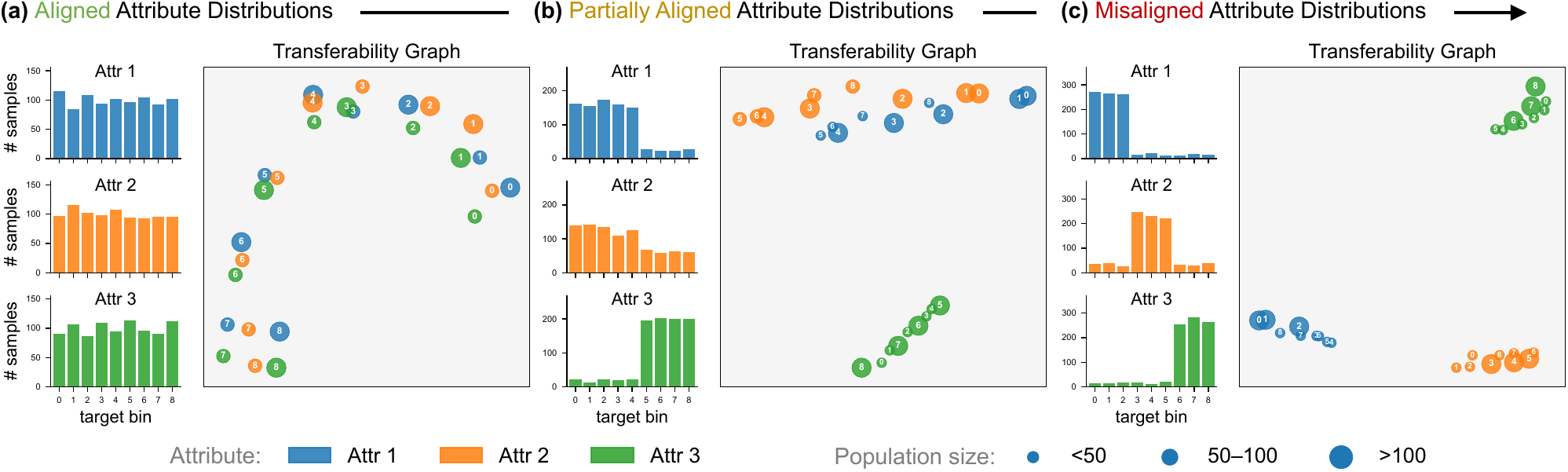}
\caption{\small{
\textbf{From aligned to misaligned spurious attributes.} We vary the similarity of target distributions across attributes: \textbf{(a)} all attributes share similar target distributions, \textbf{(b)} two attributes are similar while one differs, and \textbf{(c)} all attributes have distinct target distributions. The learned feature-space embeddings reflect these distributional relationships, motivating MDS-based information sharing across related attributes.
}}
\label{fig:trans-graph}
\vspace{-5pt}
\end{figure}

\textbf{Observation 2: Attribute similarity enables cross-attribute smoothing.}
Target continuity alone does not fully address DSR because spurious attributes may have different but related target distributions. To illustrate this, we construct three synthetic distribution scenarios with three attributes $a \in \{\text{Red}, \text{Blue}, \text{Green}\}$. In the \textbf{\textit{aligned}} setting, all three attributes have similar target distributions; in the \textbf{\textit{partially aligned}} setting, two attributes have similar distributions while the third occupies a different target range; and in the \textbf{\textit{misaligned}} setting, all attributes occupy distinct target regions.
Fig. \ref{fig:trans-graph} shows that the learned feature-space embeddings under ERM reflect these distributional relationships. When attributes share similar target distributions, their per-bin feature centroids are mixed or close to each other (Fig. \ref{fig:trans-graph}a).
When one attribute differs, its embeddings separate from the other two (Fig. \ref{fig:trans-graph}b).
When all attributes are distributionally distinct, the embeddings form separated attribute-specific clusters (Fig. \ref{fig:trans-graph}c).
This suggests that spurious attributes should not be treated as isolated groups. Instead, attributes with similar label or feature distributions can share information, especially for sparse target bins. We therefore smooth not only along the continuous target axis, but also across related attributes through an attribute affinity structure.

\textbf{Two-Dimensional Distribution Smoothing.}
Together, the observations above motivate a distribution smoothing strategy that operates along two axes. Along the \textbf{\textit{target axis}}, we exploit label continuity within each spurious attribute, so that nearby target values can support each other (i.e., training data with nearby target values can borrow statistical strength from each other). Along the \textbf{\textit{attribute axis}}, we pool information across related attributes, so that attributes with similar target or feature distributions can share statistical strength. The resulting smoothed distributions are then used to derive sample weights for training. Together, these two forms of smoothing address the joint imbalance over $(y,a)$ that is not captured by marginal label reweighting or discrete group-level debiasing alone.

\textbf{Along the Target Axis.}
For each attribute $a \in \mathcal{A}$, the target values ${y_i:a_i=a}$ may follow an uneven distribution across bins. We apply label distribution smoothing (LDS) \cite{yang2021delving} independently within each attribute, yielding a kernel-smoothed density estimate $\hat{p}_a(y)$ that captures within-attribute label frequency. Each sample receives a target-axis weight inversely proportional to this density:
\begin{equation*}
\small
w_i^{\mathrm{LDS}} \;=\; \frac{1}{\hat{p}_{a_i}(y_i)^\alpha},
\label{eq:lds}
\end{equation*}
where $\alpha$ controls the reweighting strength. This preserves the idea of LDS, but applies it conditionally on the spurious attribute, instead of estimating a single marginal density over all samples.

\textbf{Along the Attribute Axis.}
Target-axis smoothing alone treats each attribute independently. However, as shown in Fig.~\ref{fig:trans-graph}, attributes with similar target or feature distributions can provide useful information to each other, especially in sparse target regions. We construct an affinity matrix $K \in \mathbb{R}^{|\mathcal{A}| \times |\mathcal{A}|}$, where $K_{aa'}$ measures how much information attribute $a'$ contributes to attribute $a$. For each group $(b,a)$, let $n_{(b,a')}$ be defined as in Eqn. (\ref{eq:group_counts}), the smoothed count and per-sample weight are
\begin{equation*}
\small
\tilde{n}_{(b, a)} \;=\; \sum\nolimits_{a'} K_{aa'}\, n_{(b, a')},
\quad\
w_i^{\mathrm{MDS}} \;=\; \frac{1}{\tilde{n}_{(b_i,\, a_i)}^{\,\alpha}}.
\label{eq:smooth_full}
\end{equation*}
The final sample weight combines target-axis and attribute-axis smoothing, $w_i \;=\; w_i^{\mathrm{LDS}} \cdot w_i^{\mathrm{MDS}}$,
and training minimizes the weighted $\ell_1$ loss:
\begin{equation*}
\small
\mathcal{L} \;=\; \frac{1}{N}\sum\nolimits_{i=1}^{N}
w_i\,\bigl\|f_\theta(\mathbf{x}_i) - y_i\bigr\|.
\label{eq:loss}
\end{equation*}

The central question is how to define the affinity matrix $K$. We propose two instantiations following the same pipeline. Given a pairwise distance matrix $\mathbf{D}$ between attributes, we embed the attributes into a Euclidean space via Multi-Dimensional Scaling (MDS) \cite{borg2005modern}, obtaining coordinates $\mathbf{Z} =[z_m]_{m=1}^{|\mathcal{A}|} \in \mathbb{R}^{|\mathcal{A}| \times 2}$ that best preserve the pairwise distances in $\mathbf{D}$. We then apply a row-normalized RBF kernel:
\begin{equation}
\small
K_{mn} \;=\;
\frac{\exp\!\left(-\|z_m - z_n\|^2\big/2\tau^2\right)}
     {\displaystyle\sum\nolimits_{n'}\exp\!\left(-\|z_m - z_{n'}\|^2\big/2\tau^2\right)},
\label{eq:kernel}
\end{equation}
$\tau$ is the median pairwise distance in $\mathbf{Z}$. The two instantiations differ only in how $\mathbf{D}$ is computed:
\labelmds uses pairwise Wasserstein distances \cite{villani2009optimal} between per-attribute target distributions, while \featuremds uses pairwise Euclidean distances between per-attribute feature centroids.

\subsection{Label-MDS: Kernel from Target Distributions}
\label{sec:label_mds}

Label-MDS (\labelmds) defines attribute similarity through the label distributions. Intuitively, two attributes should be close if their training targets follow similar distributions, as they can provide useful count information to each other during attribute-axis smoothing. This is especially important in DSR, where related attributes may contain nearby target evidence.

\textbf{Wasserstein distance between attributes.}
Let $\hat{p}_a$ be the empirical target distribution of $a$, estimated from training. We measure pairwise dissimilarity between attributes via Wasserstein-1 distance \cite{villani2009optimal}:
\begin{equation*}
\small
    D_{aa'} \;=\; W_1\!\left(\hat{p}_a,\, \hat{p}_{a'}\right),
    \quad a,a' \in \mathcal{A},
    \label{eq:wasserstein}
\end{equation*}
forming $\mathbf{D} \in \mathbb{R}^{|\mathcal{A}| \times |\mathcal{A}|}$. We use the Wasserstein-1 distance because it respects the geometry of the continuous space \cite{peyre2019computational}: two distributions supported on nearby targets are considered close even when their supports do not exactly overlap \cite{rubner2000earth}. This property is well suited for DSR, where sparse targets can make divergence-based measures such as $\chi^2$ or Kullback--Leibler unstable or ill-defined \cite{csiszar2004information}.

\textbf{Kernel construction.}
Given $\mathbf{D}$, we apply the shared MDS pipeline: embed the attributes into $\mathbf{Z}$, form the RBF kernel $K$ as in Eqn. \eqref{eq:kernel}, and use $K$ for attribute-axis smoothing. Since $\mathbf{D}$ depends only on the training targets, \labelmds is computed once before training and adds no additional overhead.

\subsection{Feature-MDS: Kernel from Learned Representations}
\label{sec:feature_mds}
\vspace{-3pt}

\labelmds measures similarity from target distributions alone, which is fixed before training and does not reflect how the model represents different attributes. Feature-MDS (\featuremds) instead computes attribute similarity in the learned representation space, yielding a kernel that adapts to the model's evolving feature geometry. This allows attribute-axis smoothing to become progressively aligned with the internal structure learned by the encoder.

\textbf{Centroid distances in representation space.}
Let $\phi$ denote the encoder. At every $T$ epochs, we perform a forward pass over the training set and extract $\ell_2$-normalized features $h_i = \phi(\mathbf{x}_i) /
\|\phi(\mathbf{x}_i)\|_2$ for all training samples.
Normalization makes distances scale-invariant across training stages. For each attribute $a \in \mathcal{A}$, we compute the attribute centroid
\begin{equation*}
\small
c_a \;=\; \frac{1}{n_a}\sum\nolimits_{i:\,a_i=a} h_i,
\label{eq:centroid}
\end{equation*}
The pairwise distance matrix is then defined by Euclidean distances between centroids:
\begin{equation*}
\small
D_{aa'} \;=\; \|c_a - c_{a'}\|_2, \quad a,a' \in \mathcal{A}.
\label{eq:centroid_dist}
\end{equation*}

\textbf{Kernel construction.}
Given $\mathbf{D}$, we apply the same MDS pipeline as \labelmds: embed the attributes into $\mathbf{Z}$ and form the row-normalized RBF kernel $K$ using Eqn. (\ref{eq:kernel}). Attributes that are close in the encoder feature space contribute more to each other's count smoothing, while distant attributes contribute less. As a result, \featuremds adaptively transfers information across attributes according to the model's learned representation geometry, rather than relying only on fixed target distributions.
\vspace{-3pt}
\section{Benchmarking DSR}
\label{sec:results}
\vspace{-3pt}

\textbf{Datasets.}
To rigorously evaluate spurious regression across a broad range of domains and tasks, we curate and benchmark diverse DSR datasets spanning computer vision, environmental sensing, and natural language processing. 
Appendix \ref{app:dataset-details} includes full dataset details and attribute distributions. 

\vspace{-5pt}
\begin{itemize}[leftmargin=*]
\item \UTKFace \textit{($y$: age, $a$: race)}: \UTKFace is based on the UTKFace dataset \cite{zhifei2017utkface}, which contains facial images annotated with age, gender, and ethnicity. We use the five ethnicity groups as the spurious attribute, resulting in 17,620 training images, 2,753 validation images, and 3,730 test images.

\item \SkyFinder \textit{($y$: temperature, $a$: camera ID)}: \SkyFinder \cite{mihail2016skyfinder} contains pixel-annotated time-lapse sky images captured by outdoor webcams. We use camera ID as the spurious attribute and predict the in-the-wild temperature associated with each image. The dataset contains 64,945 training images, 9,335 validation images, and 6,766 test images across 47 webcams.

\item \PovertyMap \textit{($y$: poverty index, $a$: country)}: \PovertyMap is based on PovertyMap-WILDS \cite{koh2021wilds}, which contains satellite images from rural and urban regions across multiple countries. We use country as the spurious attribute and poverty index as the regression target. The training set contains 6,034 images across 20 countries, with 475 validation images and 545 test images.
 
\item \CodeNet \textit{($y$: run time, $a$: language)}: \CodeNet is based on the IBM Project CodeNet dataset \cite{puri2021project}, which contains code submissions in multiple programming languages with association metadata. We use CPU run time, clamped between 0 and 1000 $ms$, as the continuous target, and programming language as the spurious attribute. The training set includes the 13 most common languages, with 1,500 samples per language for a total of 19,500 samples. The validation and test sets each contain 6,374 samples. We use this dataset to evaluate DSR in LLM-based regression.
\end{itemize}
\vspace{-5pt}

\textbf{Network Architecture and Experiment Settings.}
For \UTKFace, \SkyFinder, and \PovertyMap, we follow the source datasets and use ResNet-18 \cite{he2016deep} as the backbone network. To evaluate \CodeNet in the context of LLM-based regression, we use RLM-GemmaS-Code-V0 \cite{akhauri2025regression}, a pre-trained encoder-decoder Regression Language Model derived from the T5-Gemma architecture \cite{zhang2025encoderdecodergemmaimprovingqualityefficiency}. We freeze the encoder and train only the decoder. Full experimental settings and details are in Appendix \ref{app:exp-settings}.

\begin{table*}[!t]
  \centering
  \small
  \caption{\small{\textbf{Main results on \UTKFace.} We report test MAE and its standard deviation across 5 random seeds.
  }}
\vspace{-3pt}
  \setlength{\tabcolsep}{3pt}
  \renewcommand{\arraystretch}{1.15}
  \resizebox{\textwidth}{!}{%
  \begin{tabular}{lccccccccccc}
    \toprule[1.5pt]
    \multirow{3}{*}{Algorithm}
      & \multirow{3}{*}{Overall} 
      & \multicolumn{2}{c}{Test Error (by attribute)}
      & \multicolumn{8}{c}{Test Error (by shot)} \\
    \cmidrule(lr){3-4} \cmidrule(lr){5-12}
      &
      & \multirow{2}{*}{Average}
      & \multirow{2}{*}{Worst}
      & \multicolumn{2}{c}{Many}
      & \multicolumn{2}{c}{Medium}
      & \multicolumn{2}{c}{Few}
      & \multicolumn{2}{c}{Zero} \\
    \cmidrule(lr){5-6} \cmidrule(lr){7-8} \cmidrule(lr){9-10} \cmidrule(lr){11-12}
      &  &  &
      & Average & Worst
      & Average & Worst
      & Average & Worst
      & Average & Worst \\
    \midrule
    
    ERM \cite{vapnik1998statistical}
    & 7.39 \scriptsize$\pm0.1$ 
    & 7.26 \scriptsize$\pm0.1$ & 9.19 \scriptsize$\pm0.2$
    & \textbf{4.34} \scriptsize$\pm0.2$ & 18.61 \scriptsize$\pm2.3$
    & 6.24 \scriptsize$\pm0.1$ & 19.18 \scriptsize$\pm0.7$
    & 7.19 \scriptsize$\pm0.1$ & 13.16 \scriptsize$\pm1.9$
    & 9.83 \scriptsize$\pm0.2$ & 73.56 \scriptsize$\pm7.3$ \\

    Resample \cite{yang2021delving}
    & 7.64 \scriptsize$\pm0.1$ 
    & 7.50 \scriptsize$\pm0.1$ & 9.40 \scriptsize$\pm0.3$
    & 4.72 \scriptsize$\pm0.1$ & 19.71 \scriptsize$\pm1.2$
    & 6.20 \scriptsize$\pm0.1$ & 17.12 \scriptsize$\pm0.9$
    & 7.00 \scriptsize$\pm0.3$ & 13.03 \scriptsize$\pm1.4$
    & 10.43 \scriptsize$\pm0.4$ & 76.58 \scriptsize$\pm7.1$ \\

    SqrtReWeight \cite{yang2021delving}
    & 7.31 \scriptsize$\pm0.1$ 
    & 7.18 \scriptsize$\pm0.1$ & 9.02 \scriptsize$\pm0.3$
    & 4.48 \scriptsize$\pm0.2$ & \textbf{16.30} \scriptsize$\pm2.3$
    & 6.10 \scriptsize$\pm0.1$ & \textbf{16.01} \scriptsize$\pm0.8$
    & 6.90 \scriptsize$\pm0.1$ & 13.18 \scriptsize$\pm1.1$
    & 9.78 \scriptsize$\pm0.4$ & \textbf{62.10} \scriptsize$\pm8.2$ \\

    ReWeight \cite{yang2021delving}
    & 8.42 \scriptsize$\pm0.1$ 
    & 8.25 \scriptsize$\pm0.1$ & 9.62 \scriptsize$\pm0.2$
    & 6.21 \scriptsize$\pm0.1$ & 17.82 \scriptsize$\pm0.9$
    & 7.06 \scriptsize$\pm0.1$ & 18.00 \scriptsize$\pm1.4$
    & 7.65 \scriptsize$\pm0.1$ & 14.78 \scriptsize$\pm1.4$
    & 10.88 \scriptsize$\pm0.3$ & 86.59 \scriptsize$\pm4.6$ \\

    CBLoss \cite{yang2021delving}
    & 8.37 \scriptsize$\pm0.1$ 
    & 8.19 \scriptsize$\pm0.1$ & 9.61 \scriptsize$\pm0.1$
    & 6.16 \scriptsize$\pm0.1$ & 18.55 \scriptsize$\pm1.8$
    & 6.98 \scriptsize$\pm0.1$ & 18.18 \scriptsize$\pm1.6$
    & 7.56 \scriptsize$\pm0.2$ & 15.23 \scriptsize$\pm1.3$
    & 10.86 \scriptsize$\pm0.2$ & 87.54 \scriptsize$\pm2.1$ \\

    DANN \cite{ganin2016dann}
    & 7.97 \scriptsize$\pm0.1$ 
    & 7.82 \scriptsize$\pm0.1$ & 9.69 \scriptsize$\pm0.2$
    & 4.63 \scriptsize$\pm0.2$ & 20.62 \scriptsize$\pm1.4$
    & 6.65 \scriptsize$\pm0.1$ & 20.32 \scriptsize$\pm0.5$
    & 7.88 \scriptsize$\pm0.1$ & 15.90 \scriptsize$\pm1.1$
    & 10.67 \scriptsize$\pm0.3$ & 76.86 \scriptsize$\pm4.7$ \\

    RnC \cite{zha2023rnc}
    & 7.38 \scriptsize$\pm0.1$ 
    & 7.25 \scriptsize$\pm0.1$ & 9.22 \scriptsize$\pm0.2$
    & \underline{4.35} \scriptsize$\pm0.1$ & 19.31 \scriptsize$\pm1.1$
    & 6.15 \scriptsize$\pm0.0$ & 17.64 \scriptsize$\pm1.3$
    & 7.12 \scriptsize$\pm0.1$ & 12.55 \scriptsize$\pm0.9$
    & 9.91 \scriptsize$\pm0.2$ & \underline{63.69} \scriptsize$\pm6.2$ \\

    LDS \cite{yang2021delving}
    & \underline{7.23} \scriptsize$\pm0.1$ 
    & \underline{7.09} \scriptsize$\pm0.1$ & \underline{8.90} \scriptsize$\pm0.1$
    & 4.58 \scriptsize$\pm0.1$ & \underline{16.59} \scriptsize$\pm1.9$
    & 6.12 \scriptsize$\pm0.0$ & \underline{16.36} \scriptsize$\pm0.6$
    & 7.01 \scriptsize$\pm0.2$ & 13.22 \scriptsize$\pm0.6$
    & \textbf{9.46} \scriptsize$\pm0.3$ & 71.49 \scriptsize$\pm10.6$ \\

    GroupDRO \cite{sagawa2020dro}
    & 7.43 \scriptsize$\pm0.1$ 
    & 7.29 \scriptsize$\pm0.1$ & 9.02 \scriptsize$\pm0.1$
    & 4.94 \scriptsize$\pm0.1$ & 16.69 \scriptsize$\pm1.7$
    & 6.13 \scriptsize$\pm0.1$ & 17.78 \scriptsize$\pm1.5$
    & 6.86 \scriptsize$\pm0.2$ & 12.71 \scriptsize$\pm0.9$
    & 9.90 \scriptsize$\pm0.1$ & 71.79 \scriptsize$\pm11.7$ \\

    \midrule
\grayrow
    \labelmds
    & 7.30 \scriptsize$\pm0.1$ 
    & 7.17 \scriptsize$\pm0.1$ & 8.94 \scriptsize$\pm0.2$
    & 4.66 \scriptsize$\pm0.2$ & 19.40 \scriptsize$\pm2.1$
    & \textbf{6.03} \scriptsize$\pm0.1$ & 17.62 \scriptsize$\pm1.0$
    & \underline{6.79} \scriptsize$\pm0.2$ & \underline{12.51} \scriptsize$\pm1.2$
    & 9.79 \scriptsize$\pm0.2$ & 76.32 \scriptsize$\pm6.2$ \\
\grayrow
    \featuremds
    & \textbf{7.22} \scriptsize$\pm0.1$ 
    & \textbf{7.08} \scriptsize$\pm0.1$ & \textbf{8.71} \scriptsize$\pm0.2$
    & 4.65 \scriptsize$\pm0.2$ & 17.49 \scriptsize$\pm1.9$
    & \underline{6.08} \scriptsize$\pm0.1$ & 17.91 \scriptsize$\pm0.2$
    & \textbf{6.71} \scriptsize$\pm0.2$ & \textbf{12.43} \scriptsize$\pm1.4$
    & \underline{9.54} \scriptsize$\pm0.4$ & 68.81 \scriptsize$\pm8.1$ \\
\grayrow
    \labelmds + \featuremds
    & 7.42 \scriptsize$\pm0.1$ 
    & 7.29 \scriptsize$\pm0.1$ & 9.04 \scriptsize$\pm0.2$
    & 4.55 \scriptsize$\pm0.2$ & 17.55 \scriptsize$\pm1.8$
    & 6.15 \scriptsize$\pm0.1$ & 17.73 \scriptsize$\pm1.8$
    & 7.02 \scriptsize$\pm0.1$ & 12.88 \scriptsize$\pm0.6$
    & 9.96 \scriptsize$\pm0.3$ & 77.43 \scriptsize$\pm9.3$ \\

\midrule
Ours (best) vs.\ ERM
& \textbf{\textcolor{green!60!black}{+0.17}}
& \textbf{\textcolor{green!60!black}{+0.18}}
& \textbf{\textcolor{green!60!black}{+0.48}}
& \textbf{\textcolor{lightblue}{-0.21}}
& \textbf{\textcolor{green!60!black}{+1.12}}
& \textbf{\textcolor{green!60!black}{+0.21}}
& \textbf{\textcolor{green!60!black}{+1.56}}
& \textbf{\textcolor{green!60!black}{+0.48}}
& \textbf{\textcolor{green!60!black}{+0.73}}
& \textbf{\textcolor{green!60!black}{+0.29}}
& \textbf{\textcolor{green!60!black}{+4.75}} \\

\bottomrule[1.5pt]
\end{tabular}%
}
\label{tab:utkface}
\vspace{-5pt}
\end{table*}
\begin{table*}[!t]
  \centering
  \small
  \caption{\small{
  \textbf{Main results on \SkyFinder.} We report test MAE and its standard deviation across 5 random seeds.
  }}
\vspace{-3pt}
  \setlength{\tabcolsep}{3pt}
  \renewcommand{\arraystretch}{1.15}
  \resizebox{\textwidth}{!}{%
  \begin{tabular}{lccccccccccc}
    \toprule[1.5pt]
    \multirow{3}{*}{Algorithm}
      & \multirow{3}{*}{Overall}
      & \multicolumn{2}{c}{Test Error (by attribute)}
      & \multicolumn{8}{c}{Test Error (by shot)} \\
    \cmidrule(lr){3-4} \cmidrule(lr){5-12}
      &
      & \multirow{2}{*}{Average}
      & \multirow{2}{*}{Worst}
      & \multicolumn{2}{c}{Many}
      & \multicolumn{2}{c}{Medium}
      & \multicolumn{2}{c}{Few}
      & \multicolumn{2}{c}{Zero} \\
    \cmidrule(lr){5-6} \cmidrule(lr){7-8} \cmidrule(lr){9-10} \cmidrule(lr){11-12}
      &  &  &
      & Average & Worst
      & Average & Worst
      & Average & Worst
      & Average & Worst \\
    \midrule
    
    ERM \cite{vapnik1998statistical}
& 3.68 \scriptsize$\pm0.0$ 
& 3.41 \scriptsize$\pm0.0$ & 5.95 \scriptsize$\pm0.1$
& \textbf{2.27} \scriptsize$\pm0.0$ & 6.45 \scriptsize$\pm0.6$
& \underline{2.94} \scriptsize$\pm0.0$ & 12.21 \scriptsize$\pm1.2$
& 4.49 \scriptsize$\pm0.0$ & 25.08 \scriptsize$\pm0.9$
& 5.22 \scriptsize$\pm0.0$ & \underline{29.78} \scriptsize$\pm1.8$ \\

Resample \cite{yang2021delving}
& 3.62 \scriptsize$\pm0.0$ 
& 3.35 \scriptsize$\pm0.0$ & \underline{5.76} \scriptsize$\pm0.1$
& 2.71 \scriptsize$\pm0.1$ & 7.28 \scriptsize$\pm0.3$
& 3.00 \scriptsize$\pm0.0$ & 13.37 \scriptsize$\pm1.2$
& 4.23 \scriptsize$\pm0.0$ & \textbf{19.23} \scriptsize$\pm0.6$
& 4.97 \scriptsize$\pm0.1$ & 36.23 \scriptsize$\pm4.0$ \\

SqrtReWeight \cite{yang2021delving}
& \underline{3.53} \scriptsize$\pm0.0$ 
& \underline{3.26} \scriptsize$\pm0.0$ & 5.85 \scriptsize$\pm0.1$
& 2.34 \scriptsize$\pm0.1$ & 6.91 \scriptsize$\pm0.8$
& 2.95 \scriptsize$\pm0.0$ & \underline{11.69} \scriptsize$\pm0.8$
& \underline{4.17} \scriptsize$\pm0.0$ & 23.76 \scriptsize$\pm1.0$
& 4.75 \scriptsize$\pm0.1$ & 32.65 \scriptsize$\pm1.7$ \\

ReWeight \cite{yang2021delving}
& 4.25 \scriptsize$\pm0.0$ 
& 3.91 \scriptsize$\pm0.0$ & 7.13 \scriptsize$\pm0.2$
& 4.03 \scriptsize$\pm0.1$ & 10.55 \scriptsize$\pm0.8$
& 3.88 \scriptsize$\pm0.0$ & 15.84 \scriptsize$\pm1.3$
& 4.52 \scriptsize$\pm0.0$ & 21.90 \scriptsize$\pm0.4$
& 5.13 \scriptsize$\pm0.1$ & 33.09 \scriptsize$\pm1.6$ \\

CBLoss \cite{yang2021delving}
& 4.23 \scriptsize$\pm0.1$ 
& 3.90 \scriptsize$\pm0.1$ & 7.24 \scriptsize$\pm0.2$
& 4.07 \scriptsize$\pm0.2$ & 10.39 \scriptsize$\pm0.8$
& 3.86 \scriptsize$\pm0.1$ & 15.04 \scriptsize$\pm2.1$
& 4.50 \scriptsize$\pm0.1$ & 20.16 \scriptsize$\pm0.6$
& 5.12 \scriptsize$\pm0.1$ & 30.88 \scriptsize$\pm1.0$ \\

DANN \cite{ganin2016dann}
& 4.04 \scriptsize$\pm0.1$ 
& 3.76 \scriptsize$\pm0.1$ & 6.75 \scriptsize$\pm0.2$
& 2.57 \scriptsize$\pm0.0$ & 7.98 \scriptsize$\pm0.7$
& 3.32 \scriptsize$\pm0.1$ & 12.60 \scriptsize$\pm1.3$
& 4.83 \scriptsize$\pm0.1$ & 24.84 \scriptsize$\pm0.4$
& 5.56 \scriptsize$\pm0.1$ & 31.02 \scriptsize$\pm1.6$ \\

RnC \cite{zha2023rnc}
& \textbf{3.49} \scriptsize$\pm0.0$ 
& \textbf{3.24} \scriptsize$\pm0.0$ & \textbf{5.69} \scriptsize$\pm0.1$
& 2.42 \scriptsize$\pm0.0$ & 7.21 \scriptsize$\pm0.5$
& \textbf{2.90} \scriptsize$\pm0.1$ & 11.99 \scriptsize$\pm0.8$
& \textbf{4.14} \scriptsize$\pm0.0$ & \underline{19.65} \scriptsize$\pm1.2$
& \textbf{4.71} \scriptsize$\pm0.1$ & 30.86 \scriptsize$\pm2.2$ \\

LDS \cite{yang2021delving}
& 3.85 \scriptsize$\pm0.1$ 
& 3.56 \scriptsize$\pm0.0$ & 6.44 \scriptsize$\pm0.4$
& 2.39 \scriptsize$\pm0.0$ & 7.95 \scriptsize$\pm0.6$
& 3.12 \scriptsize$\pm0.0$ & 13.51 \scriptsize$\pm1.0$
& 4.69 \scriptsize$\pm0.1$ & 21.71 \scriptsize$\pm0.9$
& 5.26 \scriptsize$\pm0.1$ & 33.98 \scriptsize$\pm2.8$ \\

GroupDRO \cite{sagawa2020dro}
& 3.62 \scriptsize$\pm0.0$ 
& 3.35 \scriptsize$\pm0.0$ & 6.00 \scriptsize$\pm0.1$
& 2.34 \scriptsize$\pm0.0$ & 6.64 \scriptsize$\pm0.5$
& \textbf{2.90} \scriptsize$\pm0.0$ & 12.43 \scriptsize$\pm1.1$
& 4.42 \scriptsize$\pm0.1$ & 25.07 \scriptsize$\pm1.4$
& 5.04 \scriptsize$\pm0.0$ & \textbf{29.66} \scriptsize$\pm1.5$ \\

\midrule
\grayrow
\labelmds
& 3.54 \scriptsize$\pm0.0$ 
& 3.27 \scriptsize$\pm0.0$ & 5.81 \scriptsize$\pm0.2$
& 2.38 \scriptsize$\pm0.0$ & 6.86 \scriptsize$\pm0.6$
& 2.95 \scriptsize$\pm0.0$ & \textbf{11.63} \scriptsize$\pm1.0$
& \underline{4.17} \scriptsize$\pm0.0$ & 23.63 \scriptsize$\pm0.7$
& 4.78 \scriptsize$\pm0.0$ & 31.47 \scriptsize$\pm2.3$ \\
\grayrow
\featuremds
& 3.56 \scriptsize$\pm0.0$ 
& 3.29 \scriptsize$\pm0.0$ & 5.81 \scriptsize$\pm0.2$
& \underline{2.33} \scriptsize$\pm0.1$ & \underline{6.44} \scriptsize$\pm0.3$
& 2.97 \scriptsize$\pm0.0$ & 11.86 \scriptsize$\pm0.4$
& 4.22 \scriptsize$\pm0.0$ & 21.40 \scriptsize$\pm1.1$
& \underline{4.74} \scriptsize$\pm0.0$ & 30.47 \scriptsize$\pm2.0$ \\
\grayrow
\labelmds + \featuremds
& 3.58 \scriptsize$\pm0.0$ 
& 3.30 \scriptsize$\pm0.0$ & 5.78 \scriptsize$\pm0.1$
& 2.39 \scriptsize$\pm0.1$ & \textbf{6.28} \scriptsize$\pm0.6$
& 2.97 \scriptsize$\pm0.0$ & 12.18 \scriptsize$\pm0.6$
& 4.23 \scriptsize$\pm0.1$ & 22.30 \scriptsize$\pm1.3$
& 4.82 \scriptsize$\pm0.1$ & 32.99 \scriptsize$\pm2.8$ \\

\midrule
Ours (best) vs.\ ERM
& \textbf{\textcolor{green!60!black}{+0.14}}
& \textbf{\textcolor{green!60!black}{+0.14}}
& \textbf{\textcolor{green!60!black}{+0.17}}
& \textbf{\textcolor{lightblue}{-0.06}}
& \textbf{\textcolor{green!60!black}{+0.17}}
& \textbf{\textcolor{lightblue}{-0.01}}
& \textbf{\textcolor{green!60!black}{+0.58}}
& \textbf{\textcolor{green!60!black}{+0.32}}
& \textbf{\textcolor{green!60!black}{+3.68}}
& \textbf{\textcolor{green!60!black}{+0.48}}
& \textbf{\textcolor{lightblue}{-0.69}} \\

\bottomrule[1.5pt]
  \end{tabular}%
  }
  \label{tab:skyfinder}
\vspace{-5pt}
\end{table*}

\textbf{Baselines.}
We compare \labelmds and \featuremds with standard regression and debiasing baselines. These include ERM \cite{vapnik1998statistical}, classical inverse-frequency reweighting and square-root inverse reweighting, Class-Balanced Loss \cite{cui2019classbalancedlossbasedeffective}, and LDS \cite{yang2021delving}, which smooths weights across nearby target values. To test whether classification-oriented spurious-correlation methods transfer to continuous prediction, we further evaluate DANN \cite{ganin2016dann}; for \UTKFace, \SkyFinder, and \PovertyMap, we also include GroupDRO \cite{sagawa2020dro} and RnC \cite{zha2023rnc}. This comparison covers standard ERM training, label-aware reweighting, regression imbalance methods, and representative spurious-correlation mitigation approaches.

\textbf{Evaluation Process and Metrics.}
We evaluate each regression task using mean absolute error (MAE) and error geometric mean (GM) \cite{yang2021delving}.
To evaluate performance across different target-density regimes, we follow established long-tailed evaluation protocols \cite{yang2021delving, yang2022mdlt} and divide target groups defined in \ref{eq:group_counts} into \textit{many-shot} ($>$100 training samples), \textit{medium-shot} (20$-$100 training samples), and \textit{few-shot} ($<$20 training samples) regions. For \UTKFace, \SkyFinder, and \PovertyMap, we also define \textit{zero-shot} bins as those with no training samples. We report both overall and shot-region results.

\vspace{-3pt}
\subsection{Main Results}
\vspace{-3pt}
We summarize the main results in this section for all DSR datasets and regression tasks. Additional results, training details, and hyperparameter settings can be found in Appendix~\ref{app:additional_results} and~\ref{app:exp-settings}.

\textbf{Age Regression Robust to Racial Attributes.} 
Table \ref{tab:utkface} confirms that both \labelmds and \featuremds improve substantially over the ERM, with \featuremds achieving the best overall performance among all methods. The gains are especially clear in \textit{few-shot} regions, where both methods outperform not only ERM, but also classification-oriented approaches such as DANN, GroupDRO, and RnC. Our methods provide a better balance between overall accuracy and robustness in sparse target regions.

\begin{table*}[!t]
  \centering
  \small
  \caption{\small{
  \textbf{Main results on \PovertyMap.} We report test MAE and its standard deviation across 5 random seeds.
  }}
  \vspace{-3pt}
  \setlength{\tabcolsep}{3pt}
  \renewcommand{\arraystretch}{1.15}
  \resizebox{\textwidth}{!}{%
  \begin{tabular}{lccccccccccc}
    \toprule[1.5pt]
    \multirow{3}{*}{Algorithm}
      & \multirow{3}{*}{Overall}
      & \multicolumn{2}{c}{Test Error (by attribute)}
      & \multicolumn{8}{c}{Test Error (by shot)} \\
    \cmidrule(lr){3-4} \cmidrule(lr){5-12}
      &
      & \multirow{2}{*}{Average}
      & \multirow{2}{*}{Worst}
      & \multicolumn{2}{c}{Many}
      & \multicolumn{2}{c}{Medium}
      & \multicolumn{2}{c}{Few}
      & \multicolumn{2}{c}{Zero} \\
    \cmidrule(lr){5-6} \cmidrule(lr){7-8} \cmidrule(lr){9-10} \cmidrule(lr){11-12}
      &  &  &
      & Average & Worst
      & Average & Worst
      & Average & Worst
      & Average & Worst \\
    \midrule
    
    ERM \cite{vapnik1998statistical}
    & 0.504 \scriptsize$\pm0.0$ 
    & 0.502 \scriptsize$\pm0.0$ & 0.679 \scriptsize$\pm0.0$
    & \textbf{0.256} \scriptsize$\pm0.0$ & \textbf{0.504} \scriptsize$\pm0.1$
    & 0.335 \scriptsize$\pm0.0$ & 1.356 \scriptsize$\pm0.1$
    & 0.494 \scriptsize$\pm0.0$ & 2.452 \scriptsize$\pm0.1$
    & 0.744 \scriptsize$\pm0.0$ & \underline{1.996} \scriptsize$\pm0.1$ \\

    Resample \cite{yang2021delving}
    & 0.506 \scriptsize$\pm0.0$ 
    & 0.503 \scriptsize$\pm0.0$ & 0.710 \scriptsize$\pm0.0$
    & 0.385 \scriptsize$\pm0.0$ & 0.781 \scriptsize$\pm0.2$
    & 0.391 \scriptsize$\pm0.0$ & 1.383 \scriptsize$\pm0.1$
    & \textbf{0.463} \scriptsize$\pm0.0$ & 2.247 \scriptsize$\pm0.1$
    & 0.737 \scriptsize$\pm0.0$ & 2.019 \scriptsize$\pm0.1$ \\

    SqrtReWeight \cite{yang2021delving}
    & 0.512 \scriptsize$\pm0.0$ 
    & 0.509 \scriptsize$\pm0.0$ & 0.670 \scriptsize$\pm0.0$
    & 0.375 \scriptsize$\pm0.0$ & 0.679 \scriptsize$\pm0.2$
    & 0.376 \scriptsize$\pm0.0$ & 1.441 \scriptsize$\pm0.1$
    & 0.478 \scriptsize$\pm0.0$ & 2.233 \scriptsize$\pm0.1$
    & 0.753 \scriptsize$\pm0.0$ & 2.037 \scriptsize$\pm0.1$ \\

    ReWeight \cite{yang2021delving}
    & 0.522 \scriptsize$\pm0.0$ 
    & 0.520 \scriptsize$\pm0.0$ & 0.750 \scriptsize$\pm0.0$
    & 0.485 \scriptsize$\pm0.1$ & 0.805 \scriptsize$\pm0.1$
    & 0.431 \scriptsize$\pm0.0$ & 1.426 \scriptsize$\pm0.1$
    & \underline{0.464} \scriptsize$\pm0.0$ & \underline{2.088} \scriptsize$\pm0.2$
    & 0.748 \scriptsize$\pm0.0$ & 2.012 \scriptsize$\pm0.1$ \\

    CBLoss \cite{yang2021delving}
    & 0.515 \scriptsize$\pm0.0$ 
    & 0.513 \scriptsize$\pm0.0$ & 0.720 \scriptsize$\pm0.0$
    & 0.450 \scriptsize$\pm0.0$ & 0.856 \scriptsize$\pm0.2$
    & 0.420 \scriptsize$\pm0.0$ & 1.447 \scriptsize$\pm0.1$
    & 0.467 \scriptsize$\pm0.0$ & 2.142 \scriptsize$\pm0.1$
    & 0.729 \scriptsize$\pm0.0$ & 2.029 \scriptsize$\pm0.1$ \\

    DANN \cite{ganin2016dann}
    & 0.689 \scriptsize$\pm0.1$ 
    & 0.685 \scriptsize$\pm0.1$ & 0.869 \scriptsize$\pm0.0$
    & 0.796 \scriptsize$\pm0.1$ & 0.996 \scriptsize$\pm0.1$
    & 0.574 \scriptsize$\pm0.1$ & 1.638 \scriptsize$\pm0.1$
    & 0.598 \scriptsize$\pm0.1$ & \textbf{1.926} \scriptsize$\pm0.1$
    & 1.003 \scriptsize$\pm0.1$ & 2.191 \scriptsize$\pm0.1$ \\

    RnC \cite{zha2023rnc}
    & 0.494 \scriptsize$\pm0.0$ 
    & 0.490 \scriptsize$\pm0.0$ & 0.675 \scriptsize$\pm0.0$
    & 0.304 \scriptsize$\pm0.0$ & 0.559 \scriptsize$\pm0.1$
    & \textbf{0.290} \scriptsize$\pm0.0$ & \textbf{1.103} \scriptsize$\pm0.1$
    & 0.486 \scriptsize$\pm0.0$ & 2.320 \scriptsize$\pm0.1$
    & 0.773 \scriptsize$\pm0.0$ & 2.153 \scriptsize$\pm0.2$ \\

    LDS \cite{yang2021delving}
    & 0.501 \scriptsize$\pm0.0$ 
    & 0.499 \scriptsize$\pm0.0$ & 0.712 \scriptsize$\pm0.0$
    & 0.331 \scriptsize$\pm0.0$ & 0.717 \scriptsize$\pm0.1$
    & 0.336 \scriptsize$\pm0.0$ & 1.458 \scriptsize$\pm0.1$
    & 0.501 \scriptsize$\pm0.0$ & 2.276 \scriptsize$\pm0.1$
    & \textbf{0.714} \scriptsize$\pm0.0$ & 2.049 \scriptsize$\pm0.1$ \\

    GroupDRO \cite{sagawa2020dro}
    & 0.492 \scriptsize$\pm0.0$ 
    & 0.489 \scriptsize$\pm0.0$ & \underline{0.648} \scriptsize$\pm0.0$
    & 0.376 \scriptsize$\pm0.1$ & 0.844 \scriptsize$\pm0.2$
    & \underline{0.319} \scriptsize$\pm0.0$ & \underline{1.245} \scriptsize$\pm0.1$
    & 0.470 \scriptsize$\pm0.0$ & 2.382 \scriptsize$\pm0.1$
    & 0.757 \scriptsize$\pm0.0$ & 2.016 \scriptsize$\pm0.1$ \\

    \midrule
    \grayrow
    \labelmds
    & \textbf{0.486} \scriptsize$\pm0.0$
    & \textbf{0.484} \scriptsize$\pm0.0$ & 0.666 \scriptsize$\pm0.0$
    & \underline{0.271} \scriptsize$\pm0.0$ & \underline{0.535} \scriptsize$\pm0.2$
    & 0.336 \scriptsize$\pm0.0$ & 1.417 \scriptsize$\pm0.1$
    & 0.467 \scriptsize$\pm0.0$ & 2.385 \scriptsize$\pm0.1$
    & 0.720 \scriptsize$\pm0.0$ & \textbf{1.987} \scriptsize$\pm0.1$ \\
\grayrow
    \featuremds
    & \underline{0.488} \scriptsize$\pm0.0$
    & \underline{0.485} \scriptsize$\pm0.0$ & 0.670 \scriptsize$\pm0.0$
    & 0.278 \scriptsize$\pm0.0$ & 0.554 \scriptsize$\pm0.1$
    & 0.327 \scriptsize$\pm0.0$ & 1.307 \scriptsize$\pm0.1$
    & 0.477 \scriptsize$\pm0.0$ & 2.492 \scriptsize$\pm0.1$
    & 0.719 \scriptsize$\pm0.0$ & 2.057 \scriptsize$\pm0.0$ \\
\grayrow
    \labelmds + \featuremds
    & 0.492 \scriptsize$\pm0.0$ 
    & 0.490 \scriptsize$\pm0.0$ & \textbf{0.642} \scriptsize$\pm0.0$
    & 0.352 \scriptsize$\pm0.1$ & 0.834 \scriptsize$\pm0.1$
    & 0.332 \scriptsize$\pm0.0$ & 1.369 \scriptsize$\pm0.1$
    & 0.483 \scriptsize$\pm0.0$ & 2.283 \scriptsize$\pm0.2$
    & \underline{0.715} \scriptsize$\pm0.0$ & 2.015 \scriptsize$\pm0.1$ \\

    \midrule
    Ours (best) vs.\ ERM (\%)
    & \textbf{\textcolor{green!60!black}{+3.57\%}}
    & \textbf{\textcolor{green!60!black}{+3.59\%}}
    & \textbf{\textcolor{green!60!black}{+5.45\%}}
    & \textbf{\textcolor{lightblue}{-5.86\%}}
    & \textbf{\textcolor{lightblue}{-6.15\%}}
    & \textbf{\textcolor{green!60!black}{+2.39\%}}
    & \textbf{\textcolor{green!60!black}{+3.61\%}}
    & \textbf{\textcolor{green!60!black}{+5.47\%}}
    & \textbf{\textcolor{green!60!black}{+6.89\%}}
    & \textbf{\textcolor{green!60!black}{+3.90\%}}
    & \textbf{\textcolor{green!60!black}{+0.45\%}} \\

    \bottomrule[1.5pt]
  \end{tabular}%
  }
  \label{tab:povertymap}
  \vspace{-5pt}
\end{table*}
\begin{table*}[!t]
    \centering
    \small
    \caption{\small{
    \textbf{Main results on \CodeNet.} We report test MAE and its standard deviation across 5 random seeds.
    }}
    \vspace{-3pt}
    \setlength{\tabcolsep}{3pt}
    \renewcommand{\arraystretch}{1.15}
    \resizebox{\textwidth}{!}{%
    \begin{tabular}{lccccccccccc}
      \toprule[1.5pt]
      \multirow{3}{*}{Algorithm}
        & \multirow{3}{*}{Overall}
        & \multicolumn{2}{c}{Test Error (by attribute)}
        & \multicolumn{6}{c}{Test Error (by shot)} \\
      \cmidrule(lr){3-4} \cmidrule(lr){5-12}
        &
        & \multirow{2}{*}{Average}
        & \multirow{2}{*}{Worst}
        & \multicolumn{2}{c}{Many}
        & \multicolumn{2}{c}{Medium}
        & \multicolumn{2}{c}{Few} \\
      \cmidrule(lr){5-6} \cmidrule(lr){7-8} \cmidrule(lr){9-10} \cmidrule(lr){11-12}
        &  &  &
        & Average & Worst
        & Average & Worst
        & Average & Worst \\
      \midrule
      ERM \cite{vapnik1998statistical}
        & 268.7 \scriptsize$\pm2.8$ 
        & 269.0 \scriptsize$\pm2.6$ & 350.3 \scriptsize$\pm11.0$   
        & 165.4 \scriptsize$\pm2.8$ & \textbf{228.7} \scriptsize$\pm8.9$   
        & 268.8 \scriptsize$\pm3.6$ & 398.4 \scriptsize$\pm13.1$
        & 529.8 \scriptsize$\pm5.2$ & 711.3 \scriptsize$\pm21.2$ \\
      ReWeight \cite{yang2021delving}
        & 253.7 \scriptsize$\pm2.8$ 
        & 253.5 \scriptsize$\pm2.5$ & 306.3 \scriptsize$\pm9.1$   
        & 179.4 \scriptsize$\pm3.4$ & 253.6 \scriptsize$\pm13.0$   
        & 249.4 \scriptsize$\pm3.6$ & 374.3 \scriptsize$\pm19.0$
        & 463.4 \scriptsize$\pm6.2$ & 616.5 \scriptsize$\pm18.3$ \\
      SqrtReWeight \cite{yang2021delving}
        & \underline{248.2} \scriptsize$\pm2.5$ 
        & \underline{248.3} \scriptsize$\pm2.6$ & 299.4 \scriptsize$\pm8.9$
        & 179.2 \scriptsize$\pm3.1$ & 247.2 \scriptsize$\pm11.5$   
        & 242.1 \scriptsize$\pm3.5$ & 328.2 \scriptsize$\pm12.4$
        & 444.6 \scriptsize$\pm6.3$ & \underline{609.0} \scriptsize$\pm23.5$ \\
      CBLoss \cite{yang2021delving}
        & 251.9 \scriptsize$\pm2.6$ 
        & 251.8 \scriptsize$\pm2.6$ & 301.3 \scriptsize$\pm9.6$   
        & \underline{161.3} \scriptsize$\pm2.8$ & 229.2 \scriptsize$\pm11.5$ 
        & 253.9 \scriptsize$\pm3.5$ & 333.9 \scriptsize$\pm15.2$
        & 472.9 \scriptsize$\pm5.9$ & 624.0 \scriptsize$\pm16.2$ \\
      DANN \cite{ganin2016dann}
        & 276.0 \scriptsize$\pm2.8$ 
        & 276.4 \scriptsize$\pm2.6$ & 348.8 \scriptsize$\pm11.0$   
        & \textbf{148.3} \scriptsize$\pm2.6$ & \underline{228.9} \scriptsize$\pm10.7$   
        & 292.9 \scriptsize$\pm3.4$ & 427.0 \scriptsize$\pm15.2$
        & 551.4 \scriptsize$\pm4.7$ & 716.9 \scriptsize$\pm16.3$ \\
      LDS \cite{yang2021delving}
        & 263.1 \scriptsize$\pm2.7$ 
        & 263.3 \scriptsize$\pm2.8$ & 322.4 \scriptsize$\pm9.5$   
        & 178.6 \scriptsize$\pm3.3$ & 275.6 \scriptsize$\pm13.1$   
        & 261.4 \scriptsize$\pm3.7$ & 365.5 \scriptsize$\pm12.1$
        & 484.6 \scriptsize$\pm6.4$ & 686.3 \scriptsize$\pm24.4$ \\
      \midrule
\grayrow
      \labelmds
        & \textbf{243.4} \scriptsize$\pm2.7$ 
        & \textbf{243.1} \scriptsize$\pm2.7$ & \underline{299.0} \scriptsize$\pm9.3$   
        & 163.7 \scriptsize$\pm3.3$ & 257.7 \scriptsize$\pm14.0$   
        & 245.0 \scriptsize$\pm3.9$ & \underline{321.8} \scriptsize$\pm11.4$
        & 440.2 \scriptsize$\pm6.5$ & 623.4 \scriptsize$\pm16.1$ \\
\grayrow
      \featuremds
        & 250.5 \scriptsize$\pm2.6$ 
        & 250.4 \scriptsize$\pm2.5$ & \textbf{287.2} \scriptsize$\pm8.7$  
        & 196.0 \scriptsize$\pm3.3$ & 279.6 \scriptsize$\pm10.4$   
        & \underline{235.6} \scriptsize$\pm3.5$ & \textbf{306.0} \scriptsize$\pm11.9$
        & \underline{429.0} \scriptsize$\pm2.6$ & \textbf{592.2} \scriptsize$\pm25.7$ \\
\grayrow
      \labelmds+ \featuremds
        & 249.4 \scriptsize$\pm2.5$ 
        & 249.2 \scriptsize$\pm2.6$ & 299.5 \scriptsize$\pm10.4$   
        & 205.4 \scriptsize$\pm3.4$ & 309.0 \scriptsize$\pm12.7$   
        & \textbf{231.2} \scriptsize$\pm3.3$ & \textbf{306.0} \scriptsize$\pm12.0$
        & \textbf{413.5} \scriptsize$\pm6.1$ & 622.6 \scriptsize$\pm18.2$ \\   
      \midrule 
       Ours (best) vs.\ ERM  
        & \textbf{\textcolor{green!60!black}{+25.3}}
        & \textbf{\textcolor{green!60!black}{+25.9}}
        & \textbf{\textcolor{green!60!black}{+63.1}}
        & \textbf{\textcolor{green!60!black}{+1.7}}
        & \textbf{\textcolor{lightblue}{-29.0}}
        & \textbf{\textcolor{green!60!black}{+37.6}}
        & \textbf{\textcolor{green!60!black}{+92.4}}
        & \textbf{\textcolor{green!60!black}{+116.3}}
        & \textbf{\textcolor{green!60!black}{+119.1}} \\
      \bottomrule[1.5pt]
    \end{tabular}%
    }        
    \label{tab:llm-regression}
    \vspace{-12pt}
  \end{table*}

\textbf{Temperature Regression Robust to Camera Location.}
Table~\ref{tab:skyfinder} reports results on \SkyFinder, where camera ID is as the spurious attribute. \labelmds, \featuremds, and their combination improve overall performance over ERM. The largest gains appear in the \textit{few-shot} and \textit{zero-shot} regions, where our methods reduce both average and worst-case MAE by substantial margins. Overall, our methods show better robustness to outliers, as reflected by lower worst-group MAE.

\textbf{Poverty Index Regression Robust to Country.}
Table~\ref{tab:povertymap} reports \PovertyMap, where country serves as the spurious attribute. All our variants improve over ERM overall, with \labelmds achieving the best overall performance among all compared methods. ERM shows clear signs of overfitting, whereas \labelmds and \featuremds generalize better across target regions with lower training prevalence.

\begin{wraptable}[11]{r}{0.3\textwidth}
\vspace{-12pt}
  \centering
  \small
  \setlength{\abovecaptionskip}{3pt}
  \setlength{\belowcaptionskip}{0pt}
  \setlength{\tabcolsep}{2pt}
  \renewcommand{\arraystretch}{1.15}
  \caption{\small{
  \textbf{Average performance ranking across all datasets.} Full results are in Appendix \ref{app:ranking_details}.
  }}
  
\adjustbox{max width=0.27\textwidth}{
  \begin{tabular}{clc}
    \toprule[1.5pt]
    \textbf{Rank} & \textbf{Method} & \textbf{Avg Rank} \\
    \midrule
    \grayrow
    1  & \labelmds               & 3.55 \\
    \grayrow
    2  & \featuremds             & 3.68 \\
    3  & RnC \cite{zha2023rnc} & 4.70 \\
    4  & GroupDRO \cite{sagawa2020dro} & 5.68 \\
    5  & LDS \cite{yang2021delving}                     & 6.30 \\
    6  & ERM \cite{vapnik1998statistical}                    & 6.41 \\
    7 & CBLoss \cite{cui2019classbalancedlossbasedeffective}                 & 8.24 \\
    8 & DANN \cite{ganin2016dann}                   & 9.69 \\
    \bottomrule[1.5pt]
  \end{tabular}
}
  \label{tab:average_rank_main}
\end{wraptable}

\textbf{Execution Time Regression Robust to Programming Language.}
For the LLM regression task trained on \CodeNet, we verify in Table \ref{tab:llm-regression} that \labelmds, \featuremds, and their combination consistently improve overall performance and across all shot regions. Although \featuremds shows a small degradation in the \textit{many-shot} region, it provides the strongest robustness, achieving the lowest worst-case MAE overall and in the \textit{medium-shot} and \textit{few-shot} regions. Consistent with the other datasets, our methods yield larger gains as data becomes sparser, with the most clear improvement in the \textit{few-shot} region.  

Across all datasets and metrics, the average performance ranking in Table \ref{tab:average_rank_main} further confirms the advantage of \labelmds and \featuremds.

\vspace{-3pt}
\subsection{Further Analysis}
\vspace{-3pt}

\textbf{Ablation Studies for \labelmds \& \featuremds (Appendix~\ref{app:ablation-results}).}
We study the robustness of \labelmds and \featuremds under several design choices.
\ding{182} Since our main experiments use Gaussian smoothing, we vary the kernel size $k$ $\in \{5, 9, 15\}$ and standard deviation $\sigma \in \{1, 2, 3\}$ to test sensitivity to smoothing strength.
\ding{183} We compare Gaussian smoothing with alternative kernel types to evaluate whether performance depends on a specific kernel choice.
\ding{184} We ablate the training loss function to examine whether the gains are tied to a particular regression objective. Across all settings, \labelmds and \featuremds remain robust and consistently outperform baselines. Detailed results are provided in Appendix \ref{app:broader-impacts}.

\begin{figure}[!t]
\centering
\includegraphics[width=\linewidth]{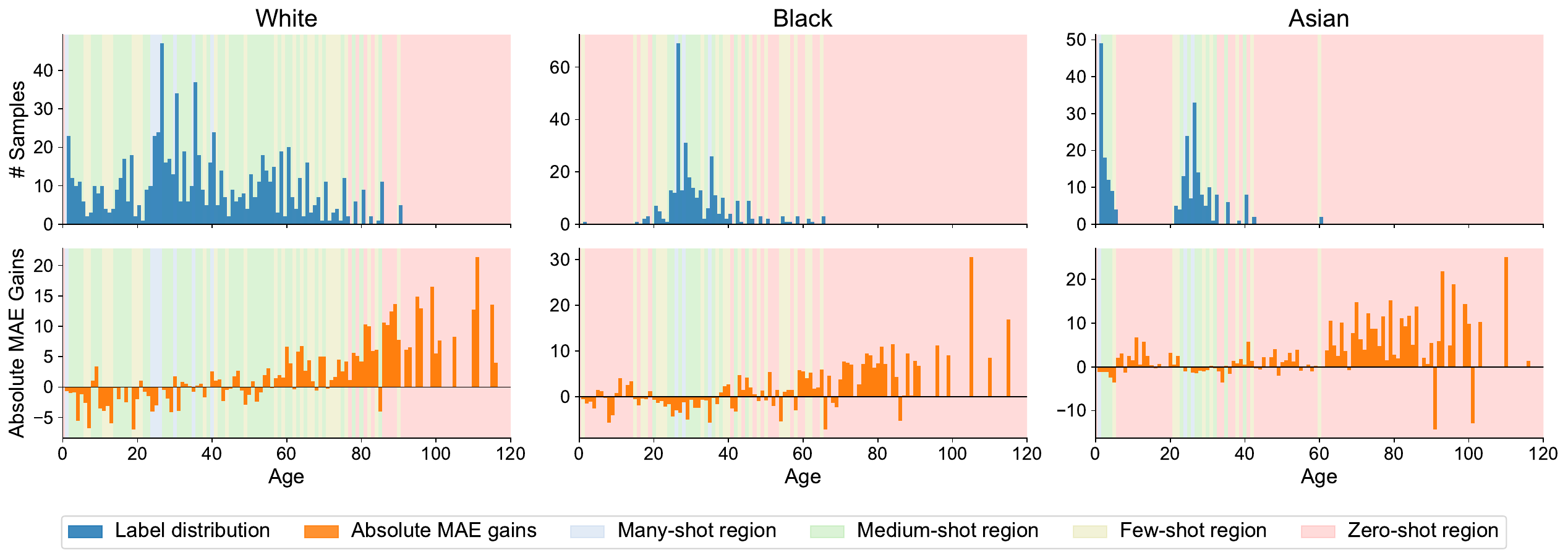}
\caption{\small{
\textbf{Interpolation and extrapolation on zero-shot target regions.} We curate a \UTKFace subset with missing age intervals and truncated age extremes. \textbf{Top:} per-attribute age distributions. \textbf{Bottom:} absolute MAE gains of \labelmds over ERM. Pink regions denote zero-shot target ranges, where \labelmds improves predictions despite no training samples. Detailed results for all attributes are provided in Appendix \ref{app:interp-extrap}.
}}
\vspace{-7pt}
\label{fig:extr-inter-dists}
\end{figure}

\textbf{Interpolation \& Extrapolation.}
Unlike classification, regression often requires predictions at target values that are unseen or missing during training. To test whether \labelmds and \featuremds can generalize to such zero-shot regions both within the training data coverage (i.e., \textit{interpolation}) and outside of it (i.e., \textit{extrapolation}), we curate a controlled subset of \UTKFace by removing selected age intervals and truncating age extremes. While our main results evaluate naturally occurring missing regions, this controlled setup allows us to isolate generalization to unseen target values. As detailed in Table \ref{tab:erm_mds_comparison}, \labelmds substantially outperforms ERM in these zero-shot regions. Fig. \ref{fig:extr-inter-dists} further visualizes the per-attribute age distributions and absolute MAE gains over ERM for three representative race groups, showing that \labelmds improves performance across both internal gaps and extrapolated age ranges. 
Detailed results for all attributes are provided in Appendix \ref{app:interp-extrap}.

\begin{wrapfigure}{r}{0.5\textwidth}
\vspace{-8pt}
\centering
\includegraphics[width=0.45\textwidth]{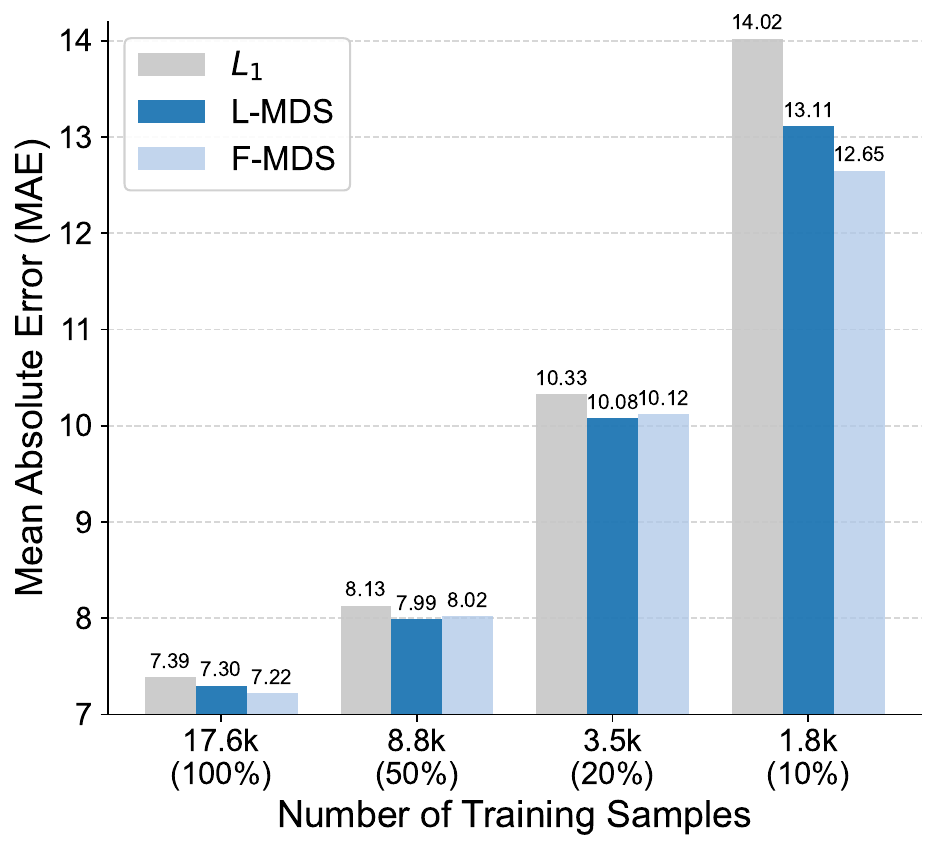}
\vspace{-6pt}
\caption{\small{
\textbf{Robustness to data scarcity.} We subsample \UTKFace to simulate limited data settings. \labelmds and \featuremds obtain larger gains under lower data regimes.
}}
\label{fig:robustness-utkface}
\vspace{-12pt}
\end{wrapfigure}

\textbf{Resilience to Reduced Training Data.}
Real-world datasets are often limited by sparse observations, annotation cost, and uneven target coverage. To test robustness under limited supervision, we subsample the \UTKFace training set to $10\%$, $20\%$, $50\%$, and $100\%$ of its original size, and compare \labelmds and \featuremds with the ERM baseline.  As shown in Fig. \ref{fig:robustness-utkface}, both \labelmds and \featuremds reduce performance degradation as the training set becomes smaller. Their gains over ERM are most clear in the low-data regimes, showing that attribute-aware smoothing is especially helpful when training coverage is sparse. Notably, \featuremds achieves the strongest improvement at $10\%$ training data, suggesting that representation-based attribute smoothing can better recover transferable structure when direct supervision is limited. Detailed shot-wise results are provided in Appendix \ref{app:reduced_shot_wise}.

\vspace{-4pt}
\section{Discussion}
\label{sec:discussion}
\vspace{-4pt}

\textbf{Limitations.}
While \labelmds and \featuremds boost overall performance, especially in \textit{few-shot} and \textit{zero-shot} regions, they may perform comparably to or slightly worse than baselines in some data-dense regions. This suggests a trade-off between preserving performance on well-represented target regions and improving robustness on sparse or unseen attribute-target combinations.
In addition, our current study focuses on settings where spurious attributes are available during training; future work may extend DSR to cases where such attributes are partially observed or unknown. 
Finally, although we evaluate across several domains, broader studies are needed to test DSR under more diverse spurious attributes, continuous targets, and deployment shifts. Further impacts are detailed in Appendix \ref{app:ablation-results}.

\textbf{Conclusion.}
We present DSR, a continuous prediction setting for studying spurious attribute-target correlations in deep regression. We introduce \labelmds and \featuremds, two MDS-based smoothing methods that exploit attribute similarity in label and feature spaces. Extensive experiments across vision, environmental sensing, and LLM-based regression show that our methods improve robustness in sparse target regions and achieve strong overall performance. Our work fills the gap in benchmarks and techniques for practical spurious correlation problems with continuous targets.


\bibliographystyle{plain}
\bibliography{ref}

\newpage
\appendix
\section{Additional Results}
\label{app:additional_results}
We report the complete evaluation results on all four datasets using error geometric mean (GM) \cite{yang2021delving}, which serves as a supplement to the provided results in the main paper. As a whole, \labelmds and \featuremds perform better than the baseline and other methods when evaluated with GM rather than MAE, indicating more uniform accuracy across attributes.

\subsection{GM results on UTKFace}
We provide GM evaluation results for \UTKFace in Table \ref{tab:utkface-gm}. As the table illustrates, our method yields consistent improvements over ERM across most GM metrics on \UTKFace with the most pronounced gains in the \textit{few-shot} and \textit{zero-shot} regions. Notably, \featuremds achieves the best overall GM and ranks among the top methods in both attribute-level and shot-based evaluations, confirming the effectiveness of our method across diverse evaluation criteria.
\begin{table*}[h]
    \centering
    \caption{\small{
Additional \UTKFace results. We report test GM and its standard deviation across 5 random seeds.
    }}            
    \setlength{\tabcolsep}{3pt}
    \renewcommand{\arraystretch}{1.15}
    \resizebox{\textwidth}{!}{%
    \begin{tabular}{lccccccccccc}
      \toprule[1.5pt]
      \multirow{3}{*}{Algorithm}
        & \multirow{3}{*}{Overall}
        & \multicolumn{2}{c}{Test Error (by attribute)}
        & \multicolumn{8}{c}{Test Error (by shot)} \\
      \cmidrule(lr){3-4} \cmidrule(lr){5-12}
        &
        & \multirow{2}{*}{Average}
        & \multirow{2}{*}{Worst}
        & \multicolumn{2}{c}{Many}
        & \multicolumn{2}{c}{Medium}
        & \multicolumn{2}{c}{Few}
        & \multicolumn{2}{c}{Zero} \\
      \cmidrule(lr){5-6} \cmidrule(lr){7-8} \cmidrule(lr){9-10} \cmidrule(lr){11-12}
        &  &  &
        & Average & Worst
        & Average & Worst
        & Average & Worst
        & Average & Worst \\
      \midrule
      ERM \cite{vapnik1998statistical}
        & 4.01 \scriptsize$\pm0.1$
        & 4.02 \scriptsize$\pm0.1$ & 5.32 \scriptsize$\pm0.1$
        & \textbf{2.18} \scriptsize$\pm0.2$ & 9.23 \scriptsize$\pm1.3$
        & 3.57 \scriptsize$\pm0.1$ & 11.64 \scriptsize$\pm0.7$
        & 4.46 \scriptsize$\pm0.1$ & 10.61 \scriptsize$\pm0.3$
        & 5.62 \scriptsize$\pm0.1$ & 73.56 \scriptsize$\pm7.2$ \\
      Resample \cite{yang2021delving}
        & 4.15 \scriptsize$\pm0.1$
        & 4.15 \scriptsize$\pm0.1$ & 5.37 \scriptsize$\pm0.3$
        & 2.39 \scriptsize$\pm0.1$ & 10.22 \scriptsize$\pm1.5$
        & 3.52 \scriptsize$\pm0.1$ & 11.62 \scriptsize$\pm0.9$
        & 4.25 \scriptsize$\pm0.3$ & 10.35 \scriptsize$\pm0.9$
        & 6.08 \scriptsize$\pm0.3$ & 76.58 \scriptsize$\pm7.1$ \\
      SqrtReWeight \cite{yang2021delving}
        & 4.01 \scriptsize$\pm0.0$
        & 4.01 \scriptsize$\pm0.0$ & 5.18 \scriptsize$\pm0.2$
        & 2.27 \scriptsize$\pm0.2$ & 8.91 \scriptsize$\pm0.4$
        & 3.47 \scriptsize$\pm0.1$ & \underline{11.12} \scriptsize$\pm1.0$
        & 4.35 \scriptsize$\pm0.1$ & 11.23 \scriptsize$\pm1.5$
        & 5.72 \scriptsize$\pm0.2$ & \textbf{62.10} \scriptsize$\pm8.2$ \\
      ReWeight \cite{yang2021delving}
        & 5.08 \scriptsize$\pm0.3$
        & 5.02 \scriptsize$\pm0.3$ & 5.91 \scriptsize$\pm0.3$
        & 3.34 \scriptsize$\pm0.3$ & 10.35 \scriptsize$\pm1.4$
        & 4.36 \scriptsize$\pm0.3$ & 14.19 \scriptsize$\pm1.4$
        & 5.23 \scriptsize$\pm0.3$ & 12.37 \scriptsize$\pm0.8$
        & 6.95 \scriptsize$\pm0.4$ & 87.63 \scriptsize$\pm3.8$ \\
      CBLoss \cite{yang2021delving}
        & 4.97 \scriptsize$\pm0.2$
        & 4.91 \scriptsize$\pm0.2$ & 5.78 \scriptsize$\pm0.3$
        & 3.29 \scriptsize$\pm0.2$ & 10.34 \scriptsize$\pm1.1$
        & 4.24 \scriptsize$\pm0.2$ & 13.52 \scriptsize$\pm1.5$
        & 5.01 \scriptsize$\pm0.3$ & 12.40 \scriptsize$\pm1.0$
        & 6.85 \scriptsize$\pm0.4$ & 87.48 \scriptsize$\pm2.4$ \\
      DANN \cite{ganin2016dann}
        & 4.34 \scriptsize$\pm0.1$
        & 4.33 \scriptsize$\pm0.1$ & 5.60 \scriptsize$\pm0.2$
        & \underline{2.25} \scriptsize$\pm0.2$ & 11.94 \scriptsize$\pm1.0$
        & 3.83 \scriptsize$\pm0.1$ & 13.13 \scriptsize$\pm1.1$
        & 5.00 \scriptsize$\pm0.1$ & 11.59 \scriptsize$\pm0.8$
        & 6.17 \scriptsize$\pm0.2$ & 76.86 \scriptsize$\pm4.7$ \\
      RnC \cite{zha2023rnc}
        & 4.07 \scriptsize$\pm0.1$
        & 4.08 \scriptsize$\pm0.1$ & 5.34 \scriptsize$\pm0.1$
        & 2.30 \scriptsize$\pm0.1$ & \underline{8.48} \scriptsize$\pm0.9$
        & 3.55 \scriptsize$\pm0.1$ & 11.72 \scriptsize$\pm1.6$
        & 4.46 \scriptsize$\pm0.1$ & 10.32 \scriptsize$\pm1.0$
        & 5.72 \scriptsize$\pm0.2$ & \underline{62.91} \scriptsize$\pm6.6$ \\
      LDS \cite{yang2021delving}
        & \underline{3.95} \scriptsize$\pm0.1$
        & \underline{3.94} \scriptsize$\pm0.1$ & \underline{5.09} \scriptsize$\pm0.1$
        & 2.35 \scriptsize$\pm0.1$ & \textbf{8.40} \scriptsize$\pm0.6$
        & 3.45 \scriptsize$\pm0.0$ & \textbf{11.04} \scriptsize$\pm0.6$
        & 4.41 \scriptsize$\pm0.2$ & 10.56 \scriptsize$\pm0.8$
        & \textbf{5.40} \scriptsize$\pm0.2$ & 71.10 \scriptsize$\pm11.3$ \\
      GroupDRO \cite{sagawa2020dro}
        & 4.08 \scriptsize$\pm0.1$
        & 4.06 \scriptsize$\pm0.1$ & 5.13 \scriptsize$\pm0.1$
        & 2.64 \scriptsize$\pm0.1$ & 9.63 \scriptsize$\pm0.6$
        & 3.46 \scriptsize$\pm0.0$ & 12.40 \scriptsize$\pm1.6$
        & 4.14 \scriptsize$\pm0.2$ & \underline{9.95} \scriptsize$\pm0.8$
        & 5.70 \scriptsize$\pm0.1$ & 71.79 \scriptsize$\pm11.7$ \\
      \midrule
      \labelmds
        & \underline{3.95} \scriptsize$\pm0.0$
        & \underline{3.94} \scriptsize$\pm0.0$ & 5.12 \scriptsize$\pm0.1$
        & 2.44 \scriptsize$\pm0.1$ & 9.94 \scriptsize$\pm1.6$
        & \textbf{3.35} \scriptsize$\pm0.1$ & 12.25 \scriptsize$\pm1.9$
        & \underline{4.10} \scriptsize$\pm0.2$ & 10.10 \scriptsize$\pm0.8$
        & 5.58 \scriptsize$\pm0.2$ & 76.32 \scriptsize$\pm6.2$ \\
      \featuremds
        & \textbf{3.91} \scriptsize$\pm0.1$
        & \textbf{3.90} \scriptsize$\pm0.1$ & \textbf{4.90} \scriptsize$\pm0.2$
        & 2.40 \scriptsize$\pm0.1$ & 8.94 \scriptsize$\pm0.7$
        & \underline{3.40} \scriptsize$\pm0.0$ & 12.56 \scriptsize$\pm0.7$
        & \textbf{4.03} \scriptsize$\pm0.2$ & \textbf{9.58} \scriptsize$\pm0.5$
        & \underline{5.43} \scriptsize$\pm0.3$ & 68.81 \scriptsize$\pm8.0$ \\
      \labelmds+ \featuremds
        & 4.06 \scriptsize$\pm0.1$
        & 4.05 \scriptsize$\pm0.0$ & \underline{5.09} \scriptsize$\pm0.1$
        & 2.31 \scriptsize$\pm0.1$ & 8.66 \scriptsize$\pm0.6$
        & 3.53 \scriptsize$\pm0.1$ & 11.63 \scriptsize$\pm1.0$
        & 4.37 \scriptsize$\pm0.1$ & 10.40 \scriptsize$\pm0.5$
        & 5.75 \scriptsize$\pm0.2$ & 77.43 \scriptsize$\pm9.3$ \\
      \midrule
      Ours (best) vs.\ ERM
        & \textbf{\textcolor{green!60!black}{+2.5\%}}
        & \textbf{\textcolor{green!60!black}{+3.0\%}}
        & \textbf{\textcolor{green!60!black}{+7.9\%}}
        & \textbf{\textcolor{lightblue}{-6.0\%}}
        & \textbf{\textcolor{green!60!black}{+6.2\%}}
        & \textbf{\textcolor{green!60!black}{+6.2\%}}
        & \textbf{\textcolor{green!60!black}{+0.1\%}}
        & \textbf{\textcolor{green!60!black}{+9.6\%}}
        & \textbf{\textcolor{green!60!black}{+9.7\%}}
        & \textbf{\textcolor{green!60!black}{+3.4\%}}
        & \textbf{\textcolor{green!60!black}{+6.5\%}} \\
      \bottomrule[1.5pt]
    \end{tabular}%
    }
    \label{tab:utkface-gm}
  \end{table*}

\subsection{GM Results on SkyFinder}
Complete GM results on \SkyFinder are shown in Table \ref{tab:skyfinder-gm}. Similar to \UTKFace, our method demonstrates stronger performance in the \textit{few-shot} and \textit{zero-shot} regions, indicating improved robustness for underrepresented target groups where standard ERM typically struggles. At the same time, the method remains competitive on the medium-shot and many-shot regions, showing that the gains do not come at the cost of performance on well-represented samples. We also observe that RnC~\cite{zha2023rnc} serves as a strong and competitive baseline on this benchmark dataset, and our method achieves comparable overall performance.
\begin{table*}[h]
    \centering
    \caption{\small{
    Additional \SkyFinder results. We report test GM and its standard deviation across 5 random seeds.
    }}      
    \setlength{\tabcolsep}{3pt}
    \renewcommand{\arraystretch}{1.15}
    \resizebox{\textwidth}{!}{%
    \begin{tabular}{lccccccccccc}
      \toprule[1.5pt]
      \multirow{3}{*}{Algorithm}
        & \multirow{3}{*}{Overall}
        & \multicolumn{2}{c}{Test Error (by attribute)}
        & \multicolumn{8}{c}{Test Error (by shot)} \\
      \cmidrule(lr){3-4} \cmidrule(lr){5-12}
        &
        & \multirow{2}{*}{Average}
        & \multirow{2}{*}{Worst}
        & \multicolumn{2}{c}{Many}
        & \multicolumn{2}{c}{Medium}
        & \multicolumn{2}{c}{Few}
        & \multicolumn{2}{c}{Zero} \\
      \cmidrule(lr){5-6} \cmidrule(lr){7-8} \cmidrule(lr){9-10} \cmidrule(lr){11-12}
        &  &  &
        & Average & Worst
        & Average & Worst
        & Average & Worst
        & Average & Worst \\
      \midrule
      ERM \cite{vapnik1998statistical}
        & 2.26 \scriptsize$\pm0.0$ 
        & 2.19 \scriptsize$\pm0.0$ & \underline{3.71} \scriptsize$\pm0.1$   
        & \textbf{1.38} \scriptsize$\pm0.0$ & \underline{6.00} \scriptsize$\pm0.8$   
        & 1.84 \scriptsize$\pm0.0$ & 10.92 \scriptsize$\pm1.2$
        & 2.86 \scriptsize$\pm0.0$ & 22.46 \scriptsize$\pm0.6$
        & 3.47 \scriptsize$\pm0.1$ & \underline{29.78} \scriptsize$\pm1.8$ \\
      Resample \cite{yang2021delving}
        & 2.23 \scriptsize$\pm0.0$ 
        & 2.14 \scriptsize$\pm0.0$ & 3.82 \scriptsize$\pm0.2$   
        & 1.69 \scriptsize$\pm0.1$ & 6.78 \scriptsize$\pm0.5$   
        & 1.85 \scriptsize$\pm0.0$ & 10.62 \scriptsize$\pm0.9$
        & 2.68 \scriptsize$\pm0.0$ & \textbf{17.47} \scriptsize$\pm0.7$
        & 3.38 \scriptsize$\pm0.1$ & 36.23 \scriptsize$\pm4.0$ \\
      SqrtReWeight \cite{yang2021delving}
        & \textbf{2.17} \scriptsize$\pm0.0$ 
        & \textbf{2.09} \scriptsize$\pm0.0$ & 3.79 \scriptsize$\pm0.2$   
        & 1.44 \scriptsize$\pm0.1$ & 6.59 \scriptsize$\pm1.0$   
        & \underline{1.83} \scriptsize$\pm0.0$ & 10.92 \scriptsize$\pm0.9$
        & \underline{2.63} \scriptsize$\pm0.1$ & 21.04 \scriptsize$\pm0.8$
        & 3.14 \scriptsize$\pm0.1$ & 32.65 \scriptsize$\pm1.7$ \\
      ReWeight \cite{yang2021delving}
        & 2.68 \scriptsize$\pm0.0$ 
        & 2.54 \scriptsize$\pm0.0$ & 4.66 \scriptsize$\pm0.2$   
        & 2.53 \scriptsize$\pm0.1$ & 10.04 \scriptsize$\pm1.0$   
        & 2.44 \scriptsize$\pm0.0$ & 14.19 \scriptsize$\pm2.1$
        & 2.83 \scriptsize$\pm0.1$ & 19.66 \scriptsize$\pm0.8$
        & 3.42 \scriptsize$\pm0.1$ & 33.09 \scriptsize$\pm1.6$ \\
      CBLoss \cite{yang2021delving}
        & 2.67 \scriptsize$\pm0.1$ 
        & 2.53 \scriptsize$\pm0.1$ & 4.76 \scriptsize$\pm0.1$   
        & 2.65 \scriptsize$\pm0.2$ & 9.77 \scriptsize$\pm0.4$   
        & 2.44 \scriptsize$\pm0.1$ & 13.81 \scriptsize$\pm1.8$
        & 2.81 \scriptsize$\pm0.1$ & \underline{17.77} \scriptsize$\pm1.2$
        & 3.33 \scriptsize$\pm0.1$ & 30.88 \scriptsize$\pm1.0$ \\
      DANN \cite{ganin2016dann}
        & 2.53 \scriptsize$\pm0.0$ 
        & 2.44 \scriptsize$\pm0.0$ & 4.55 \scriptsize$\pm0.1$   
        & 1.52 \scriptsize$\pm0.1$ & 7.40 \scriptsize$\pm0.9$   
        & 2.12 \scriptsize$\pm0.1$ & 12.09 \scriptsize$\pm1.3$
        & 3.14 \scriptsize$\pm0.0$ & 22.99 \scriptsize$\pm0.4$
        & 3.68 \scriptsize$\pm0.1$ & 31.02 \scriptsize$\pm1.6$ \\
      RnC \cite{zha2023rnc}
        & \textbf{2.17} \scriptsize$\pm0.0$ 
        & 2.10 \scriptsize$\pm0.0$ & \textbf{3.68} \scriptsize$\pm0.2$   
        & 1.56 \scriptsize$\pm0.1$ & 7.08 \scriptsize$\pm0.5$   
        & \underline{1.83} \scriptsize$\pm0.1$ & 10.50 \scriptsize$\pm0.9$
        & \textbf{2.62} \scriptsize$\pm0.0$ & 18.19 \scriptsize$\pm1.3$
        & \textbf{3.03} \scriptsize$\pm0.1$ & 30.86 \scriptsize$\pm2.2$ \\
      LDS \cite{yang2021delving}
        & 2.38 \scriptsize$\pm0.0$ 
        & 2.29 \scriptsize$\pm0.0$ & 4.00 \scriptsize$\pm0.2$   
        & 1.41 \scriptsize$\pm0.0$ & 7.63 \scriptsize$\pm0.7$   
        & 1.96 \scriptsize$\pm0.0$ & 11.36 \scriptsize$\pm0.5$
        & 3.05 \scriptsize$\pm0.1$ & 20.84 \scriptsize$\pm1.0$
        & 3.48 \scriptsize$\pm0.1$ & 33.98 \scriptsize$\pm2.8$ \\
      GroupDRO \cite{sagawa2020dro}
        & 2.22 \scriptsize$\pm0.0$ 
        & 2.14 \scriptsize$\pm0.0$ & 3.76 \scriptsize$\pm0.1$   
        & 1.42 \scriptsize$\pm0.1$ & 6.48 \scriptsize$\pm0.5$   
        & \textbf{1.79} \scriptsize$\pm0.0$ & 11.00 \scriptsize$\pm1.6$
        & 2.85 \scriptsize$\pm0.0$ & 23.21 \scriptsize$\pm1.0$
        & 3.33 \scriptsize$\pm0.0$ & \textbf{29.66} \scriptsize$\pm1.5$ \\
      \midrule
      \labelmds
        & 2.19 \scriptsize$\pm0.0$ 
        & 2.10 \scriptsize$\pm0.0$ & 3.72 \scriptsize$\pm0.2$   
        & 1.44 \scriptsize$\pm0.1$ & 6.46 \scriptsize$\pm0.5$   
        & 1.85 \scriptsize$\pm0.0$ & \textbf{10.00} \scriptsize$\pm0.9$
        & \textbf{2.62} \scriptsize$\pm0.0$ & 20.97 \scriptsize$\pm0.7$
        & 3.22 \scriptsize$\pm0.1$ & 31.47 \scriptsize$\pm2.3$ \\
      \featuremds
        & \underline{2.18} \scriptsize$\pm0.0$ 
        & \textbf{2.09} \scriptsize$\pm0.0$ & 3.75 \scriptsize$\pm0.1$   
        & \underline{1.40} \scriptsize$\pm0.0$ & 6.11 \scriptsize$\pm0.5$   
        & 1.84 \scriptsize$\pm0.0$ & \underline{10.36} \scriptsize$\pm1.2$
        & 2.70 \scriptsize$\pm0.1$ & 18.53 \scriptsize$\pm0.9$
        & \underline{3.09} \scriptsize$\pm0.0$ & 30.47 \scriptsize$\pm2.0$ \\
      \labelmds+ \featuremds
        & 2.20 \scriptsize$\pm0.0$ 
        & 2.10 \scriptsize$\pm0.0$ & \textbf{3.68} \scriptsize$\pm0.1$   
        & 1.47 \scriptsize$\pm0.1$ & \textbf{5.90} \scriptsize$\pm0.4$   
        & 1.85 \scriptsize$\pm0.0$ & 10.83 \scriptsize$\pm0.7$
        & 2.64 \scriptsize$\pm0.1$ & 20.13 \scriptsize$\pm1.4$
        & 3.20 \scriptsize$\pm0.1$ & 32.99 \scriptsize$\pm2.8$ \\   
      \midrule 
       Ours (best) vs.\ ERM  
        & \textbf{\textcolor{green!60!black}{+3.5\%}}
        & \textbf{\textcolor{green!60!black}{+4.6\%}}
        & \textbf{\textcolor{green!60!black}{+0.8\%}}
        & \textbf{\textcolor{lightblue}{-1.4\%}}
        & \textbf{\textcolor{green!60!black}{+1.7\%}}
        & \textbf{\textcolor{green!60!black}{+0.0\%}}
        & \textbf{\textcolor{green!60!black}{+8.4\%}}
        & \textbf{\textcolor{green!60!black}{+8.4\%}}
        & \textbf{\textcolor{green!60!black}{+17.5\%}}
        & \textbf{\textcolor{green!60!black}{+11.0\%}}
        & \textbf{\textcolor{lightblue}{-2.3\%}} \\
      \bottomrule[1.5pt]
    \end{tabular}%
    }        
    \label{tab:skyfinder-gm}
  \end{table*}

\subsection{GM Results on PovertyMap}
Table \ref{tab:povertymap-gm} contains all GM metrics calculated from evaluating the \PovertyMap dataset. When comparing performance with GM, \labelmds and \featuremds actually improve on the baseline across all shot regions, as opposed to a slight degradation in the \textit{many-shot} region when evaluating with MAE. The strong GM metrics for \labelmds and \featuremds suggest that both methods balance out prediction errors across the entire distribution, while the baseline instead focuses on reducing error in \textit{many-shot} regions. 
\begin{table*}[h]
    \centering
    \small
    \caption{\small{
    Additional  \PovertyMap results. We report test GM and its standard deviation across 5 random seeds.
    }}
    \setlength{\tabcolsep}{3pt}
    \renewcommand{\arraystretch}{1.15}
    \resizebox{\textwidth}{!}{%
    \begin{tabular}{lccccccccccc}
      \toprule[1.5pt]
      \multirow{3}{*}{Algorithm}
        & \multirow{3}{*}{Overall}
        & \multicolumn{2}{c}{Test Error (by attribute)}
        & \multicolumn{8}{c}{Test Error (by shot)} \\
      \cmidrule(lr){3-4} \cmidrule(lr){5-12}
        &
        & \multirow{2}{*}{Average}
        & \multirow{2}{*}{Worst}
        & \multicolumn{2}{c}{Many}
        & \multicolumn{2}{c}{Medium}
        & \multicolumn{2}{c}{Few}
        & \multicolumn{2}{c}{Zero} \\
      \cmidrule(lr){5-6} \cmidrule(lr){7-8} \cmidrule(lr){9-10} \cmidrule(lr){11-12}
        &  &  &
        & Average & Worst
        & Average & Worst
        & Average & Worst
        & Average & Worst \\
      \midrule
      ERM \cite{vapnik1998statistical}
        & \underline{0.33} \scriptsize$\pm0.0$
        & 0.34 \scriptsize$\pm0.0$ & 0.50 \scriptsize$\pm0.0$
        & 0.20 \scriptsize$\pm0.0$ & \textbf{0.50} \scriptsize$\pm0.1$
        & 0.21 \scriptsize$\pm0.0$ & 1.36 \scriptsize$\pm0.1$
        & 0.34 \scriptsize$\pm0.0$ & 2.45 \scriptsize$\pm0.1$
        & 0.59 \scriptsize$\pm0.0$ & \underline{2.00} \scriptsize$\pm0.1$ \\
      Resample \cite{yang2021delving}
        & \underline{0.33} \scriptsize$\pm0.0$
        & 0.34 \scriptsize$\pm0.0$ & 0.54 \scriptsize$\pm0.0$
        & 0.26 \scriptsize$\pm0.1$ & 0.78 \scriptsize$\pm0.2$
        & 0.25 \scriptsize$\pm0.0$ & 1.38 \scriptsize$\pm0.1$
        & \textbf{0.30} \scriptsize$\pm0.0$ & 2.25 \scriptsize$\pm0.1$
        & 0.56 \scriptsize$\pm0.0$ & 2.02 \scriptsize$\pm0.1$ \\
      SqrtReWeight \cite{yang2021delving}
        & \underline{0.33} \scriptsize$\pm0.0$
        & 0.34 \scriptsize$\pm0.0$ & 0.49 \scriptsize$\pm0.0$
        & 0.28 \scriptsize$\pm0.1$ & 0.68 \scriptsize$\pm0.2$
        & 0.23 \scriptsize$\pm0.0$ & 1.44 \scriptsize$\pm0.1$
        & \underline{0.32} \scriptsize$\pm0.0$ & 2.23 \scriptsize$\pm0.1$
        & 0.57 \scriptsize$\pm0.0$ & 2.04 \scriptsize$\pm0.1$ \\
      ReWeight \cite{yang2021delving}
        & 0.34 \scriptsize$\pm0.0$
        & 0.35 \scriptsize$\pm0.0$ & 0.58 \scriptsize$\pm0.0$
        & 0.33 \scriptsize$\pm0.1$ & 0.80 \scriptsize$\pm0.1$
        & 0.29 \scriptsize$\pm0.0$ & 1.43 \scriptsize$\pm0.1$
        & \textbf{0.30} \scriptsize$\pm0.0$ & \underline{2.09} \scriptsize$\pm0.2$
        & \textbf{0.53} \scriptsize$\pm0.0$ & 2.01 \scriptsize$\pm0.1$ \\
      CBLoss \cite{yang2021delving}
        & \underline{0.33} \scriptsize$\pm0.0$
        & 0.34 \scriptsize$\pm0.0$ & 0.55 \scriptsize$\pm0.0$
        & 0.31 \scriptsize$\pm0.0$ & 0.86 \scriptsize$\pm0.2$
        & 0.28 \scriptsize$\pm0.0$ & 1.45 \scriptsize$\pm0.1$
        & \textbf{0.30} \scriptsize$\pm0.0$ & 2.14 \scriptsize$\pm0.1$
        & \textbf{0.53} \scriptsize$\pm0.0$ & 2.03 \scriptsize$\pm0.1$ \\
      DANN \cite{ganin2016dann}
        & 0.48 \scriptsize$\pm0.1$
        & 0.48 \scriptsize$\pm0.1$ & 0.63 \scriptsize$\pm0.0$
        & 0.78 \scriptsize$\pm0.1$ & 1.00 \scriptsize$\pm0.1$
        & 0.41 \scriptsize$\pm0.1$ & 1.64 \scriptsize$\pm0.1$
        & 0.41 \scriptsize$\pm0.0$ & \textbf{1.93} \scriptsize$\pm0.1$
        & 0.76 \scriptsize$\pm0.1$ & 2.19 \scriptsize$\pm0.1$ \\
      RnC \cite{zha2023rnc}
        & \textbf{0.32} \scriptsize$\pm0.0$
        & \textbf{0.32} \scriptsize$\pm0.0$ & \underline{0.46} \scriptsize$\pm0.0$
        & 0.26 \scriptsize$\pm0.0$ & 0.56 \scriptsize$\pm0.1$
        & \textbf{0.19} \scriptsize$\pm0.0$ & \textbf{1.10} \scriptsize$\pm0.1$
        & 0.33 \scriptsize$\pm0.0$ & 2.32 \scriptsize$\pm0.1$
        & 0.61 \scriptsize$\pm0.0$ & 2.15 \scriptsize$\pm0.2$ \\
      LDS \cite{yang2021delving}
        & \underline{0.33} \scriptsize$\pm0.0$
        & \underline{0.33} \scriptsize$\pm0.0$ & 0.51 \scriptsize$\pm0.0$
        & 0.22 \scriptsize$\pm0.1$ & 0.72 \scriptsize$\pm0.1$
        & \underline{0.20} \scriptsize$\pm0.0$ & 1.46 \scriptsize$\pm0.1$
        & 0.34 \scriptsize$\pm0.0$ & 2.28 \scriptsize$\pm0.1$
        & 0.56 \scriptsize$\pm0.0$ & 2.05 \scriptsize$\pm0.1$ \\
      GroupDRO \cite{sagawa2020dro}
        & \textbf{0.32} \scriptsize$\pm0.0$
        & \underline{0.33} \scriptsize$\pm0.0$ & \textbf{0.45} \scriptsize$\pm0.0$
        & 0.26 \scriptsize$\pm0.1$ & 0.84 \scriptsize$\pm0.2$
        & \underline{0.20} \scriptsize$\pm0.0$ & \underline{1.24} \scriptsize$\pm0.1$
        & 0.33 \scriptsize$\pm0.0$ & 2.38 \scriptsize$\pm0.1$
        & 0.58 \scriptsize$\pm0.0$ & 2.02 \scriptsize$\pm0.1$ \\
      \midrule
      \labelmds
        & \textbf{0.32} \scriptsize$\pm0.0$
        & \textbf{0.32} \scriptsize$\pm0.0$ & 0.49 \scriptsize$\pm0.0$
        & \underline{0.19} \scriptsize$\pm0.1$ & \underline{0.54} \scriptsize$\pm0.2$
        & \underline{0.20} \scriptsize$\pm0.0$ & 1.42 \scriptsize$\pm0.1$
        & \underline{0.32} \scriptsize$\pm0.0$ & 2.39 \scriptsize$\pm0.1$
        & 0.57 \scriptsize$\pm0.0$ & \textbf{1.99} \scriptsize$\pm0.1$ \\
      \featuremds
        & \textbf{0.32} \scriptsize$\pm0.0$
        & \textbf{0.32} \scriptsize$\pm0.0$ & 0.48 \scriptsize$\pm0.0$
        & \textbf{0.17} \scriptsize$\pm0.0$ & 0.55 \scriptsize$\pm0.1$
        & \underline{0.20} \scriptsize$\pm0.0$ & 1.31 \scriptsize$\pm0.1$
        & 0.33 \scriptsize$\pm0.0$ & 2.49 \scriptsize$\pm0.1$
        & \underline{0.55} \scriptsize$\pm0.0$ & 2.06 \scriptsize$\pm0.0$ \\
      \labelmds+ \featuremds
        & \textbf{0.32} \scriptsize$\pm0.0$
        & \textbf{0.32} \scriptsize$\pm0.0$ & \underline{0.46} \scriptsize$\pm0.0$
        & \underline{0.19} \scriptsize$\pm0.1$ & 0.83 \scriptsize$\pm0.1$
        & \underline{0.20} \scriptsize$\pm0.0$ & 1.37 \scriptsize$\pm0.1$
        & 0.33 \scriptsize$\pm0.0$ & 2.28 \scriptsize$\pm0.2$
        & \underline{0.55} \scriptsize$\pm0.0$ & 2.02 \scriptsize$\pm0.1$ \\
      \midrule
      Ours (best) vs.\ ERM
        & \textbf{\textcolor{green!60!black}{+3.0\%}}
        & \textbf{\textcolor{green!60!black}{+5.9\%}}
        & \textbf{\textcolor{green!60!black}{+8.0\%}}
        & \textbf{\textcolor{green!60!black}{+15.0\%}}
        & \textbf{\textcolor{lightblue}{-8.0\%}}
        & \textbf{\textcolor{green!60!black}{+4.8\%}}
        & \textbf{\textcolor{green!60!black}{+3.7\%}}
        & \textbf{\textcolor{green!60!black}{+5.9\%}}
        & \textbf{\textcolor{green!60!black}{+6.9\%}}
        & \textbf{\textcolor{green!60!black}{+6.8\%}}
        & \textbf{\textcolor{green!60!black}{+0.5\%}} \\
      \bottomrule[1.5pt]
    \end{tabular}%
    }
    \label{tab:povertymap-gm}
  \end{table*}

\subsection{GM Results on CodeNet}
We report all GM results for the \CodeNet dataset in Table \ref{tab:llm-regression-gm}. We observe that GM metrics for \labelmds, \featuremds, and \labelmds and \featuremds combined further amplify the MAE results discussed in the main paper. As demonstrated with MAE, all three methods improve the baseline performance and make substantial gains in \textit{medium-shot} and \textit{few-shot} regions. The test GM results additionally demonstrate the dominance of \labelmds, where it beats all other methods and is directly comparable to the baseline, even in \textit{many-shot} regions.  
\begin{table*}[h]
    \centering
    \caption{\small{
    Additional \CodeNet results. We report test GM and its standard deviation across 5 random seeds.
    }}
    \setlength{\tabcolsep}{3pt}
    \renewcommand{\arraystretch}{1.15}
    \resizebox{\textwidth}{!}{%
    \begin{tabular}{lccccccccccc}
      \toprule[1.5pt]
      \multirow{3}{*}{Algorithm}
        & \multirow{3}{*}{Overall}
        & \multicolumn{2}{c}{Test GM (by attribute)}
        & \multicolumn{6}{c}{Test GM (by shot)} \\
      \cmidrule(lr){3-4} \cmidrule(lr){5-12}
        &
        & \multirow{2}{*}{Average}
        & \multirow{2}{*}{Worst}
        & \multicolumn{2}{c}{Many}
        & \multicolumn{2}{c}{Medium}
        & \multicolumn{2}{c}{Few} \\
      \cmidrule(lr){5-6} \cmidrule(lr){7-8} \cmidrule(lr){9-10} \cmidrule(lr){11-12}
        &  &  &
        & Average & Worst
        & Average & Worst
        & Average & Worst \\
      \midrule
      ERM \cite{vapnik1998statistical}
        & 266.4 \scriptsize$\pm2.7$
        & 142.8 \scriptsize$\pm3.3$ & 203.7 \scriptsize$\pm13.6$
        & \textbf{51.9} \scriptsize$\pm2.8$ & \textbf{92.5} \scriptsize$\pm11.0$
        & 127.1 \scriptsize$\pm3.9$ & 249.3 \scriptsize$\pm13.1$
        & 377.3 \scriptsize$\pm6.5$ & 455.6 \scriptsize$\pm13.4$ \\
      ReWeight \cite{yang2021delving}
        & 251.9 \scriptsize$\pm2.5$
        & 132.5 \scriptsize$\pm3.2$ & 172.0 \scriptsize$\pm11.3$
        & 71.5 \scriptsize$\pm4.7$ & 120.8 \scriptsize$\pm23.8$
        & 111.9 \scriptsize$\pm3.9$ & 209.3 \scriptsize$\pm13.3$
        & 279.4 \scriptsize$\pm7.9$ & 378.9 \scriptsize$\pm20.1$ \\
      SqrtReWeight \cite{yang2021delving}
        & \underline{246.6} \scriptsize$\pm2.5$
        & \underline{125.2} \scriptsize$\pm3.3$ & \underline{159.7} \scriptsize$\pm10.2$
        & 57.4 \scriptsize$\pm3.4$ & 112.9 \scriptsize$\pm10.0$
        & 113.5 \scriptsize$\pm3.5$ & 208.8 \scriptsize$\pm12.2$
        & 284.4 \scriptsize$\pm9.4$ & \textbf{371.3} \scriptsize$\pm14.7$ \\
      CBLoss \cite{yang2021delving}
        & 250.7 \scriptsize$\pm2.6$
        & 128.5 \scriptsize$\pm3.1$ & 160.7 \scriptsize$\pm9.4$
        & 64.0 \scriptsize$\pm3.3$ & 103.3 \scriptsize$\pm8.5$
        & \underline{103.0} \scriptsize$\pm3.3$ & 193.1 \scriptsize$\pm18.6$
        & 294.7 \scriptsize$\pm9.1$ & 401.9 \scriptsize$\pm14.8$ \\
      DANN \cite{ganin2016dann}
        & 273.7 \scriptsize$\pm2.6$
        & 153.5 \scriptsize$\pm3.2$ & 189.1 \scriptsize$\pm10.7$
        & 54.5 \scriptsize$\pm3.4$ & 104.6 \scriptsize$\pm25.2$
        & 138.1 \scriptsize$\pm3.7$ & 222.2 \scriptsize$\pm9.1$
        & 418.3 \scriptsize$\pm5.5$ & 531.5 \scriptsize$\pm10.0$ \\
      LDS \cite{yang2021delving}
        & 260.9 \scriptsize$\pm2.7$
        & 126.4 \scriptsize$\pm3.4$ & 175.6 \scriptsize$\pm11.1$
        & 52.9 \scriptsize$\pm3.2$ & 103.1 \scriptsize$\pm8.1$
        & 115.2 \scriptsize$\pm4.1$ & \underline{190.7} \scriptsize$\pm12.9$
        & 297.9 \scriptsize$\pm8.9$ & 427.7 \scriptsize$\pm16.3$ \\
      \midrule
      \labelmds
        & \textbf{240.8} \scriptsize$\pm2.7$
        & \textbf{109.8} \scriptsize$\pm2.9$ & \textbf{159.0} \scriptsize$\pm11.4$
        & \underline{52.0} \scriptsize$\pm2.9$ & \underline{101.9} \scriptsize$\pm8.1$
        & \textbf{94.5} \scriptsize$\pm3.4$ & \textbf{159.6} \scriptsize$\pm11.9$
        & \textbf{248.2} \scriptsize$\pm8.2$ & 375.4 \scriptsize$\pm17.0$ \\
      \featuremds
        & 249.6 \scriptsize$\pm2.5$
        & 131.4 \scriptsize$\pm3.2$ & 165.9 \scriptsize$\pm8.1$
        & 69.7 \scriptsize$\pm4.6$ & 116.7 \scriptsize$\pm25.0$
        & 111.0 \scriptsize$\pm3.5$ & 206.1 \scriptsize$\pm12.9$
        & 282.7 \scriptsize$\pm6.6$ & \underline{373.8} \scriptsize$\pm13.4$ \\
      \labelmds+\featuremds
        & 248.1 \scriptsize$\pm2.5$
        & 133.5 \scriptsize$\pm3.1$ & 178.4 \scriptsize$\pm9.0$
        & 85.5 \scriptsize$\pm5.4$ & 163.8 \scriptsize$\pm33.4$
        & 106.8 \scriptsize$\pm3.4$ & 202.0 \scriptsize$\pm13.6$
        & \underline{251.4} \scriptsize$\pm8.9$ & 377.3 \scriptsize$\pm14.7$ \\
      \midrule
       Ours (best) vs.\ ERM
        & \textbf{\textcolor{green!60!black}{+25.6}}
        & \textbf{\textcolor{green!60!black}{+33.0}}
        & \textbf{\textcolor{green!60!black}{+44.7}}
        & \textbf{\textcolor{lightblue}{-0.1}}
        & \textbf{\textcolor{lightblue}{-9.4}}
        & \textbf{\textcolor{green!60!black}{+32.6}}
        & \textbf{\textcolor{green!60!black}{+89.7}}
        & \textbf{\textcolor{green!60!black}{+129.1}}
        & \textbf{\textcolor{green!60!black}{+81.8}} \\
      \bottomrule[1.5pt]
    \end{tabular}%
    }
    \label{tab:llm-regression-gm}
  \end{table*}

\section{Further Analysis \&\ Ablation Studies}
\label{app:ablation-results}
\subsection{Hyper-parameter choices for \labelmds and \featuremds}
We explore the effects of different hyper-parameter choices on both \labelmds and \featuremds. Since we primarily use the Gaussian kernel for smoothing, we select different kernel sizes $k$ $\in \{5, 9, 15\}$ and standard deviations $\sigma \in \{1, 2, 3\}$ for \labelmds. For \featuremds, we vary the choice of kernel size $k$ $\in \{3, 5, 9\}$ and standard deviation $\sigma \in \{1, 2\}$. For \featuremds, we additionally experiment with two different reweighting methods: inverse reweight and square-root inverse reweight on our final weights.

\paragraph{\textbf{UTKFace.}} We show results for \UTKFace in Table \ref{tab:utkface_ablation}. The overall performance for different hyper-parameter choices is quite stable. Interestingly, for \UTKFace, \labelmds has better results with smaller standard deviations, while \featuremds has slightly better performance with the square-root inverse reweighting scheme.

\begin{table*}[!h]
\caption{\small{Ablation study of different hyper-parameters for \labelmds and \featuremds on \UTKFace}}
\centering
\resizebox{\textwidth}{!}{%
\begin{tabular}{c c c c cc cc cc cc cc cc}
\toprule[1.5pt]
& & & 
& \multicolumn{2}{c}{Overall}
& \multicolumn{2}{c}{Attribute}
& \multicolumn{2}{c}{Many-shot}
& \multicolumn{2}{c}{Medium-shot}
& \multicolumn{2}{c}{Few-shot}
& \multicolumn{2}{c}{Zero-shot} \\
\cmidrule(lr){5-6}
\cmidrule(lr){7-8}
\cmidrule(lr){9-10}
\cmidrule(lr){11-12}
\cmidrule(lr){13-14}
\cmidrule(lr){15-16}
Method & $k$ & $\sigma$ & RW
& MAE$\downarrow$ & GM$\downarrow$
& Avg$\downarrow$ & Worst$\downarrow$
& Avg$\downarrow$ & Worst$\downarrow$
& Avg$\downarrow$ & Worst$\downarrow$
& Avg$\downarrow$ & Worst$\downarrow$
& Avg$\downarrow$ & Worst$\downarrow$ \\

\midrule
\multicolumn{16}{l}{\textit{LDS + \labelmds variants}} \\

\multirow{9}{*}{\labelmds} 
& \multirow{3}{*}{5}  & 1 & Square-Root Inverse 
& 7.34 \scriptsize$\pm0.1$ & 3.97 \scriptsize$\pm0.0$
& 7.21 \scriptsize$\pm0.1$ & 9.03 \scriptsize$\pm0.2$
& 4.58 \scriptsize$\pm0.1$ & 17.12 \scriptsize$\pm2.3$
& 6.06 \scriptsize$\pm0.1$ & 17.85 \scriptsize$\pm1.1$
& 6.87 \scriptsize$\pm0.1$ & 13.04 \scriptsize$\pm0.5$
& 9.88 \scriptsize$\pm0.1$ & 70.04 \scriptsize$\pm13.7$ \\

& & 2 & Square-Root Inverse 
& 7.28 \scriptsize$\pm0.1$ & 3.91 \scriptsize$\pm0.1$
& 7.13 \scriptsize$\pm0.1$ & 8.90 \scriptsize$\pm0.1$
& 4.58 \scriptsize$\pm0.1$ & 19.53 \scriptsize$\pm1.5$
& 6.07 \scriptsize$\pm0.1$ & 18.87 \scriptsize$\pm0.3$
& 6.77 \scriptsize$\pm0.2$ & 13.09 \scriptsize$\pm0.6$
& 9.71 \scriptsize$\pm0.2$ & 72.36 \scriptsize$\pm3.8$ \\

& & 3 & Square-Root Inverse 
& 7.41 \scriptsize$\pm0.1$ & 4.05 \scriptsize$\pm0.1$
& 7.27 \scriptsize$\pm0.1$ & 8.96 \scriptsize$\pm0.1$
& 4.66 \scriptsize$\pm0.2$ & 18.67 \scriptsize$\pm1.9$
& 6.15 \scriptsize$\pm0.1$ & 18.15 \scriptsize$\pm0.4$
& 6.77 \scriptsize$\pm0.1$ & 12.35 \scriptsize$\pm0.4$
& 9.96 \scriptsize$\pm0.3$ & 77.14 \scriptsize$\pm9.5$ \\

\addlinespace[1pt]
\cmidrule(lr){2-16}
\addlinespace[1pt]

& \multirow{3}{*}{9}  & 1 & Square-Root Inverse 
& 7.30 \scriptsize$\pm0.1$ & 3.95 \scriptsize$\pm0.1$
& 7.17 \scriptsize$\pm0.1$ & 8.94 \scriptsize$\pm0.2$
& 4.66 \scriptsize$\pm0.2$ & 19.40 \scriptsize$\pm2.1$
& 6.03 \scriptsize$\pm0.1$ & 17.62 \scriptsize$\pm1.0$
& 6.79 \scriptsize$\pm0.2$ & 12.51 \scriptsize$\pm1.2$
& 9.79 \scriptsize$\pm0.2$ & 76.32 \scriptsize$\pm6.2$ \\

& & 2 & Square-Root Inverse 
& 7.36 \scriptsize$\pm0.2$ & 3.98 \scriptsize$\pm0.0$
& 7.22 \scriptsize$\pm0.2$ & 8.92 \scriptsize$\pm0.3$
& 4.57 \scriptsize$\pm0.1$ & 18.97 \scriptsize$\pm1.3$
& 6.13 \scriptsize$\pm0.1$ & 17.90 \scriptsize$\pm1.5$
& 6.87 \scriptsize$\pm0.1$ & 12.44 \scriptsize$\pm0.9$
& 9.86 \scriptsize$\pm0.4$ & 73.51 \scriptsize$\pm8.0$ \\

& & 3 & Square-Root Inverse 
& 7.40 \scriptsize$\pm0.1$ & 4.06 \scriptsize$\pm0.1$
& 7.26 \scriptsize$\pm0.1$ & 9.00 \scriptsize$\pm0.1$
& 4.51 \scriptsize$\pm0.2$ & 18.17 \scriptsize$\pm1.5$
& 6.17 \scriptsize$\pm0.1$ & 17.46 \scriptsize$\pm0.7$
& 6.84 \scriptsize$\pm0.2$ & 12.35 \scriptsize$\pm0.7$
& 9.97 \scriptsize$\pm0.1$ & 72.44 \scriptsize$\pm3.8$ \\

\addlinespace[1pt]
\cmidrule(lr){2-16}
\addlinespace[1pt]

& \multirow{3}{*}{15} & 1 & Square-Root Inverse 
& 7.27 \scriptsize$\pm0.1$ & 3.90 \scriptsize$\pm0.1$
& 7.14 \scriptsize$\pm0.1$ & 8.94 \scriptsize$\pm0.2$
& 4.58 \scriptsize$\pm0.1$ & 16.84 \scriptsize$\pm2.6$
& 6.09 \scriptsize$\pm0.1$ & 17.58 \scriptsize$\pm1.0$
& 6.75 \scriptsize$\pm0.1$ & 12.17 \scriptsize$\pm0.7$
& 9.69 \scriptsize$\pm0.3$ & 78.96 \scriptsize$\pm5.6$ \\

& & 2 & Square-Root Inverse 
& 7.30 \scriptsize$\pm0.1$ & 3.98 \scriptsize$\pm0.1$
& 7.16 \scriptsize$\pm0.1$ & 8.81 \scriptsize$\pm0.2$
& 4.74 \scriptsize$\pm0.2$ & 19.12 \scriptsize$\pm1.5$
& 6.14 \scriptsize$\pm0.2$ & 17.34 \scriptsize$\pm1.2$
& 6.76 \scriptsize$\pm0.1$ & 12.22 \scriptsize$\pm0.4$
& 9.66 \scriptsize$\pm0.2$ & 74.90 \scriptsize$\pm9.3$ \\

& & 3 & Square-Root Inverse 
& 7.38 \scriptsize$\pm0.1$ & 3.98 \scriptsize$\pm0.1$
& 7.24 \scriptsize$\pm0.1$ & 8.91 \scriptsize$\pm0.2$
& 4.62 \scriptsize$\pm0.1$ & 18.26 \scriptsize$\pm0.6$
& 6.16 \scriptsize$\pm0.1$ & 18.94 \scriptsize$\pm1.9$
& 6.88 \scriptsize$\pm0.2$ & 12.67 \scriptsize$\pm1.2$
& 9.86 \scriptsize$\pm0.3$ & 77.52 \scriptsize$\pm5.3$ \\

\midrule
\multicolumn{16}{l}{\textit{LDS + \featuremds variants}} \\[2pt]

\multirow{12}{*}{\featuremds}
& \multirow{4}{*}{3} & \multirow{2}{*}{1} & Inverse
& 7.36 \scriptsize$\pm0.1$ & 4.02 \scriptsize$\pm0.1$
& 7.22 \scriptsize$\pm0.1$ & 8.97 \scriptsize$\pm0.1$
& 4.90 \scriptsize$\pm0.2$ & 18.65 \scriptsize$\pm1.7$
& 6.08 \scriptsize$\pm0.1$ & 17.66 \scriptsize$\pm0.8$
& 6.58 \scriptsize$\pm0.2$ & 12.41 \scriptsize$\pm0.6$
& 9.86 \scriptsize$\pm0.2$ & 77.05 \scriptsize$\pm7.5$ \\

& & & Square-Root Inverse
& 7.40 \scriptsize$\pm0.1$ & 3.98 \scriptsize$\pm0.1$
& 7.26 \scriptsize$\pm0.1$ & 9.03 \scriptsize$\pm0.2$
& 4.47 \scriptsize$\pm0.1$ & 18.20 \scriptsize$\pm1.3$
& 6.16 \scriptsize$\pm0.1$ & 18.47 \scriptsize$\pm0.5$
& 6.90 \scriptsize$\pm0.2$ & 12.34 \scriptsize$\pm0.6$
& 9.98 \scriptsize$\pm0.2$ & 72.79 \scriptsize$\pm12.7$ \\

& & \multirow{2}{*}{2} & Inverse
& 7.46 \scriptsize$\pm0.1$ & 4.05 \scriptsize$\pm0.0$
& 7.31 \scriptsize$\pm0.1$ & 9.05 \scriptsize$\pm0.1$
& 4.99 \scriptsize$\pm0.1$ & 18.59 \scriptsize$\pm1.9$
& 6.09 \scriptsize$\pm0.1$ & 17.97 \scriptsize$\pm0.7$
& 6.65 \scriptsize$\pm0.1$ & 13.11 \scriptsize$\pm0.9$
& 10.06 \scriptsize$\pm0.1$ & 81.63 \scriptsize$\pm5.1$ \\

& & & Square-Root Inverse
& 7.31 \scriptsize$\pm0.1$ & 3.90 \scriptsize$\pm0.1$
& 7.17 \scriptsize$\pm0.1$ & 8.89 \scriptsize$\pm0.2$
& 4.58 \scriptsize$\pm0.1$ & 17.97 \scriptsize$\pm0.9$
& 6.10 \scriptsize$\pm0.0$ & 17.60 \scriptsize$\pm0.7$
& 6.76 \scriptsize$\pm0.2$ & 12.82 \scriptsize$\pm0.8$
& 9.77 \scriptsize$\pm0.2$ & 75.33 \scriptsize$\pm14.5$ \\

\cmidrule(lr){2-16}

& \multirow{4}{*}{5} & \multirow{2}{*}{1} & Inverse
& 7.45 \scriptsize$\pm0.1$ & 4.04 \scriptsize$\pm0.1$
& 7.30 \scriptsize$\pm0.1$ & 9.02 \scriptsize$\pm0.2$
& 4.97 \scriptsize$\pm0.1$ & 19.39 \scriptsize$\pm1.9$
& 6.11 \scriptsize$\pm0.1$ & 17.26 \scriptsize$\pm1.1$
& 6.69 \scriptsize$\pm0.1$ & 13.10 \scriptsize$\pm1.6$
& 10.00 \scriptsize$\pm0.2$ & 81.48 \scriptsize$\pm5.6$ \\

& & & Square-Root Inverse
& 7.43 \scriptsize$\pm0.1$ & 3.99 \scriptsize$\pm0.1$
& 7.30 \scriptsize$\pm0.1$ & 9.17 \scriptsize$\pm0.2$
& 4.57 \scriptsize$\pm0.1$ & 17.96 \scriptsize$\pm2.3$
& 6.09 \scriptsize$\pm0.1$ & 17.85 \scriptsize$\pm1.0$
& 6.96 \scriptsize$\pm0.1$ & 12.55 \scriptsize$\pm0.7$
& 10.06 \scriptsize$\pm0.3$ & 77.69 \scriptsize$\pm7.7$ \\

& & \multirow{2}{*}{2} & Inverse
& 7.46 \scriptsize$\pm0.1$ & 4.05 \scriptsize$\pm0.0$
& 7.31 \scriptsize$\pm0.1$ & 9.11 \scriptsize$\pm0.0$
& 4.92 \scriptsize$\pm0.1$ & 17.63 \scriptsize$\pm1.8$
& 6.13 \scriptsize$\pm0.1$ & 17.02 \scriptsize$\pm0.7$
& 6.55 \scriptsize$\pm0.1$ & 12.25 \scriptsize$\pm0.5$
& 10.08 \scriptsize$\pm0.1$ & 82.71 \scriptsize$\pm3.6$ \\

& & & Square-Root Inverse
& 7.32 \scriptsize$\pm0.1$ & 4.00 \scriptsize$\pm0.1$
& 7.18 \scriptsize$\pm0.1$ & 8.93 \scriptsize$\pm0.2$
& 4.55 \scriptsize$\pm0.1$ & 18.94 \scriptsize$\pm1.1$
& 6.11 \scriptsize$\pm0.1$ & 17.98 \scriptsize$\pm0.9$
& 6.80 \scriptsize$\pm0.1$ & 12.45 \scriptsize$\pm0.5$
& 9.79 \scriptsize$\pm0.2$ & 72.24 \scriptsize$\pm12.0$ \\

\cmidrule(lr){2-16}

& \multirow{4}{*}{9} & \multirow{2}{*}{1} & Inverse
& 7.42 \scriptsize$\pm0.1$ & 4.02 \scriptsize$\pm0.1$
& 7.28 \scriptsize$\pm0.1$ & 8.96 \scriptsize$\pm0.1$
& 4.97 \scriptsize$\pm0.1$ & 19.75 \scriptsize$\pm2.3$
& 6.09 \scriptsize$\pm0.1$ & 17.24 \scriptsize$\pm1.5$
& 6.67 \scriptsize$\pm0.2$ & 13.30 \scriptsize$\pm1.4$
& 9.97 \scriptsize$\pm0.2$ & 77.73 \scriptsize$\pm6.7$ \\

& & & Square-Root Inverse
& 7.22 \scriptsize$\pm0.1$ & 3.91 \scriptsize$\pm0.1$
& 7.08 \scriptsize$\pm0.1$ & 8.71 \scriptsize$\pm0.2$
& 4.65 \scriptsize$\pm0.2$ & 17.49 \scriptsize$\pm1.9$
& 6.08 \scriptsize$\pm0.1$ & 17.91 \scriptsize$\pm0.2$
& 6.71 \scriptsize$\pm0.2$ & 12.43 \scriptsize$\pm1.4$
& 9.54 \scriptsize$\pm0.4$ & 68.81 \scriptsize$\pm8.1$ \\

& & \multirow{2}{*}{2} & Inverse
& 7.36 \scriptsize$\pm0.1$ & 4.00 \scriptsize$\pm0.0$
& 7.21 \scriptsize$\pm0.1$ & 9.08 \scriptsize$\pm0.0$
& 4.88 \scriptsize$\pm0.1$ & 18.43 \scriptsize$\pm2.2$
& 6.03 \scriptsize$\pm0.0$ & 17.35 \scriptsize$\pm0.9$
& 6.59 \scriptsize$\pm0.1$ & 12.93 \scriptsize$\pm0.6$
& 9.92 \scriptsize$\pm0.2$ & 74.21 \scriptsize$\pm4.2$ \\

& & & Square-Root Inverse
& 7.39 \scriptsize$\pm0.1$ & 4.02 \scriptsize$\pm0.1$
& 7.26 \scriptsize$\pm0.1$ & 9.00 \scriptsize$\pm0.2$
& 4.67 \scriptsize$\pm0.1$ & 17.68 \scriptsize$\pm1.6$
& 6.07 \scriptsize$\pm0.1$ & 17.80 \scriptsize$\pm0.9$
& 6.85 \scriptsize$\pm0.2$ & 13.29 \scriptsize$\pm1.3$
& 9.97 \scriptsize$\pm0.2$ & 71.28 \scriptsize$\pm10.1$ \\

\bottomrule[1.5pt]
\end{tabular}}

\label{tab:utkface_ablation}
\end{table*}

\paragraph{\textbf{SkyFinder.}} We report similar results for \SkyFinder in Table \ref{tab:skyfinder_ablation}, where both \labelmds and \featuremds demonstrate consistent performance metrics regardless of kernel size $k$ or standard deviation $\sigma$. With larger kernel sizes of 9 and 15 in \labelmds, smaller standard deviations (e.g. $\sigma$ = 1) generally lead to the best results, but gains are marginal. For \featuremds, we report that square-root inverse reweighting generally outperforms inverse reweighting in \textit{many-shot} regions, while inverse reweighting yields slightly better results on \textit{few-shot} and \textit{zero-shot} regions. 

\begin{table*}[!h]
\caption{\small{Ablation study of different hyper-parameters for \labelmds and \featuremds on \SkyFinder}}
\centering
\resizebox{\textwidth}{!}{%
\begin{tabular}{c c c c cc cc cc cc cc cc}
\toprule[1.5pt]
& & &
& \multicolumn{2}{c}{Overall}
& \multicolumn{2}{c}{Attribute}
& \multicolumn{2}{c}{Many-shot}
& \multicolumn{2}{c}{Medium-shot}
& \multicolumn{2}{c}{Few-shot}
& \multicolumn{2}{c}{Zero-shot} \\
\cmidrule(lr){5-6}
\cmidrule(lr){7-8}
\cmidrule(lr){9-10}
\cmidrule(lr){11-12}
\cmidrule(lr){13-14}
\cmidrule(lr){15-16}
Method & $k$ & $\sigma$ & RW
& MAE$\downarrow$ & GM$\downarrow$
& Avg$\downarrow$ & Worst$\downarrow$
& Avg$\downarrow$ & Worst$\downarrow$
& Avg$\downarrow$ & Worst$\downarrow$
& Avg$\downarrow$ & Worst$\downarrow$
& Avg$\downarrow$ & Worst$\downarrow$ \\

\midrule

\multicolumn{16}{l}{\textit{LDS + \labelmds variants}} \\

\multirow{8}{*}{\labelmds}
& \multirow{3}{*}{5} & 1 & Square-Root Inverse
& 3.54 \scriptsize$\pm0.0$ & 2.17 \scriptsize$\pm0.0$
& 3.27 \scriptsize$\pm0.0$ & 5.94 \scriptsize$\pm0.1$
& 2.35 \scriptsize$\pm0.0$ & 6.88 \scriptsize$\pm0.5$
& 2.94 \scriptsize$\pm0.0$ & 12.40 \scriptsize$\pm1.8$
& 4.19 \scriptsize$\pm0.0$ & 23.21 \scriptsize$\pm1.2$
& 4.80 \scriptsize$\pm0.0$ & 30.94 \scriptsize$\pm1.5$ \\

& & 2 & Square-Root Inverse
& 3.57 \scriptsize$\pm0.0$ & 2.20 \scriptsize$\pm0.0$
& 3.30 \scriptsize$\pm0.0$ & 5.82 \scriptsize$\pm0.1$
& 2.39 \scriptsize$\pm0.0$ & 6.71 \scriptsize$\pm0.7$
& 2.98 \scriptsize$\pm0.0$ & 13.13 \scriptsize$\pm1.5$
& 4.24 \scriptsize$\pm0.0$ & 23.20 \scriptsize$\pm0.8$
& 4.79 \scriptsize$\pm0.1$ & 33.25 \scriptsize$\pm2.9$ \\

& & 3 & Square-Root Inverse
& 3.59 \scriptsize$\pm0.0$ & 2.22 \scriptsize$\pm0.0$
& 3.30 \scriptsize$\pm0.0$ & 5.80 \scriptsize$\pm0.1$
& 2.33 \scriptsize$\pm0.0$ & 6.14 \scriptsize$\pm0.6$
& 2.99 \scriptsize$\pm0.0$ & 11.33 \scriptsize$\pm0.3$
& 4.26 \scriptsize$\pm0.1$ & 23.38 \scriptsize$\pm0.5$
& 4.83 \scriptsize$\pm0.0$ & 28.82 \scriptsize$\pm2.7$ \\

\addlinespace[1pt]
\cmidrule(lr){2-16}
\addlinespace[1pt]

& \multirow{3}{*}{9} & 1 & Square-Root Inverse
& 3.57 \scriptsize$\pm0.0$ & 2.21 \scriptsize$\pm0.0$
& 3.30 \scriptsize$\pm0.0$ & 5.89 \scriptsize$\pm0.2$
& 2.29 \scriptsize$\pm0.0$ & 6.25 \scriptsize$\pm1.1$
& 3.00 \scriptsize$\pm0.0$ & 12.48 \scriptsize$\pm1.2$
& 4.21 \scriptsize$\pm0.1$ & 23.51 \scriptsize$\pm1.0$
& 4.82 \scriptsize$\pm0.1$ & 30.54 \scriptsize$\pm2.3$ \\

& & 2 & Square-Root Inverse
& 3.55 \scriptsize$\pm0.0$ & 2.18 \scriptsize$\pm0.0$
& 3.28 \scriptsize$\pm0.0$ & 5.79 \scriptsize$\pm0.1$
& 2.35 \scriptsize$\pm0.0$ & 6.38 \scriptsize$\pm0.8$
& 2.95 \scriptsize$\pm0.0$ & 12.09 \scriptsize$\pm1.1$
& 4.22 \scriptsize$\pm0.0$ & 23.74 \scriptsize$\pm1.5$
& 4.77 \scriptsize$\pm0.1$ & 28.15 \scriptsize$\pm3.0$ \\

& & 3 & Square-Root Inverse
& 3.54 \scriptsize$\pm0.0$ & 2.19 \scriptsize$\pm0.0$
& 3.27 \scriptsize$\pm0.0$ & 5.81 \scriptsize$\pm0.2$
& 2.38 \scriptsize$\pm0.0$ & 6.86 \scriptsize$\pm0.6$
& 2.95 \scriptsize$\pm0.0$ & 11.63 \scriptsize$\pm1.0$
& 4.17 \scriptsize$\pm0.0$ & 23.63 \scriptsize$\pm0.7$
& 4.78 \scriptsize$\pm0.0$ & 31.47 \scriptsize$\pm2.3$ \\

\addlinespace[1pt]
\cmidrule(lr){2-16}
\addlinespace[1pt]

& \multirow{2}{*}{15} & 2 & Square-Root Inverse
& 3.57 \scriptsize$\pm0.0$ & 2.19 \scriptsize$\pm0.0$
& 3.30 \scriptsize$\pm0.0$ & 5.83 \scriptsize$\pm0.1$
& 2.34 \scriptsize$\pm0.0$ & 6.75 \scriptsize$\pm0.4$
& 2.98 \scriptsize$\pm0.0$ & 13.34 \scriptsize$\pm1.1$
& 4.24 \scriptsize$\pm0.1$ & 23.34 \scriptsize$\pm1.9$
& 4.77 \scriptsize$\pm0.1$ & 33.20 \scriptsize$\pm4.4$ \\

& & 3 & Square-Root Inverse
& 3.54 \scriptsize$\pm0.0$ & 2.17 \scriptsize$\pm0.0$
& 3.26 \scriptsize$\pm0.0$ & 5.83 \scriptsize$\pm0.1$
& 2.32 \scriptsize$\pm0.1$ & 6.48 \scriptsize$\pm0.6$
& 2.95 \scriptsize$\pm0.0$ & 11.59 \scriptsize$\pm0.6$
& 4.20 \scriptsize$\pm0.0$ & 22.27 \scriptsize$\pm1.7$
& 4.74 \scriptsize$\pm0.0$ & 29.98 \scriptsize$\pm2.6$ \\

\midrule
\multicolumn{16}{l}{\textit{LDS + \featuremds variants}} \\[2pt]

\multirow{12}{*}{\featuremds}
& \multirow{4}{*}{3} & \multirow{2}{*}{1} & Inverse
& 3.62 \scriptsize$\pm0.0$ & 2.24 \scriptsize$\pm0.0$
& 3.32 \scriptsize$\pm0.0$ & 5.74 \scriptsize$\pm0.1$
& 2.68 \scriptsize$\pm0.1$ & 7.18 \scriptsize$\pm0.4$
& 3.10 \scriptsize$\pm0.0$ & 12.84 \scriptsize$\pm0.5$
& 4.13 \scriptsize$\pm0.0$ & 20.03 \scriptsize$\pm0.8$
& 4.79 \scriptsize$\pm0.0$ & 30.30 \scriptsize$\pm3.2$ \\

& & & Square-Root Inverse
& 3.56 \scriptsize$\pm0.0$ & 2.19 \scriptsize$\pm0.0$
& 3.28 \scriptsize$\pm0.0$ & 5.76 \scriptsize$\pm0.1$
& 2.41 \scriptsize$\pm0.1$ & 7.27 \scriptsize$\pm0.3$
& 2.97 \scriptsize$\pm0.0$ & 11.76 \scriptsize$\pm1.0$
& 4.20 \scriptsize$\pm0.0$ & 23.03 \scriptsize$\pm0.6$
& 4.78 \scriptsize$\pm0.0$ & 30.44 \scriptsize$\pm2.0$ \\

& & \multirow{2}{*}{2} & Inverse
& 3.63 \scriptsize$\pm0.0$ & 2.24 \scriptsize$\pm0.0$
& 3.33 \scriptsize$\pm0.0$ & 5.88 \scriptsize$\pm0.1$
& 2.62 \scriptsize$\pm0.1$ & 7.50 \scriptsize$\pm0.2$
& 3.10 \scriptsize$\pm0.0$ & 13.07 \scriptsize$\pm1.1$
& 4.17 \scriptsize$\pm0.1$ & 21.27 \scriptsize$\pm1.4$
& 4.76 \scriptsize$\pm0.0$ & 28.81 \scriptsize$\pm1.6$ \\

& & & Square-Root Inverse
& 3.59 \scriptsize$\pm0.0$ & 2.20 \scriptsize$\pm0.0$
& 3.30 \scriptsize$\pm0.0$ & 5.80 \scriptsize$\pm0.0$
& 2.42 \scriptsize$\pm0.1$ & 6.79 \scriptsize$\pm1.0$
& 3.00 \scriptsize$\pm0.0$ & 12.91 \scriptsize$\pm1.0$
& 4.23 \scriptsize$\pm0.1$ & 23.46 \scriptsize$\pm0.7$
& 4.83 \scriptsize$\pm0.1$ & 29.68 \scriptsize$\pm1.9$ \\

\cmidrule(lr){2-16}

& \multirow{4}{*}{5} & \multirow{2}{*}{1} & Inverse
& 3.57 \scriptsize$\pm0.0$ & 2.21 \scriptsize$\pm0.0$
& 3.28 \scriptsize$\pm0.0$ & 5.79 \scriptsize$\pm0.1$
& 2.55 \scriptsize$\pm0.0$ & 6.65 \scriptsize$\pm0.5$
& 3.05 \scriptsize$\pm0.0$ & 12.73 \scriptsize$\pm1.0$
& 4.12 \scriptsize$\pm0.1$ & 21.91 \scriptsize$\pm1.3$
& 4.72 \scriptsize$\pm0.0$ & 29.46 \scriptsize$\pm2.8$ \\

& & & Square-Root Inverse
& 3.57 \scriptsize$\pm0.0$ & 2.19 \scriptsize$\pm0.0$
& 3.29 \scriptsize$\pm0.0$ & 5.83 \scriptsize$\pm0.1$
& 2.31 \scriptsize$\pm0.1$ & 6.41 \scriptsize$\pm0.5$
& 2.98 \scriptsize$\pm0.0$ & 12.18 \scriptsize$\pm1.3$
& 4.23 \scriptsize$\pm0.0$ & 22.98 \scriptsize$\pm0.9$
& 4.80 \scriptsize$\pm0.0$ & 31.36 \scriptsize$\pm1.6$ \\

& & \multirow{2}{*}{2} & Inverse
& 3.58 \scriptsize$\pm0.0$ & 2.21 \scriptsize$\pm0.0$
& 3.29 \scriptsize$\pm0.0$ & 5.76 \scriptsize$\pm0.2$
& 2.51 \scriptsize$\pm0.0$ & 6.87 \scriptsize$\pm0.5$
& 3.04 \scriptsize$\pm0.0$ & 11.72 \scriptsize$\pm0.8$
& 4.13 \scriptsize$\pm0.1$ & 20.41 \scriptsize$\pm1.3$
& 4.78 \scriptsize$\pm0.1$ & 31.76 \scriptsize$\pm2.7$ \\

& & & Square-Root Inverse
& 3.56 \scriptsize$\pm0.0$ & 2.18 \scriptsize$\pm0.0$
& 3.28 \scriptsize$\pm0.0$ & 5.88 \scriptsize$\pm0.2$
& 2.30 \scriptsize$\pm0.0$ & 6.17 \scriptsize$\pm0.6$
& 2.95 \scriptsize$\pm0.0$ & 12.58 \scriptsize$\pm0.2$
& 4.25 \scriptsize$\pm0.0$ & 22.52 \scriptsize$\pm0.7$
& 4.80 \scriptsize$\pm0.1$ & 32.55 \scriptsize$\pm2.8$ \\

\cmidrule(lr){2-16}

& \multirow{4}{*}{9} & \multirow{2}{*}{1} & Inverse
& 3.59 \scriptsize$\pm0.0$ & 2.22 \scriptsize$\pm0.0$
& 3.30 \scriptsize$\pm0.0$ & 5.84 \scriptsize$\pm0.1$
& 2.56 \scriptsize$\pm0.1$ & 6.89 \scriptsize$\pm0.7$
& 3.07 \scriptsize$\pm0.0$ & 12.67 \scriptsize$\pm0.6$
& 4.12 \scriptsize$\pm0.0$ & 19.94 \scriptsize$\pm2.2$
& 4.72 \scriptsize$\pm0.1$ & 29.46 \scriptsize$\pm1.8$ \\

& & & Square-Root Inverse
& 3.58 \scriptsize$\pm0.0$ & 2.20 \scriptsize$\pm0.0$
& 3.30 \scriptsize$\pm0.0$ & 5.97 \scriptsize$\pm0.1$
& 2.33 \scriptsize$\pm0.1$ & 6.48 \scriptsize$\pm0.5$
& 2.97 \scriptsize$\pm0.0$ & 12.45 \scriptsize$\pm1.9$
& 4.27 \scriptsize$\pm0.0$ & 23.02 \scriptsize$\pm0.5$
& 4.79 \scriptsize$\pm0.1$ & 30.35 \scriptsize$\pm1.4$ \\

& & \multirow{2}{*}{2} & Inverse
& 3.58 \scriptsize$\pm0.0$ & 2.20 \scriptsize$\pm0.0$
& 3.29 \scriptsize$\pm0.0$ & 5.76 \scriptsize$\pm0.1$
& 2.52 \scriptsize$\pm0.1$ & 6.98 \scriptsize$\pm0.6$
& 3.04 \scriptsize$\pm0.0$ & 13.62 \scriptsize$\pm0.6$
& 4.13 \scriptsize$\pm0.1$ & 21.86 \scriptsize$\pm1.3$
& 4.76 \scriptsize$\pm0.0$ & 28.52 \scriptsize$\pm1.2$ \\

& & & Square-Root Inverse
& 3.56 \scriptsize$\pm0.0$ & 2.18 \scriptsize$\pm0.0$
& 3.29 \scriptsize$\pm0.0$ & 5.81 \scriptsize$\pm0.2$
& 2.33 \scriptsize$\pm0.1$ & 6.44 \scriptsize$\pm0.3$
& 2.97 \scriptsize$\pm0.0$ & 11.86 \scriptsize$\pm0.4$
& 4.22 \scriptsize$\pm0.0$ & 21.40 \scriptsize$\pm1.1$
& 4.74 \scriptsize$\pm0.0$ & 30.47 \scriptsize$\pm2.0$ \\

\bottomrule[1.5pt]
\end{tabular}}
\label{tab:skyfinder_ablation}
\end{table*}

\paragraph{\textbf{PovertyMap.}} Again, minimal fluctuations in MAE and GM despite varying hyper-parameter choices reflect the robustness of \labelmds and \featuremds on the \PovertyMap dataset. We report no consistent trend on kernel size and standard deviation for \labelmds - for instance, $k$ = 15 and $\sigma$ = 1 has the lowest overall MAE and GM, while $k$ = 9 and $\sigma$ = 1 performs the worse, and $k$ = 5 and $\sigma$ = 1 is in-between the two. Reweighting methods balance out overall for \featuremds, as square-root inverse reweighting performs better in \textit{many-shot} regions overall, while inverse reweighting does better on more data-scarce regions. 
\begin{table*}[!h]
\centering
\caption{\small{Ablation study of different hyper-parameters for \labelmds and \featuremds on \PovertyMap}}
\setlength{\tabcolsep}{3pt}
    \renewcommand{\arraystretch}{1.15}
    \resizebox{\textwidth}{!}{%
\begin{tabular}{c c c c cc cc cc cc cc cc}
\toprule[1.5pt]
& & &
& \multicolumn{2}{c}{Overall}
& \multicolumn{2}{c}{Attribute}
& \multicolumn{2}{c}{Many-shot}
& \multicolumn{2}{c}{Medium-shot}
& \multicolumn{2}{c}{Few-shot}
& \multicolumn{2}{c}{Zero-shot} \\
\cmidrule(lr){5-6}
\cmidrule(lr){7-8}
\cmidrule(lr){9-10}
\cmidrule(lr){11-12}
\cmidrule(lr){13-14}
\cmidrule(lr){15-16}
Method & $k$ & $\sigma$ & RW
& MAE$\downarrow$ & GM$\downarrow$
& Avg$\downarrow$ & Worst$\downarrow$
& Avg$\downarrow$ & Worst$\downarrow$
& Avg$\downarrow$ & Worst$\downarrow$
& Avg$\downarrow$ & Worst$\downarrow$
& Avg$\downarrow$ & Worst$\downarrow$ \\

\midrule
\multicolumn{16}{l}{\textit{LDS + \labelmds variants}} \\

\multirow{9}{*}{\labelmds}
& \multirow{3}{*}{5} & 1 & Square-Root Inverse
& 0.492 \scriptsize$\pm0.0$ & 0.320 \scriptsize$\pm0.0$
& 0.489 \scriptsize$\pm0.0$ & 0.659 \scriptsize$\pm0.0$
& 0.321 \scriptsize$\pm0.1$ & 0.716 \scriptsize$\pm0.2$
& 0.338 \scriptsize$\pm0.0$ & 1.383 \scriptsize$\pm0.2$
& 0.478 \scriptsize$\pm0.0$ & 2.278 \scriptsize$\pm0.2$
& 0.717 \scriptsize$\pm0.0$ & 1.992 \scriptsize$\pm0.0$ \\

& & 2 & Square-Root Inverse
& 0.494 \scriptsize$\pm0.0$ & 0.319 \scriptsize$\pm0.0$
& 0.492 \scriptsize$\pm0.0$ & 0.694 \scriptsize$\pm0.0$
& 0.306 \scriptsize$\pm0.0$ & 0.626 \scriptsize$\pm0.1$
& 0.348 \scriptsize$\pm0.0$ & 1.338 \scriptsize$\pm0.1$
& 0.471 \scriptsize$\pm0.0$ & 2.369 \scriptsize$\pm0.1$
& 0.728 \scriptsize$\pm0.0$ & 1.992 \scriptsize$\pm0.1$ \\

& & 3 & Square-Root Inverse
& 0.492 \scriptsize$\pm0.0$ & 0.320 \scriptsize$\pm0.0$
& 0.490 \scriptsize$\pm0.0$ & 0.657 \scriptsize$\pm0.0$
& 0.324 \scriptsize$\pm0.0$ & 0.824 \scriptsize$\pm0.1$
& 0.331 \scriptsize$\pm0.0$ & 1.289 \scriptsize$\pm0.0$
& 0.477 \scriptsize$\pm0.0$ & 2.354 \scriptsize$\pm0.1$
& 0.728 \scriptsize$\pm0.0$ & 1.911 \scriptsize$\pm0.1$ \\

\addlinespace[1pt]
\cmidrule(lr){2-16}
\addlinespace[1pt]

& \multirow{3}{*}{9} & 1 & Square-Root Inverse
& 0.496 \scriptsize$\pm0.0$ & 0.327 \scriptsize$\pm0.0$
& 0.493 \scriptsize$\pm0.0$ & 0.662 \scriptsize$\pm0.0$
& 0.320 \scriptsize$\pm0.1$ & 0.666 \scriptsize$\pm0.2$
& 0.347 \scriptsize$\pm0.0$ & 1.307 \scriptsize$\pm0.1$
& 0.475 \scriptsize$\pm0.0$ & 2.302 \scriptsize$\pm0.2$
& 0.729 \scriptsize$\pm0.0$ & 2.035 \scriptsize$\pm0.1$ \\

& & 2 & Square-Root Inverse
& 0.486 \scriptsize$\pm0.0$ & 0.317 \scriptsize$\pm0.0$
& 0.484 \scriptsize$\pm0.0$ & 0.666 \scriptsize$\pm0.0$
& 0.271 \scriptsize$\pm0.0$ & 0.535 \scriptsize$\pm0.2$
& 0.336 \scriptsize$\pm0.0$ & 1.417 \scriptsize$\pm0.1$
& 0.467 \scriptsize$\pm0.0$ & 2.385 \scriptsize$\pm0.1$
& 0.720 \scriptsize$\pm0.0$ & 1.987 \scriptsize$\pm0.1$ \\

& & 3 & Square-Root Inverse
& 0.494 \scriptsize$\pm0.0$ & 0.318 \scriptsize$\pm0.0$
& 0.492 \scriptsize$\pm0.0$ & 0.656 \scriptsize$\pm0.0$
& 0.310 \scriptsize$\pm0.0$ & 0.683 \scriptsize$\pm0.1$
& 0.343 \scriptsize$\pm0.0$ & 1.321 \scriptsize$\pm0.1$
& 0.476 \scriptsize$\pm0.0$ & 2.265 \scriptsize$\pm0.1$
& 0.727 \scriptsize$\pm0.0$ & 2.045 \scriptsize$\pm0.1$ \\

\addlinespace[1pt]
\cmidrule(lr){2-16}
\addlinespace[1pt]

& \multirow{3}{*}{15} & 1 & Square-Root Inverse
& 0.489 \scriptsize$\pm0.0$ & 0.313 \scriptsize$\pm0.0$
& 0.486 \scriptsize$\pm0.0$ & 0.665 \scriptsize$\pm0.0$
& 0.300 \scriptsize$\pm0.1$ & 0.643 \scriptsize$\pm0.2$
& 0.340 \scriptsize$\pm0.0$ & 1.437 \scriptsize$\pm0.2$
& 0.470 \scriptsize$\pm0.0$ & 2.228 \scriptsize$\pm0.1$
& 0.717 \scriptsize$\pm0.0$ & 2.082 \scriptsize$\pm0.1$ \\

& & 2 & Square-Root Inverse
& 0.493 \scriptsize$\pm0.0$ & 0.324 \scriptsize$\pm0.0$
& 0.491 \scriptsize$\pm0.0$ & 0.654 \scriptsize$\pm0.0$
& 0.317 \scriptsize$\pm0.1$ & 0.681 \scriptsize$\pm0.3$
& 0.337 \scriptsize$\pm0.0$ & 1.370 \scriptsize$\pm0.1$
& 0.479 \scriptsize$\pm0.0$ & 2.309 \scriptsize$\pm0.2$
& 0.724 \scriptsize$\pm0.0$ & 2.051 \scriptsize$\pm0.1$ \\

& & 3 & Square-Root Inverse
& 0.495 \scriptsize$\pm0.0$ & 0.321 \scriptsize$\pm0.0$
& 0.492 \scriptsize$\pm0.0$ & 0.655 \scriptsize$\pm0.0$
& 0.314 \scriptsize$\pm0.1$ & 0.669 \scriptsize$\pm0.3$
& 0.347 \scriptsize$\pm0.0$ & 1.306 \scriptsize$\pm0.1$
& 0.471 \scriptsize$\pm0.0$ & 2.241 \scriptsize$\pm0.1$
& 0.733 \scriptsize$\pm0.0$ & 2.026 \scriptsize$\pm0.0$ \\

\midrule
\multicolumn{16}{l}{\textit{LDS + \featuremds variants}} \\[2pt]
\multirow{12}{*}{\featuremds}

& \multirow{4}{*}{3} & \multirow{2}{*}{1} & Inverse
& 0.497 \scriptsize$\pm0.0$ & 0.319 \scriptsize$\pm0.0$
& 0.494 \scriptsize$\pm0.0$ & 0.670 \scriptsize$\pm0.0$
& 0.361 \scriptsize$\pm0.1$ & 0.751 \scriptsize$\pm0.1$
& 0.344 \scriptsize$\pm0.0$ & 1.359 \scriptsize$\pm0.1$
& 0.465 \scriptsize$\pm0.0$ & 2.170 \scriptsize$\pm0.1$
& 0.756 \scriptsize$\pm0.0$ & 1.994 \scriptsize$\pm0.1$ \\

& & & Square-Root Inverse
& 0.498 \scriptsize$\pm0.0$ & 0.323 \scriptsize$\pm0.0$
& 0.495 \scriptsize$\pm0.0$ & 0.675 \scriptsize$\pm0.0$
& 0.316 \scriptsize$\pm0.1$ & 0.637 \scriptsize$\pm0.2$
& 0.353 \scriptsize$\pm0.0$ & 1.348 \scriptsize$\pm0.1$
& 0.475 \scriptsize$\pm0.0$ & 2.325 \scriptsize$\pm0.1$
& 0.729 \scriptsize$\pm0.0$ & 2.031 \scriptsize$\pm0.1$ \\

& & \multirow{2}{*}{2} & Inverse
& 0.490 \scriptsize$\pm0.0$ & 0.314 \scriptsize$\pm0.0$
& 0.488 \scriptsize$\pm0.0$ & 0.673 \scriptsize$\pm0.0$
& 0.330 \scriptsize$\pm0.0$ & 0.653 \scriptsize$\pm0.1$
& 0.343 \scriptsize$\pm0.0$ & 1.347 \scriptsize$\pm0.1$
& 0.467 \scriptsize$\pm0.0$ & 2.227 \scriptsize$\pm0.1$
& 0.726 \scriptsize$\pm0.0$ & 2.046 \scriptsize$\pm0.1$ \\

& & & Square-Root Inverse
& 0.488 \scriptsize$\pm0.0$ & 0.319 \scriptsize$\pm0.0$
& 0.485 \scriptsize$\pm0.0$ & 0.641 \scriptsize$\pm0.0$
& 0.274 \scriptsize$\pm0.0$ & 0.545 \scriptsize$\pm0.1$
& 0.323 \scriptsize$\pm0.0$ & 1.269 \scriptsize$\pm0.1$
& 0.476 \scriptsize$\pm0.0$ & 2.246 \scriptsize$\pm0.1$
& 0.725 \scriptsize$\pm0.0$ & 1.949 \scriptsize$\pm0.1$ \\

\cmidrule(lr){2-16}

& \multirow{4}{*}{5} & \multirow{2}{*}{1} & Inverse
& 0.493 \scriptsize$\pm0.0$ & 0.320 \scriptsize$\pm0.0$
& 0.490 \scriptsize$\pm0.0$ & 0.663 \scriptsize$\pm0.0$
& 0.369 \scriptsize$\pm0.1$ & 0.759 \scriptsize$\pm0.3$
& 0.347 \scriptsize$\pm0.0$ & 1.328 \scriptsize$\pm0.2$
& 0.467 \scriptsize$\pm0.0$ & 2.264 \scriptsize$\pm0.1$
& 0.732 \scriptsize$\pm0.0$ & 2.000 \scriptsize$\pm0.1$ \\

& & & Square-Root Inverse
& 0.490 \scriptsize$\pm0.0$ & 0.316 \scriptsize$\pm0.0$
& 0.487 \scriptsize$\pm0.0$ & 0.651 \scriptsize$\pm0.0$
& 0.295 \scriptsize$\pm0.1$ & 0.624 \scriptsize$\pm0.2$
& 0.333 \scriptsize$\pm0.0$ & 1.272 \scriptsize$\pm0.1$
& 0.474 \scriptsize$\pm0.0$ & 2.229 \scriptsize$\pm0.2$
& 0.723 \scriptsize$\pm0.0$ & 1.997 \scriptsize$\pm0.1$ \\

& & \multirow{2}{*}{2} & Inverse
& 0.493 \scriptsize$\pm0.0$ & 0.321 \scriptsize$\pm0.0$
& 0.490 \scriptsize$\pm0.0$ & 0.650 \scriptsize$\pm0.0$
& 0.357 \scriptsize$\pm0.1$ & 0.695 \scriptsize$\pm0.2$
& 0.342 \scriptsize$\pm0.0$ & 1.304 \scriptsize$\pm0.1$
& 0.471 \scriptsize$\pm0.0$ & 2.224 \scriptsize$\pm0.1$
& 0.728 \scriptsize$\pm0.0$ & 2.000 \scriptsize$\pm0.1$ \\

& & & Square-Root Inverse
& 0.489 \scriptsize$\pm0.0$ & 0.315 \scriptsize$\pm0.0$
& 0.486 \scriptsize$\pm0.0$ & 0.655 \scriptsize$\pm0.0$
& 0.262 \scriptsize$\pm0.0$ & 0.538 \scriptsize$\pm0.1$
& 0.340 \scriptsize$\pm0.0$ & 1.318 \scriptsize$\pm0.1$
& 0.476 \scriptsize$\pm0.0$ & 2.414 \scriptsize$\pm0.1$
& 0.707 \scriptsize$\pm0.0$ & 2.088 \scriptsize$\pm0.1$ \\

\cmidrule(lr){2-16}

& \multirow{4}{*}{9} & \multirow{2}{*}{1} & Inverse
& 0.500 \scriptsize$\pm0.0$ & 0.325 \scriptsize$\pm0.0$
& 0.497 \scriptsize$\pm0.0$ & 0.670 \scriptsize$\pm0.0$
& 0.335 \scriptsize$\pm0.0$ & 0.708 \scriptsize$\pm0.1$
& 0.346 \scriptsize$\pm0.0$ & 1.352 \scriptsize$\pm0.1$
& 0.471 \scriptsize$\pm0.0$ & 2.205 \scriptsize$\pm0.1$
& 0.753 \scriptsize$\pm0.0$ & 1.950 \scriptsize$\pm0.1$ \\

& & & Square-Root Inverse
& 0.489 \scriptsize$\pm0.0$ & 0.315 \scriptsize$\pm0.0$
& 0.487 \scriptsize$\pm0.0$ & 0.665 \scriptsize$\pm0.0$
& 0.289 \scriptsize$\pm0.0$ & 0.603 \scriptsize$\pm0.1$
& 0.338 \scriptsize$\pm0.0$ & 1.398 \scriptsize$\pm0.1$
& 0.468 \scriptsize$\pm0.0$ & 2.381 \scriptsize$\pm0.1$
& 0.728 \scriptsize$\pm0.0$ & 2.067 \scriptsize$\pm0.0$ \\

& & \multirow{2}{*}{2} & Inverse
& 0.494 \scriptsize$\pm0.0$ & 0.317 \scriptsize$\pm0.0$
& 0.491 \scriptsize$\pm0.0$ & 0.664 \scriptsize$\pm0.0$
& 0.328 \scriptsize$\pm0.0$ & 0.673 \scriptsize$\pm0.1$
& 0.340 \scriptsize$\pm0.0$ & 1.344 \scriptsize$\pm0.1$
& 0.470 \scriptsize$\pm0.0$ & 2.326 \scriptsize$\pm0.1$
& 0.736 \scriptsize$\pm0.0$ & 2.042 \scriptsize$\pm0.1$ \\

& & & Square-Root Inverse
& 0.488 \scriptsize$\pm0.0$ & 0.318 \scriptsize$\pm0.0$
& 0.485 \scriptsize$\pm0.0$ & 0.670 \scriptsize$\pm0.0$
& 0.278 \scriptsize$\pm0.0$ & 0.554 \scriptsize$\pm0.1$
& 0.327 \scriptsize$\pm0.0$ & 1.307 \scriptsize$\pm0.1$
& 0.477 \scriptsize$\pm0.0$ & 2.492 \scriptsize$\pm0.1$
& 0.719 \scriptsize$\pm0.0$ & 2.057 \scriptsize$\pm0.0$ \\

\bottomrule[1.5pt]
\end{tabular}%
    }
\label{tab:povertymap_ablation}
\end{table*}

\subsection{Kernel Type for \labelmds and \featuremds}
We further investigate the impact of different kernel choices for \labelmds and \featuremds, beyond the default configuration that uses Gaussian kernels. We experiment with three different kernel types: \textit{Gaussian}, \textit{Laplacian}, and \textit{Triangular} kernel, evaluate their effects on \labelmds and \featuremds. We use kernel size $l = 5$ and the standard deviation $\sigma = 2$ for all kernels and report results on \PovertyMap in Table~\ref{tab:povertymap-kernel-ablation}. As the table illustrates, all kernel types provide improvements over the ERM baseline, especially in \textit{few-shot} and \textit{zero-shot} regions. Moreover, Laplacian gives the best results for both \labelmds and \featuremds. These results suggest that both \labelmds and \featuremds are robust to different smoothing kernel types.
\begin{table*}[!t]
  \centering
  \caption{\small{Ablation study of different kernel types for \labelmds and \featuremds on \PovertyMap}}
  \vspace{-3pt}
  \setlength{\tabcolsep}{3pt}
  \renewcommand{\arraystretch}{1.15}
  \resizebox{\textwidth}{!}{%
  \begin{tabular}{lccccccccccc}
    \toprule[1.5pt]
    \multirow{3}{*}{Algorithm}
      & \multirow{3}{*}{Overall}
      & \multicolumn{2}{c}{Test Error (by attribute)}
      & \multicolumn{8}{c}{Test Error (by shot)} \\
    \cmidrule(lr){3-4} \cmidrule(lr){5-12}
      &
      & \multirow{2}{*}{Average}
      & \multirow{2}{*}{Worst}
      & \multicolumn{2}{c}{Many}
      & \multicolumn{2}{c}{Medium}
      & \multicolumn{2}{c}{Few}
      & \multicolumn{2}{c}{Zero} \\
    \cmidrule(lr){5-6} \cmidrule(lr){7-8} \cmidrule(lr){9-10} \cmidrule(lr){11-12}
      &  &  &
      & Average & Worst
      & Average & Worst
      & Average & Worst
      & Average & Worst \\
    \midrule
    ERM
      & 0.504
      & 0.502 & 0.679
      & 0.256 & 0.504
      & 0.335 & 1.356
      & 0.494 & 2.452
      & 0.744 & 1.996 \\

    \midrule\midrule
    \multicolumn{12}{l}{\labelmds:} \\
    \midrule
    GAUSSIAN KERNEL
      & 0.486
      & 0.484 & 0.666
      & 0.271 & 0.535
      & 0.336 & 1.417
      & 0.467 & 2.385
      & 0.720 & 1.987 \\
    LAPLACIAN KERNEL
      & 0.479
      & 0.477 & 0.624
      & 0.243 & 0.534
      & 0.312 & 1.191
      & 0.468 & 2.494
      & 0.716 & 2.037 \\
    TRIANGULAR KERNEL
      & 0.480
      & 0.478 & 0.626
      & 0.221 & 0.359
      & 0.323 & 1.230
      & 0.481 & 2.302
      & 0.684 & 2.057 \\

    \midrule\midrule
    \multicolumn{12}{l}{\featuremds:} \\
    \midrule
    GAUSSIAN KERNEL
      & 0.488
      & 0.485 & 0.670
      & 0.278 & 0.554
      & 0.327 & 1.307
      & 0.477 & 2.492
      & 0.719 & 2.057 \\
    LAPLACIAN KERNEL
      & 0.485
      & 0.482 & 0.639
      & 0.434 & 0.862
      & 0.356 & 1.325
      & 0.452 & 2.084
      & 0.713 & 1.929 \\
    TRIANGULAR KERNEL
      & 0.486
      & 0.483 & 0.668
      & 0.371 & 0.558
      & 0.335 & 1.277
      & 0.459 & 2.541
      & 0.733 & 2.007 \\

    \bottomrule[1.5pt]
  \end{tabular}%
  }
  \label{tab:povertymap-kernel-ablation}
  \vspace{-5pt}
\end{table*}

\subsection{Training Loss for \labelmds and \featuremds}
In the main paper, we use $L_1$ loss during training for all datasets. Besides $L_1$, we also study the effect of different training loss functions on \labelmds and \featuremds. Specifically, we compare three common loss functions that people use for regression tasks: $L_1$ loss, MSE loss, and the Huber loss. Results on \PovertyMap are shown in Table~\ref{tab:povertymap-loss-ablation}. We notice that there are no significant performance differences between the losses and all three losses gain improvements from the baseline, indicating that \labelmds and \featuremds are robust to the choice of different loss functions.
\begin{table*}[!t]
  \centering
  \caption{\small{Ablation study of different loss functions used during training for \labelmds and \featuremds on \PovertyMap}}
  \vspace{-3pt}
  \setlength{\tabcolsep}{3pt}
  \renewcommand{\arraystretch}{1.15}
  \resizebox{\textwidth}{!}{%
  \begin{tabular}{lccccccccccc}
    \toprule[1.5pt]
    \multirow{3}{*}{Algorithm}
      & \multirow{3}{*}{Overall}
      & \multicolumn{2}{c}{Test Error (by attribute)}
      & \multicolumn{8}{c}{Test Error (by shot)} \\
    \cmidrule(lr){3-4} \cmidrule(lr){5-12}
      &
      & \multirow{2}{*}{Average}
      & \multirow{2}{*}{Worst}
      & \multicolumn{2}{c}{Many}
      & \multicolumn{2}{c}{Medium}
      & \multicolumn{2}{c}{Few}
      & \multicolumn{2}{c}{Zero} \\
    \cmidrule(lr){5-6} \cmidrule(lr){7-8} \cmidrule(lr){9-10} \cmidrule(lr){11-12}
      &  &  &
      & Average & Worst
      & Average & Worst
      & Average & Worst
      & Average & Worst \\
    \midrule
    ERM
      & 0.504
      & 0.502 & 0.679
      & 0.256 & 0.504
      & 0.335 & 1.356
      & 0.494 & 2.452
      & 0.744 & 1.996 \\

    \midrule\midrule
    \multicolumn{12}{l}{\labelmds:} \\
    \midrule
    L1
      & 0.486
      & 0.484 & 0.666
      & 0.271 & 0.535
      & 0.336 & 1.417
      & 0.467 & 2.385
      & 0.720 & 1.987 \\
    MSE
      & 0.481
      & 0.479 & 0.700
      & 0.321 & 0.749
      & 0.307 & 1.227
      & 0.465 & 2.501
      & 0.734 & 2.142 \\
    HUBER LOSS
      & 0.484
      & 0.481 & 0.650
      & 0.341 & 0.806
      & 0.335 & 1.429
      & 0.471 & 2.142
      & 0.699 & 1.981 \\

    \midrule\midrule
    \multicolumn{12}{l}{\featuremds:} \\
    \midrule
    L1
      & 0.488
      & 0.485 & 0.670
      & 0.278 & 0.554
      & 0.327 & 1.307
      & 0.477 & 2.492
      & 0.719 & 2.057 \\
    L2
      & 0.494
      & 0.490 & 0.665
      & 0.160 & 0.199
      & 0.316 & 1.285
      & 0.475 & 2.615
      & 0.765 & 2.287 \\
    HUBER LOSS
      & 0.486
      & 0.483 & 0.667
      & 0.285 & 0.378
      & 0.335 & 1.279
      & 0.471 & 2.422
      & 0.713 & 1.967 \\

    \bottomrule[1.5pt]
  \end{tabular}%
  }
  \label{tab:povertymap-loss-ablation}
  \vspace{-5pt}
\end{table*}

\subsection{Average Metric Rank}
\label{app:ranking_details}
In this section, we provide the average ranking of methods across metrics for each dataset separately. Average ranking for \UTKFace, \SkyFinder, \PovertyMap, \CodeNet are presented in Table \ref{tab:utkface_ranking}, Table \ref{tab:skyfinder_ranking}, Table \ref{tab:povertymap_ranking} and Table \ref{tab:codenet_ranking} respectively. For each metric, methods are ranked based on their performance, where rank 1 corresponds to the best-performing method. We then compute the average rank of each method across all reported metrics within a dataset. The ranking includes overall metrics as well as group-wise metrics across many-shot, medium-shot, few-shot, and zero-shot regions. Our method achieves the best average ranking on \UTKFace, \PovertyMap, and \CodeNet, while remaining highly competitive on \SkyFinder. These results suggest that our method provides consistently strong performance across diverse evaluation metrics and datasets, demonstrating strong generalization ability across diverse task domains.

\begin{table}[!t]
\centering
\small
\setlength{\abovecaptionskip}{3pt}
\setlength{\belowcaptionskip}{0pt}
\setlength{\tabcolsep}{10pt}
\renewcommand{\arraystretch}{1.15}
  \caption{\small{Average ranking of methods across metrics on \UTKFace. Lower average rank is better.}}
  \begin{tabular}{clc}
    \toprule[1.5pt]
    Rank & Method & Average Metric Rank \\
    \midrule
    1  & \featuremds         & 2.67  \\
    2  & SqrtReWeight  & 3.42  \\
    3  & LDS           & 3.46  \\
    4  & \labelmds         & 4.13  \\
    5  & RnC           & 5.63  \\
    6  & GroupDRO      & 5.92  \\
    7  & \labelmds + \featuremds   & 6.50  \\
    8  & ERM           & 6.63  \\
    9  & Resample      & 7.92  \\
    10 & DANN          & 10.42 \\
    11 & CBLoss        & 10.50 \\
    12 & ReWeight      & 10.83 \\
    \bottomrule[1.5pt]
  \end{tabular}

  \label{tab:utkface_ranking}
\end{table}
\begin{table}[!t]
\centering
\small
\setlength{\abovecaptionskip}{3pt}
\setlength{\belowcaptionskip}{0pt}
\setlength{\tabcolsep}{10pt}
\renewcommand{\arraystretch}{1.15}
\caption{\small{Average ranking of methods across metrics on \SkyFinder. Lower average rank is better.}}
\begin{tabular}{clc}
\toprule[1.5pt]
Rank & Method & Average Metric Rank \\
\midrule
1  & RnC & 2.75 \\
2  & \featuremds & 3.50 \\
3  & SqrtReWeight & 4.17 \\
4  & \labelmds & 4.29 \\
5  & \labelmds + \featuremds & 5.29 \\
6  & GroupDRO & 5.75 \\
7  & ERM & 6.33 \\
8  & Resample & 6.79 \\
9  & LDS & 9.04 \\
10 & CBLoss & 9.58 \\
11 & DANN & 9.75 \\
12 & ReWeight & 10.75 \\
\bottomrule[1.5pt]
\end{tabular}
\label{tab:skyfinder_ranking}
\end{table}
\begin{table}[!t]
\centering
\small
\setlength{\abovecaptionskip}{3pt}
\setlength{\belowcaptionskip}{0pt}
\setlength{\tabcolsep}{10pt}
\renewcommand{\arraystretch}{1.15}
\caption{\small{Average ranking of methods across metrics on \PovertyMap. Lower average rank is better.}}
\begin{tabular}{clc}
\toprule[1.5pt]
Rank & Method & Average Metric Rank\\
\midrule
1  & \labelmds & 3.50 \\
2  & \featuremds & 4.50 \\
3  & \labelmds + \featuremds & 4.71 \\
4  & GroupDRO & 5.38 \\
5  & RnC & 5.71 \\
6  & ERM & 5.88 \\
7  & Resample & 6.67 \\
8  & LDS & 6.88 \\
9  & SqrtReWeight & 7.33 \\
10 & ReWeight & 8.08 \\
11 & CBLoss & 8.29 \\
12 & DANN & 11.08 \\
\bottomrule[1.5pt]
\end{tabular}
\label{tab:povertymap_ranking}
\end{table}
\begin{table}[!t]
  \centering
  \small                                       \setlength{\abovecaptionskip}{3pt}
  \setlength{\belowcaptionskip}{0pt}           \setlength{\tabcolsep}{10pt}                 \renewcommand{\arraystretch}{1.15}
  \caption{\small{Average ranking of methods across metrics on \CodeNet. Lower average rank is better.}}
  \begin{tabular}{clc}
  \toprule[1.5pt]
  Rank & Method & Average Metric Rank \\
  \midrule          
  1  & \labelmds & 2.28 \\
  2  & SqrtReWeight & 3.61 \\              
  3  & \featuremds & 4.06 \\          
  4  & CBLoss & 4.56 \\
  5  & \labelmds + \featuremds & 4.67 \\
  6  & ReWeight & 5.72 \\
  7  & LDS & 5.83 \\
  8  & ERM & 6.78 \\
  9  & DANN & 7.50 \\
\bottomrule[1.5pt]
\end{tabular}
\label{tab:codenet_ranking}
\end{table}

\subsection{Analysis of Interpolation \&\ Extrapolation}
\label{app:interp-extrap}
We construct a curated subset of \UTKFace with missing target regions in the train sets while evaluating on the original test set. As shown in Table
\ref{tab:erm_mds_comparison}, both \labelmds and \featuremds consistently improve over ERM across both MAE and GM metrics. Notably, the gains are more pronounced in the \textit{interpolation} and \textit{extrapolation} regions, suggesting that our smoothing methods help transfer information across related target regions and improve generalization to unseen or underrepresented targets.

 \begin{table*}[!t]
\centering
\setlength{\tabcolsep}{3pt}
\caption{\small{Interpolation \&\ extrapolation results on a curated subset of \UTKFace}}
\begin{tabular}{l|cccc|cccc}
\toprule[1.5pt]
Metrics 
& \multicolumn{4}{c|}{MAE \ $\downarrow$} 
& \multicolumn{4}{c}{GM \ $\downarrow$} \\
\midrule
Shot 
& All & w/ data & Interp. & Extrap. 
& All & w/ data & Interp. & Extrap. \\
\midrule

ERM 
& 14.02 & 10.10 & 13.53 & 25.71 
& 8.20 & 5.79 & 9.42 & 17.92 \\

\midrule

\featuremds 
& 12.65 & 10.05 & 11.98 & 20.89 
& 7.56 & 5.93 & 7.86 & 14.07 \\

\labelmds
& 13.11 & 10.30 & 12.95 & 21.21 & 7.97 & 6.20 & 8.75 &14.08 \\
\midrule

Ours (best) vs.\ ERM  
& \textbf{\textcolor{green!60!black}{+1.37}} 
& \textbf{\textcolor{green!60!black}{+0.05}}
& \textbf{\textcolor{green!60!black}{+1.56}} 
& \textbf{\textcolor{green!60!black}{+4.82}}
& \textbf{\textcolor{green!60!black}{+0.65}}
& \textbf{\textcolor{lightblue}{-0.15}}
& \textbf{\textcolor{green!60!black}{+1.56}}
& \textbf{\textcolor{green!60!black}{+3.84}} \\

\bottomrule[1.5pt]
\end{tabular}

\label{tab:erm_mds_comparison}
\end{table*}

\subsection{Resilience to Reduced Training Data}
\label{app:reduced_shot_wise}
The success of modern deep learning methods has largely relied on the availability of large-scale labeled datasets. However, collecting and annotating such datasets is often costly and time-consuming in real-world applications. Thus, it is important to evaluate models under limited training data. We subsample \UTKFace to $50\%$, $20\%$, and $10\%$ of the original training data, and train ERM, \labelmds, and \featuremds separately on each subsampled subset. Shot-wise results for $50\%$, $20\%$ and $10\%$ are in 
Table~\ref{tab:utkface_50pct}, Table~\ref{tab:utkface_20pct}, Table~\ref{tab:utkface_10pct}. We observe \labelmds and \featuremds are more robust to reduced training data and achieve better performance gains. Moreover, as the training set becomes smaller, the performance gap between our methods and the baseline increases and both \labelmds and \featuremds gain from \textit{zero-shot} and \textit{few-shot} regions.

\begin{table*}[!t]
 \centering

  \setlength{\tabcolsep}{3pt}
  \renewcommand{\arraystretch}{1.15}
  
  \caption{\small{Results on \UTKFace with 50\% training data}}
     \resizebox{\textwidth}{!}{%
  \begin{tabular}{lcccccccccccc}
    \toprule[1.5pt]
    \multirow{3}{*}{Algorithm}
      & \multicolumn{2}{c}{Overall}
      & \multicolumn{2}{c}{MAE (by attribute)}
      & \multicolumn{8}{c}{MAE (by shot)} \\
    \cmidrule(lr){2-3} \cmidrule(lr){4-5} \cmidrule(lr){6-13}
      & \multirow{2}{*}{MAE}
      & \multirow{2}{*}{GM}
      & \multirow{2}{*}{Average}
      & \multirow{2}{*}{Worst}
      & \multicolumn{2}{c}{Many}
      & \multicolumn{2}{c}{Medium}
      & \multicolumn{2}{c}{Few}
      & \multicolumn{2}{c}{Zero} \\
    \cmidrule(lr){6-7} \cmidrule(lr){8-9} \cmidrule(lr){10-11} \cmidrule(lr){12-13}
      &  &  &  &
      & Average & Worst
      & Average & Worst
      & Average & Worst
      & Average & Worst \\
    \midrule
    ERM   & $8.13$ & $7.84$ & $7.98$ & $10.33$ & $2.83$ & $6.42$ & $6.06$ & $23.12$ & $7.72$ & $17.65$ & $10.99$ & $82.25$ \\
    \labelmds & $7.99$ & $7.74$ & $7.84$ & $9.71$  & $3.22$ & $5.97$ & $6.32$ & $19.85$ & $7.55$ & $14.77$ & $10.41$ & $82.99$ \\
    \featuremds & $8.02$ & $7.74$ & $7.86$ & $9.49$  & $3.22$ & $5.99$ & $6.31$ & $18.09$ & $7.50$ & $15.17$ & $10.52$ & $82.91$ \\
    \midrule
    Ours (best) vs.\ ERM
    & \textbf{\textcolor{green!60!black}{+0.14}}
    & \textbf{\textcolor{green!60!black}{+0.10}}
    & \textbf{\textcolor{green!60!black}{+0.14}}
    & \textbf{\textcolor{green!60!black}{+0.84}}
    & \textbf{\textcolor{lightblue}{-0.39}}
    & \textbf{\textcolor{green!60!black}{+0.45}}
    & \textbf{\textcolor{lightblue}{-0.25}}
    & \textbf{\textcolor{green!60!black}{+5.03}}
    & \textbf{\textcolor{green!60!black}{+0.22}}
    & \textbf{\textcolor{green!60!black}{+2.88}}
    & \textbf{\textcolor{green!60!black}{+0.58}}
    & \textbf{\textcolor{lightblue}{-0.66}} \\
    \bottomrule[1.5pt]
  \end{tabular}
  }
  \label{tab:utkface_50pct}

\end{table*}
\begin{table*}[!t]
  \centering

  \setlength{\tabcolsep}{3pt}
  \renewcommand{\arraystretch}{1.15}

   \caption{\small{Results on \UTKFace with 20\% training data}}
   \resizebox{\textwidth}{!}{%
  \begin{tabular}{lcccccccccccc}
    \toprule[1.5pt]
    \multirow{3}{*}{Algorithm}
      & \multicolumn{2}{c}{Overall}
      & \multicolumn{2}{c}{MAE (by attribute)}
      & \multicolumn{8}{c}{MAE (by shot)} \\
    \cmidrule(lr){2-3} \cmidrule(lr){4-5} \cmidrule(lr){6-13}
      & \multirow{2}{*}{MAE}
      & \multirow{2}{*}{GM}
      & \multirow{2}{*}{Average}
      & \multirow{2}{*}{Worst}
      & \multicolumn{2}{c}{Many}
      & \multicolumn{2}{c}{Medium}
      & \multicolumn{2}{c}{Few}
      & \multicolumn{2}{c}{Zero} \\
    \cmidrule(lr){6-7} \cmidrule(lr){8-9} \cmidrule(lr){10-11} \cmidrule(lr){12-13}
      &  &  &  &
      & Average & Worst
      & Average & Worst
      & Average & Worst
      & Average & Worst \\
    \midrule
    ERM   & $10.33$ & $9.96$ & $10.11$ & $12.14$ & $3.75$ & $4.02$  & $6.49$ & $24.89$ & $9.20$ & $22.69$ & $13.69$ & $95.90$ \\
    \labelmds & $10.08$ & $9.79$ & $9.89$  & $11.19$ & $5.17$ & $5.61$  & $6.74$ & $23.44$ & $8.87$ & $22.67$ & $13.14$ & $82.98$ \\
    \featuremds & $10.12$ & $9.81$ & $9.92$  & $11.66$ & $4.60$ & $5.73$  & $6.77$ & $20.27$ & $8.91$ & $19.17$ & $13.18$ & $69.23$ \\
    \midrule
    Ours (best) vs.\ ERM
    & \textbf{\textcolor{green!60!black}{+0.25}}
    & \textbf{\textcolor{green!60!black}{+0.17}}
    & \textbf{\textcolor{green!60!black}{+0.22}}
    & \textbf{\textcolor{green!60!black}{+0.95}}
    & \textbf{\textcolor{lightblue}{-0.85}}
    & \textbf{\textcolor{lightblue}{-1.59}}
    & \textbf{\textcolor{lightblue}{-0.25}}
    & \textbf{\textcolor{green!60!black}{+4.62}}
    & \textbf{\textcolor{green!60!black}{+0.33}}
    & \textbf{\textcolor{green!60!black}{+3.52}}
    & \textbf{\textcolor{green!60!black}{+0.55}}
    & \textbf{\textcolor{green!60!black}{+26.67}} \\
    \bottomrule[1.5pt]
  \end{tabular}
  }
  \label{tab:utkface_20pct}
 
\end{table*}
\begin{table*}[!t]
  \centering
    \setlength{\tabcolsep}{3pt}
   \caption{\small{Results on \UTKFace with 10\% training data}}

\resizebox{\textwidth}{!}{%
\begin{tabular}{lcccccccccccc}
    \toprule[1.5pt]
    \multirow{3}{*}{Algorithm}
      & \multicolumn{2}{c}{Overall}
      & \multicolumn{2}{c}{MAE (by attribute)}
      & \multicolumn{8}{c}{MAE (by shot)} \\
    \cmidrule(lr){2-3} \cmidrule(lr){4-5} \cmidrule(lr){6-13}
      & \multirow{2}{*}{MAE}
      & \multirow{2}{*}{GM}
      & \multirow{2}{*}{Average}
      & \multirow{2}{*}{Worst}
      & \multicolumn{2}{c}{Many}
      & \multicolumn{2}{c}{Medium}
      & \multicolumn{2}{c}{Few}
      & \multicolumn{2}{c}{Zero} \\
    \cmidrule(lr){6-7} \cmidrule(lr){8-9} \cmidrule(lr){10-11} \cmidrule(lr){12-13}
      &  &  &  &
      & Average & Worst
      & Average & Worst
      & Average & Worst
      & Average & Worst \\
    \midrule
    ERM   & $14.02$ & $13.42$ & $13.65$ & $15.81$ & $--$ & $--$ & $8.32$ & $31.49$ & $10.83$ & $34.13$ & $18.55$ & $86.31$ \\
    \labelmds & $13.11$ & $12.64$ & $12.80$ & $14.75$ & $--$ & $--$ & $9.28$ & $34.90$ & $10.71$ & $27.64$ & $16.35$ & $85.21$ \\
    \featuremds & $12.65$ & $12.19$ & $12.35$ & $14.19$ & $--$ & $--$ & $9.04$ & $35.48$ & $10.46$ & $26.35$ & $15.65$ & $85.07$ \\
    \midrule
    Ours (best) vs.\ ERM
    & \textbf{\textcolor{green!60!black}{+1.37}}
    & \textbf{\textcolor{green!60!black}{+1.23}}
    & \textbf{\textcolor{green!60!black}{+1.30}}
    & \textbf{\textcolor{green!60!black}{+1.62}}
    & $--$
    & $--$
    & \textbf{\textcolor{lightblue}{-0.72}}
    & \textbf{\textcolor{lightblue}{-3.99}}
    & \textbf{\textcolor{green!60!black}{+0.37}}
    & \textbf{\textcolor{green!60!black}{+7.78}}
    & \textbf{\textcolor{green!60!black}{+2.90}}
    & \textbf{\textcolor{green!60!black}{+1.24}} \\
    \bottomrule[1.5pt]
  \end{tabular}%
  }
\label{tab:utkface_10pct}
\end{table*}

\subsection{Broader Impacts}
\label{app:broader-impacts}
We believe \labelmds and \featuremds can have positive societal impact by making continuous prediction systems more reliable in real-world settings where data is observational, imbalanced, and affected by deployment shifts. Such settings include medical risk prediction, environmental sensing, poverty estimation, and other domains where models may otherwise rely on shortcuts tied to demographics, locations, devices, or data-collection conditions. By explicitly evaluating subgroup errors and improving performance in sparse or unseen target regions, our work may help reduce hidden failure modes missed by average regression metrics.

At the same time, several risks require careful consideration. First, our methods depend on the availability and quality of spurious attribute annotations. If the attributes are incomplete, noisy, or themselves sensitive, the resulting model may provide a false sense of robustness or introduce new biases. Second, improving regression robustness can make it easier to deploy models for predicting sensitive personal attributes, such as health or socioeconomic status, from human data. Such uses may reinforce discrimination if the predictions are used for screening, ranking, surveillance, or resource allocation without proper oversight. Finally, better interpolation and extrapolation over continuous targets may be misused to justify predictions in regions with little or no reliable training evidence. Therefore, our methods should be used with careful validation, transparent reporting of subgroup performance, and domain-specific ethical review before deployment in high-stakes applications.

\section{Dataset Details}
\label{app:dataset-details}

We provide detailed information of the four datasets used in our experiments to investigate DSR in this section. Table \ref{tab:dataset-summary} provides an overview of the each dataset.

\begin{table*}[!t]
  \centering
  \small
  \caption{Overview of the four DSR datasets used in our experiments.}
  \setlength{\tabcolsep}{8pt} 
  \renewcommand{\arraystretch}{1.15}
  \label{tab:dataset-summary}
  \begin{tabular}{l c cc cc ccc}
    \toprule[1.5pt]
    \multirow{2}{*}{Dataset} & \multirow{2}{*}{\# Attrs} & \multicolumn{2}{c}{Target Range} & \multicolumn{2}{c}{Attr Density} & \multicolumn{3}{c}{Split Sizes} \\
    \cmidrule(lr){3-4} \cmidrule(lr){5-6} \cmidrule(lr){7-9}
    & & Min & Max & Min & Max & Train & Val & Test \\
    \midrule

    UTKFace    & 5  & 1     & 116   & 830   & 8,392 & 17,620 & 2,753 & 3,730 \\
    SkyFinder  & 47 & -27.2 & 50.0  & 14    & 3,464 & 64,945 & 9,335 & 6,766 \\
    PovertyMap & 20 & -1.1  & 2.5   & 97    & 1,159 & 6,034  & 475   & 545   \\
    CodeNet    & 13 & 0     & 1,000 & 1,500 & 1,500 & 19,500 & 6,374 & 6,374 \\

    \bottomrule[1.5pt]
  \end{tabular}
\end{table*}

\subsection{ColoredRotatedMNIST}
\paragraph{Dataset Construction.}
\label{app:colored_rotated_mnist_details}
\ColoredRotatedMNIST is a synthetic regression dataset built on
MNIST~\cite{lecun1998mnist} digit ``2''. Each image is rotated to a
continuous angle $y \in (0^\circ, 180^\circ)$ and
placed on a solid-color background serving as the spurious attribute
$a \in \{\text{Red},\, \text{Blue},\, \text{Green},\, \text{Yellow}\}$.
The task is to predict the rotation angle $y$ from the image.
In the training set, we divide the $180^\circ$ range into four equal
subintervals of $45^\circ$ each, with each color assigned a dominant
subinterval: Red to $[0^\circ, 45^\circ)$, Blue to $[45^\circ, 90^\circ)$,
Green to $[90^\circ, 135^\circ)$, and Yellow to $[135^\circ, 180^\circ)$.
Within its dominant subinterval, each color has 10 samples per degree
(450 samples total); outside its dominant subinterval, each color has
only 50 samples uniformly drawn from each of the remaining three
$45^\circ$ subintervals (150 samples total). This induces a strong
spurious correlation between background color and rotation angle in
training. The validation and test sets are uniformly distributed across
all $(y, a)$ combinations (10 samples per degree per color), providing
an unbiased evaluation of generalization.

\paragraph{Detailed ERM Results}
\label{app:colored_rotated_mnist_results}
Table~\ref{tab:crmnist_results} reports results on \ColoredRotatedMNIST{} trained using ERM to evaluate the impact of DSR across MAE, MSE, and GM metrics. Many-shot groups achieve the lowest average errors across all metrics, confirming that sufficient training coverage leads to better generalization. In contrast, few- and zero-shot groups suffer the most, with worst-case errors far exceeding their averages, highlighting the existence of DSR.
\begin{table*}[!t]
  \centering
  \small
  \caption{\small{Results on \ColoredRotatedMNIST using ERM across MAE, MSE, and GM metrics}}
  \setlength{\tabcolsep}{3pt}
  \renewcommand{\arraystretch}{1.15}
  \label{tab:crmnist}
  \resizebox{\textwidth}{!}{%
  \begin{tabular}{lcccccccccc}
    \toprule[1.5pt]
    \multirow{3}{*}{Metric}
      & \multicolumn{2}{c}{Test Error (by attribute)}
      & \multicolumn{8}{c}{Test Error (by shot)} \\
    \cmidrule(lr){2-3} \cmidrule(lr){4-11}
      & \multirow{2}{*}{Average}
      & \multirow{2}{*}{Worst}
      & \multicolumn{2}{c}{Many}
      & \multicolumn{2}{c}{Medium}
      & \multicolumn{2}{c}{Few}
      & \multicolumn{2}{c}{Zero} \\
    \cmidrule(lr){4-5} \cmidrule(lr){6-7} \cmidrule(lr){8-9} \cmidrule(lr){10-11}
      &  &
      & Average & Worst
      & Average & Worst
      & Average & Worst
      & Average & Worst \\
    \midrule

    MAE
    & 12.90 & 14.16
    & 7.47  & 21.27
    & 20.18 & 20.81
    & 15.26 & 91.90
    & 13.50 & 74.29 \\

    MSE
    & 509.77 & 671.32
    & 122.76 & 1335.52
    & 819.46 &  908.84
    & 700.00 & 12399.98
    & 507.52 & 9507.13 \\

    GM
    & 6.88 & 7.18
    & 4.52  & 11.23
    & 13.12 & 13.45
    &  8.05 & 54.82
    &  7.56 & 37.89 \\

    \bottomrule[1.5pt]
  \end{tabular}%
  }
  \label{tab:crmnist_results}
\end{table*}

\subsection{UTKFace}
The \UTKFace dataset \cite{zhifei2017utkface} is a collection of more than 20,000 facial images, each labeled with age, gender, ethnicity. We predict age as the regression task, and we employ the five ethnicity groups as the spurious attribute defined as White, Black, Asian, Indian, and Others (e.g. Hispanic, Latino, Middle Eastern). From the complete dataset, we construct a training set with 17,620 images, a validation set with 2,753 images, and a test set with 3,730 images. Age ranges from 1 to 116 years. The train set exhibits significant imbalance across ethnic groups, with the White ethnicity comprising 10,222 samples (42.4\%) while the Others ethnicity accounts for only 1,711 samples (7.1\%). The test and validation sets are constructed uniformly.

\subsection{SkyFinder}
The original \SkyFinder dataset \cite{mihail2016skyfinder} dataset is a large-scale dataset of pixel-annotated images of the sky and other regions taken by outdoor webcam images. We use the camera ID as the spurious attribute and the in-the-wild temperature associated with each images as the regression target. We stratify the data across 47 webcams and split the full dataset into 64,945 training images, 9,335 validation images, and 6,766 test images.
Temperature ranges from -27.2 to 50.0 degrees Celsius. The train set exhibits significant imbalance across cameras, with the most frequent camera (ID 684) comprising 3,464 samples (5.3\%) while the least frequent camera (ID 4232) accounts for only 14 samples (0.02\%). The test and validation sets are constructed more uniformly.

\subsection{PovertyMap}
The \PovertyMap dataset is a subset of the original PovertyMap-WILDS \cite{koh2021wilds} benchmark dataset. The dataset consists of satellite images of rural and urban regions across multiple countries. We consider the spurious attribute as the country an image is of, and we predict on poverty index as the regression target. We create a training set of 6,034 images total, with 475 validation images and 545 test images. The wealth index ranges from $-1.1$ to $2.5$. The train set exhibits significant imbalance across countries, with Tanzania comprising 1,521 samples (25.2\%) while Togo accounts for only 97 samples (1.6\%), reflecting the uneven country representation in the original survey data. The test and validation sets are constructed to have a more uniform distribution across countries.

\subsection{CodeNet}
The original CodeNet dataset \cite{puri2021project} contains 14M coding samples along with metadata on metrics like programming language, code size, and code execution time. The pre-trained RLM-GemmaS-Code-V0 model filters the dataset to 7.3M  "Accepted" solutions \cite{akhauri2025regression} and predicts over the memory column. To avoid data leakage, we predict the CPU execution time column as the regression target instead, and we clamp target values between 0 and 1000. We consider programming language as the spurious attribute. Although the original dataset includes samples from 57 different programming languages, a select few languages (Python, C++, Java) dominate, while other languages have few samples. We curate a subset of the filtered dataset by taking samples from the 13 most common programming languages with at least 10K samples each within the target range. We then create a training set with 19,500 total samples, where each programming language has 1,500 samples and the target distribution follows a Gaussian distribution. We make the length of each bin 20 to construct the normal distribution. From the filtered dataset, we also curate the test and validation sets with approximately 500 samples each per language for a total of 6,374 samples per set and 12,748 total. Both the test and validation sets follow a uniform distribution per language with a bin size of 50. 

\section{Experimental Settings}
\label{app:exp-settings}

\paragraph{Image Regression Datasets.}
All experiments on image regression datasets (\UTKFace, \SkyFinder, \PovertyMap) 
are trained using a single NVIDIA A40 GPU with batch size 256 for 400 epochs.  We use SGD with learning rate 0.2, momentum 0.9, weight decay $10^{-4}$, and a cosine annealing learning rate schedule with decay rate 0.1 applied over 400 epochs.

For DANN \cite{ganin2016dann}, a 3-layer MLP domain classifier (hidden width 256) is attached to the shared encoder via a gradient reversal layer, with encoder and discriminator learning rates both set to 0.2, adversarial weight $\lambda = 1.0$, and 1 discriminator step per generator step.

For GroupDRO\cite{sagawa2020dro}, we use the same base optimizer as above with group step size $\eta = 0.01$.

For RnC \cite{zha2023rnc}, training follows a two-phase procedure. In the first 
phase, the encoder is trained for 400 epochs with learning rate 0.5, momentum 0.9, 
weight decay $10^{-4}$, temperature $\tau = 2$, $L_{1}$ label distance, and $L_{2}$ feature 
similarity. In the second phase, the encoder is frozen and a linear regressor is trained on top for 100 epochs using SGD with learning rate 0.05, momentum 0.9, weight decay 0, and cosine annealing schedule with decay rate 0.2.

For both \labelmds and \featuremds, we use kernel size $k = 5$ and standard 
deviation $\sigma = 2$ with \texttt{sqrt\_inv} reweighting. For \featuremds 
specifically, the feature centroid update begins at epoch 5.

\paragraph{Code Regression Dataset.}
Experiments on the code regression dataset fine-tune a pretrained seq2seq LLM \cite{akhauri2025regression} (\texttt{akhauriyash/RLM-GemmaS-Code-v0}) on a single NVIDIA A40 GPU, where only the decoder is fine-tuned while the encoder is kept frozen. We use AdamW with a learning rate of 2e-5, weight decay of 0.01, gradient clipping norm of 1.0, batch size of 16, and train for 20 epochs.

\end{document}